%% file: main.tex
\documentclass[Afour,sageh,times]{sagej}

\usepackage{moreverb,url}
\usepackage[T1]{fontenc}
\usepackage[colorlinks,bookmarksopen,bookmarksnumbered,citecolor=red,urlcolor=red]{hyperref}

\newcommand\BibTeX{{\rmfamily B\kern-.05em \textsc{i\kern-.025em b}\kern-.08em
T\kern-.1667em\lower.7ex\hbox{E}\kern-.125emX}}

\usepackage{array,tabularx}
\definecolor{qixuangreen}{rgb}{0,0.5,0}

\usepackage{placeins}
\usepackage{xltabular}   
\usepackage{colortbl}
\usepackage{xcolor}

\usepackage{fontawesome5}
\usepackage{tikz}
\definecolor{overviewbluefill}{HTML}{EDF3F9}
\definecolor{overviewblueline}{HTML}{6096CB}
\newcommand{\titlepanelmark}[1]{\tikz[remember picture,overlay]{\coordinate (#1) at (0,0);}}
\newcommand{\titlepanelbackground}{%
  \begin{tikzpicture}[remember picture,overlay]
    \coordinate (panelLeft) at ([xshift=-.5\textwidth-9pt]current page.center);
    \coordinate (panelRight) at ([xshift=.5\textwidth+9pt]current page.center);
    \draw[rounded corners=7pt,fill=overviewbluefill!72!white,
          draw=overviewblueline!28!white,line width=.4pt]
      ([yshift=22pt]panelLeft |- panelTitle) rectangle
      ([yshift=-8pt]panelRight |- panelKeywords);
  \end{tikzpicture}}
\newcommand{\titlelogo}[2][]{%
  \tikz[baseline=0pt]{\node[anchor=south west,inner sep=0pt,outer sep=0pt,
       blend mode=multiply] at (0,0){\includegraphics[#1]{#2}};}}
\usepackage{microtype,balance}
\begin{document}
\microtypesetup{activate=false}
\raggedbottom

\runninghead{}

\title{Agentic Robotics: A Functional Perspective}

\author{Shaoli Hu\affilnum{1,*},
Kaichen Zhou\affilnum{2,3,*},
Xuan Qi\affilnum{4,*},
Yuhong Zhao\affilnum{1,*},
Yue Ma\affilnum{1},
Feng Chen\affilnum{5},
Yuelei Li\affilnum{6},
Zhuoyang Liu\affilnum{7},
Ao Qu\affilnum{2},
Mazeyu Ji\affilnum{1},
Ruihai Wu\affilnum{6},
Xiaomeng Hu\affilnum{8},
Zhiqin Yang\affilnum{1},
Chenxin Li\affilnum{8},
Tsung-Yi Ho\affilnum{8},
Zhiyong Fan\affilnum{1},
Qifeng Chen\affilnum{1},
Jinhua Zhao\affilnum{2},
Wenhan Luo\affilnum{1},
Mengyu Wang\affilnum{3},
Mac Schwager\affilnum{9},
Paul Liang\affilnum{2},
Fangneng Zhan\affilnum{1}\,\href{mailto:fnzhan@ust.hk}{\textsuperscript{\scriptsize\faEnvelope[regular]}}}

\affiliation{\affilnum{1}The Hong Kong University of Science and Technology\\
\affilnum{2}Massachusetts Institute of Technology\\
\affilnum{3}Harvard University\\
\affilnum{4}University of Washington\\
\affilnum{5}The University of Hong Kong\\
\affilnum{6}California Institute of Technology\\
\affilnum{7}University of California, Berkeley\\
\affilnum{8}The Chinese University of Hong Kong\\
\affilnum{9}Stanford University\\
}

\corrauth{Fangneng Zhan, The Hong Kong University of Science and Technology, Hong Kong.}
\email{fnzhan@ust.hk}




\begin{abstract}

Agentic robotics studies how robots autonomously select and coordinate available capabilities to achieve goals, using interaction feedback to revise their behavior. Recent foundation models offer new ways to interpret goals, plan actions, and assess outcomes, but translating these abilities into effective physical behavior requires decisions grounded in the robot’s observations, execution capabilities, and operating constraints. This survey investigates how recent robotic systems address this challenge through four functional aspects: perception and state representation; action interfaces and execution requirements; decision-making, planning, and coordination; and closed-loop feedback and capability improvement.
Across these aspects, we analyze how foundation models, classical robotic methods, and learned policies contribute to acquiring relevant information, choosing feasible actions, and adjusting execution in agentic robotic pipelines.
Within these pipelines, we further investigate how closed-loop feedback connects task execution with capability improvement, including changes to skills, programs, and models.
We review benchmarks and reported results to assess how existing evaluations measure agentic execution and capability improvement, accounting for task demands, operating conditions, and externally provided information and execution support. 
We conclude by outlining research directions for reliable agentic execution and capability improvement, emphasizing how to perform evaluation reliably across tasks in real-world operation.
\end{abstract}


\onecolumn
\makeatletter
\AddToHookNext{shipout/background}{\titlepanelbackground}
\@maketitle\thispagestyle{title}\label{FirstPage}
\noindent\begin{minipage}{\textwidth}\def\@captype{figure}\long\def\@makecaption#1#2{\par\vskip4pt{\raggedright\sffamily\fontsize{8.5}{10}\selectfont\textbf{#1.}~#2\par}}
\centering\footnotesize
\includegraphics[width=\textwidth]{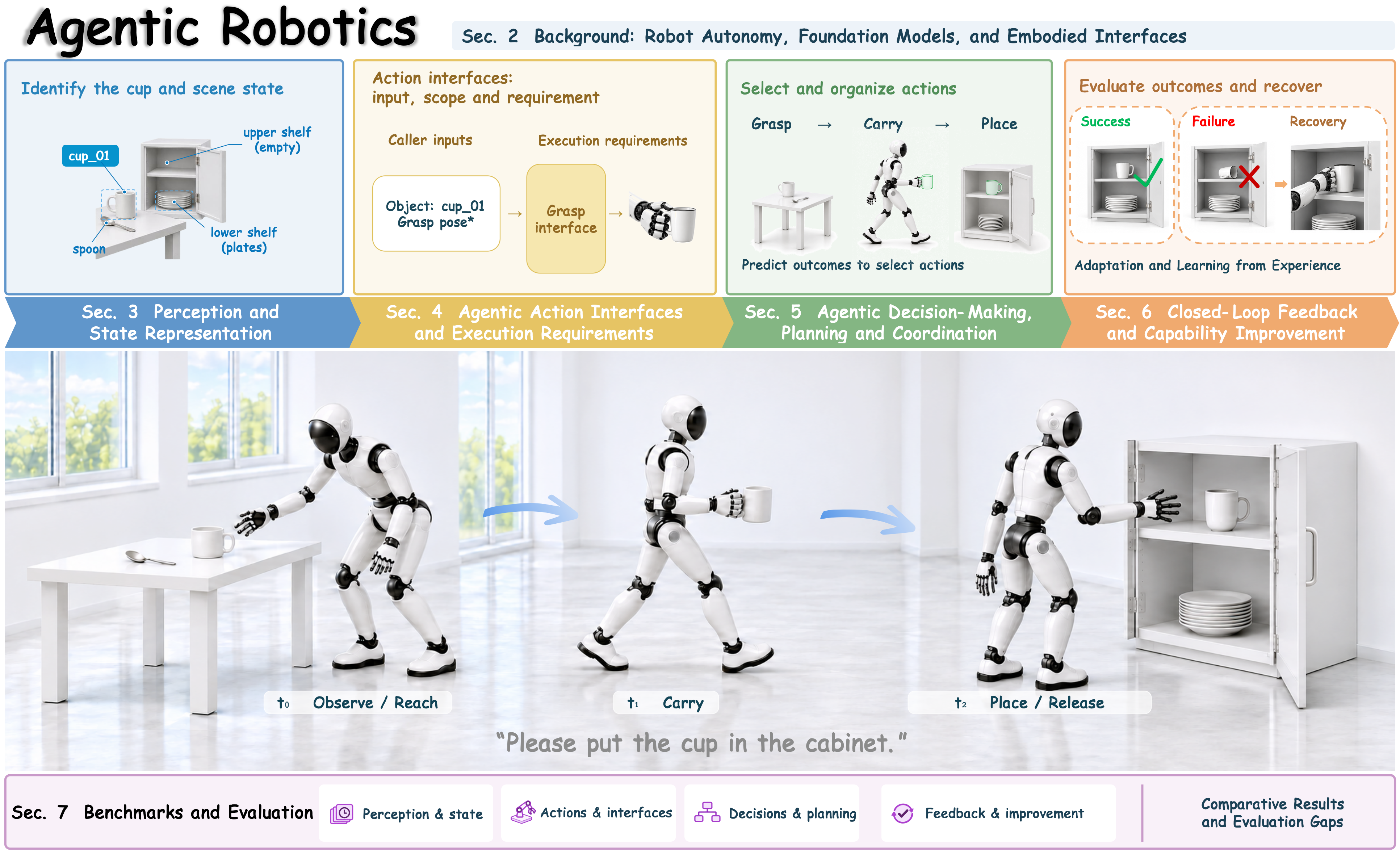}
\caption{Overview of the survey through an illustrative cup-placement task. The four colored panels connect perception and state representation (Section~\ref{sec:perception}), action interfaces and execution requirements (Section~\ref{sec:actionSpaces}), decision-making, planning and coordination (Section~\ref{sec:decision-making}), and closed-loop feedback and capability improvement (Section~\ref{sec:feedback}). Translucent cups depict predicted states used to inform action selection. Execution outcomes support recovery, while experience may support adaptation and learning whose gains and regressions require evaluation. These functions may share components; the section arrows indicate the organization of the survey rather than a fixed execution pipeline. The lower strip links these concerns to benchmarks, comparative results, and evaluation gaps (Section~\ref{sec:evaluation}). \textbf{*} Required inputs depend on the action interface: a grasp pose may be supplied by the caller or determined internally by the executor.}
\label{fig:survey-overview}
\end{minipage}
\makeatother
\twocolumn

\microtypesetup{activate=true}
\setlength{\dblfloatsep}{10pt plus 2pt minus 2pt}
\setlength{\dbltextfloatsep}{14pt plus 2pt minus 2pt}
\setlength{\textfloatsep}{14pt plus 2pt minus 2pt}
\section{Introduction}
\label{sec:introduction}

Robotic autonomy involves selecting and coordinating available capabilities toward a goal \citep{sutton1999between} and revising behavior in response to execution feedback \citep{astrom2008feedback}. Planning, monitoring, and feedback control have long addressed these issues: robot deliberation studies their integration \citep{ingrand2017deliberation}, while task and motion planning relates task dependencies to geometric feasibility \citep{garrett2021integrated}. Consider a robot instructed to pick up a cup and place it inside a cabinet (Figure~\ref{fig:survey-overview}). To achieve this goal, the robot must decide which capabilities to invoke and in what order, given the scene and its available skills \citep{ahn2022can}. It may need to open the cabinet before grasping the cup, while execution feedback may require revising its choices if the cup slips during transport \citep{huang2022inner,c7pub_REFLECT}. These decisions motivate the agentic perspective adopted in this survey: how robots autonomously select, coordinate, and adapt the use of their capabilities throughout task execution.

Recent advances in foundation models such as Astra~\citep{openai2026astraintro} and Claude Opus 5.5~\citep{anthropic2026opus55intro} have supported agents for multi-step tasks in computer use.
Extending these capabilities to robotics requires grounding decisions in the current physical state and executing them through available skills and controllers~\citep{zhang2026unexpectedrobotpolicyearly}. 
Within robotic systems, foundation models support task and scene interpretation, planning, and action generation. They can generate programs that invoke perception and control interfaces, as in Code as Policies~\citep{liang2023code}, or map observations and instructions directly to actions, as in RT-2~\citep{pmlr-v229-zitkovich23a}. 
These functions can also be combined within a model: the world action model (WAM) DreamZero~\citep{ye2026dreamzero} jointly models future video and actions, while Gemini Robotics 1.5~\citep{geminiroboticsteam2025gr15} incorporates reasoning within a VLA.
Predicted outcomes and execution feedback can guide action selection and revision, while interaction experience can support capability improvement. RoboClaw~\citep{li2026roboclawagenticframeworkscalable}, for example, coordinates robotic data collection and learning. From an agentic perspective, the key question is how systems autonomously coordinate these functions during task execution and capability development.

Existing surveys offer complementary perspectives on this literature. \citet{firoozi2025foundation} review foundation-model applications across perception, decision-making, and control. \citet{salimpour2025towards} classify integration approaches and agent roles, while \citet{raptis2025agentic} investigate agenticness and ethical implications in LLM-based robotic systems evaluated in physical settings. 
Our survey instead takes a system-level view of autonomy, rather than attributing it to an individual model or agent component. We analyze how perception, planning, execution, feedback, and learning jointly support autonomous behavior in robotic systems.

\begin{figure*}[!t]
\centering
\includegraphics[width=\textwidth]{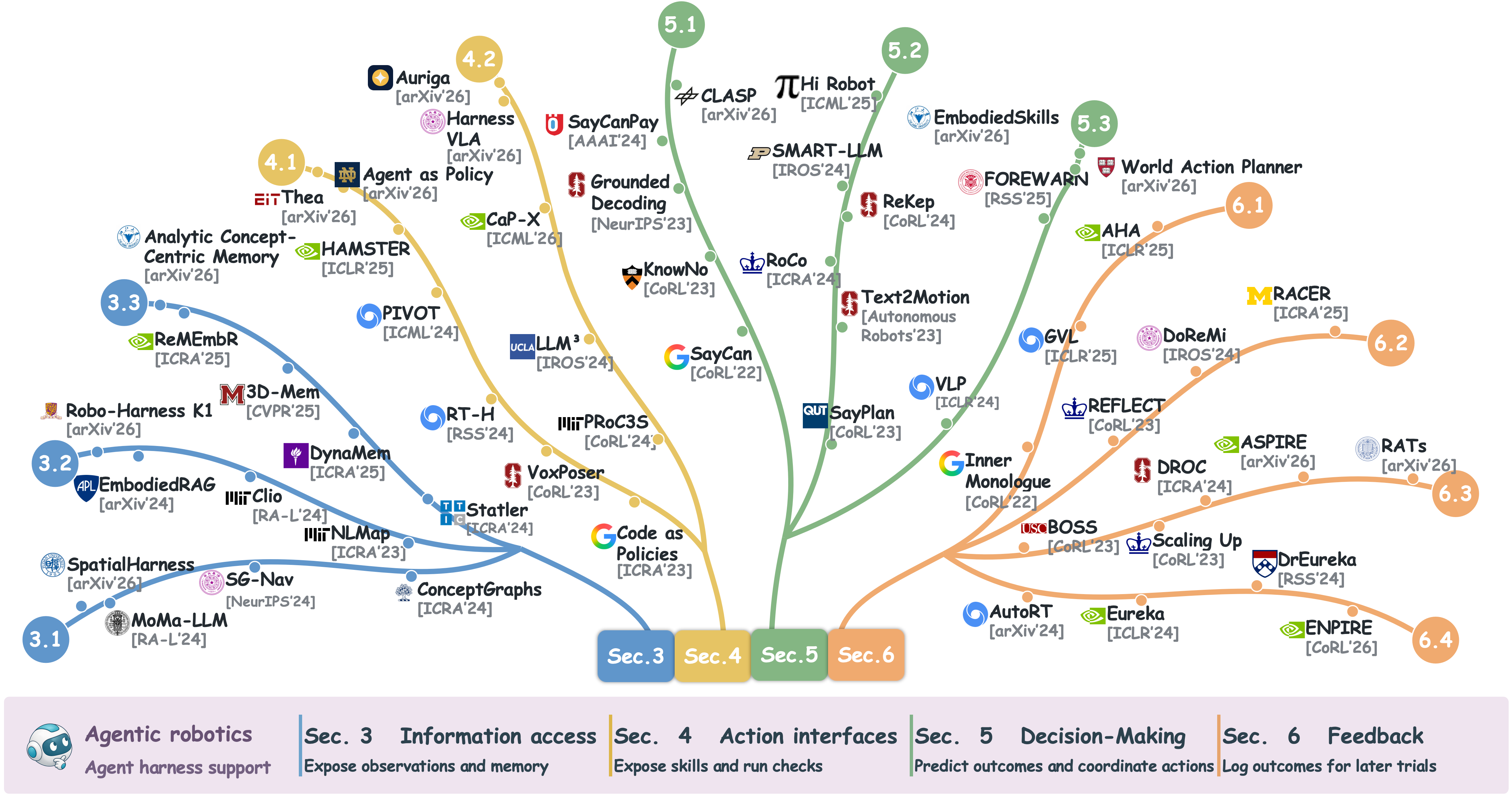}
\caption{Representative studies associated with the survey's functional discussions. The four main branches represent the sections below, with numbered branch tips identifying their subsection topics.
\textbf{Section~\ref{sec:perception}, perception and state representation:} scene representation and understanding (\ref{sec:scene-representations}); task-relevant information selection and access (\ref{sec:task-relevant-state}); dynamic state maintenance and environmental memory (\ref{sec:embodied-memory}).
\textbf{Section~\ref{sec:actionSpaces}, agentic action interfaces and execution requirements:} action specification and delegation (\ref{sec:action-specification-delegation}); execution requirements and feasibility checks (\ref{sec:action-grounding-feasibility}).
\textbf{Section~\ref{sec:decision-making}, agentic decision-making, planning and coordination:} action proposal and selection (\ref{sec:selection}); task planning and execution coordination (\ref{sec:hierarchy}); predictive models for action evaluation and planning (\ref{sec:prediction}).
\textbf{Section~\ref{sec:feedback}, closed-loop feedback and capability improvement:} semantic feedback and outcome evaluation (\ref{sec:feedback-evaluation}); closed-loop correction and recovery (\ref{sec:feedback-recovery}); adaptation and learning from experience (\ref{sec:feedback-learning}); learning harnesses and recursive self-improvement (\ref{sec:feedback-orchestration}).
Study placement highlights a representative functional association rather than an exclusive classification; the same work may be discussed in multiple subsections. The branches provide a reading map organized by function.
} 
\label{fig:survey-test}
\end{figure*}


In this survey, \textbf{agentic robotics} refers to the study and development of robotic systems that autonomously organize and adapt physical behavior toward specified goals. We focus on recent systems in which foundation models contribute to this process alongside classical robotic methods and learned policies, including autonomous learning and capability acquisition where present. Within an architecture, an \textbf{agent} may denote a decision-making component, such as a model, planner, or program; we assess autonomy at the level of the complete system, considering its operating conditions and provided goals and subgoals, information, and interventions. 
To investigate how these systems organize and adapt their behavior, we consider four connected functional aspects. Figure~\ref{fig:survey-overview} illustrates their relationships, while Figure~\ref{fig:survey-test} maps representative studies to each aspect. These aspects describe functions rather than separate architectural modules: one component may serve several functions, and several components may jointly support one function. The discussion below outlines what we examine within each aspect.

\textbf{Perception and state representation.} We examine how systems acquire, maintain, and retrieve information relevant to their goals and ongoing execution. Where systems choose what to observe or retrieve, we consider how these choices address missing information and changes in the task state.

\textbf{Agentic action interfaces and execution requirements.} We examine what an action request specifies, what inputs it requires, and which execution conditions can be checked or enforced. This clarifies what the agent can specify or adjust and what it delegates to planners, controllers, or learned policies.

\textbf{Decision-making, planning, and coordination.} We examine how systems select and order actions, account for dependencies and feasibility, and use predicted outcomes to guide decisions. We also consider when plans are revised and how execution is coordinated across components, robots, and people, including control handovers and completion checks.

\textbf{Closed-loop feedback and capability improvement.} We examine how feedback supports adjustments during execution and capability updates for later tasks. This includes how agents and learning harnesses organize practice, verify proposed changes, and determine which updates to retain. We also discuss recursive self-improvement (RSI), distinguishing better task performance from an improved ability to generate or evaluate further updates.

The survey makes three contributions:
\begin{itemize}
\setlength{\itemsep}{2pt}
\item We organize recent methods around four connected functional aspects to compare how different agentic robotic architectures support autonomous behavior.
\item We analyze how state information, action interfaces, and feedback connect decisions to physical execution, supporting both task-level adaptation and capability improvement across tasks.
\item We review benchmarks and reported results to assess the evidence for autonomy and capability improvement, accounting for each study's task demands, system configuration, and external support.
\end{itemize}



The remainder of the survey introduces the background, develops the analysis across the four functional aspects, reviews benchmarks and reported results, and discusses open challenges and future directions.

\input{sections/sec2_background}

\input{sections/sec3_perception}

\input{sections/sec4_actions}

\input{sections/sec5_decision}

\FloatBarrier
\input{sections/sec6_feedback}

\input{sections/sec7_evaluation}


\section{Challenges and Future Directions}
\label{sec:discussion}

\begin{figure*}[!t]
\centering
\includegraphics[width=\textwidth]{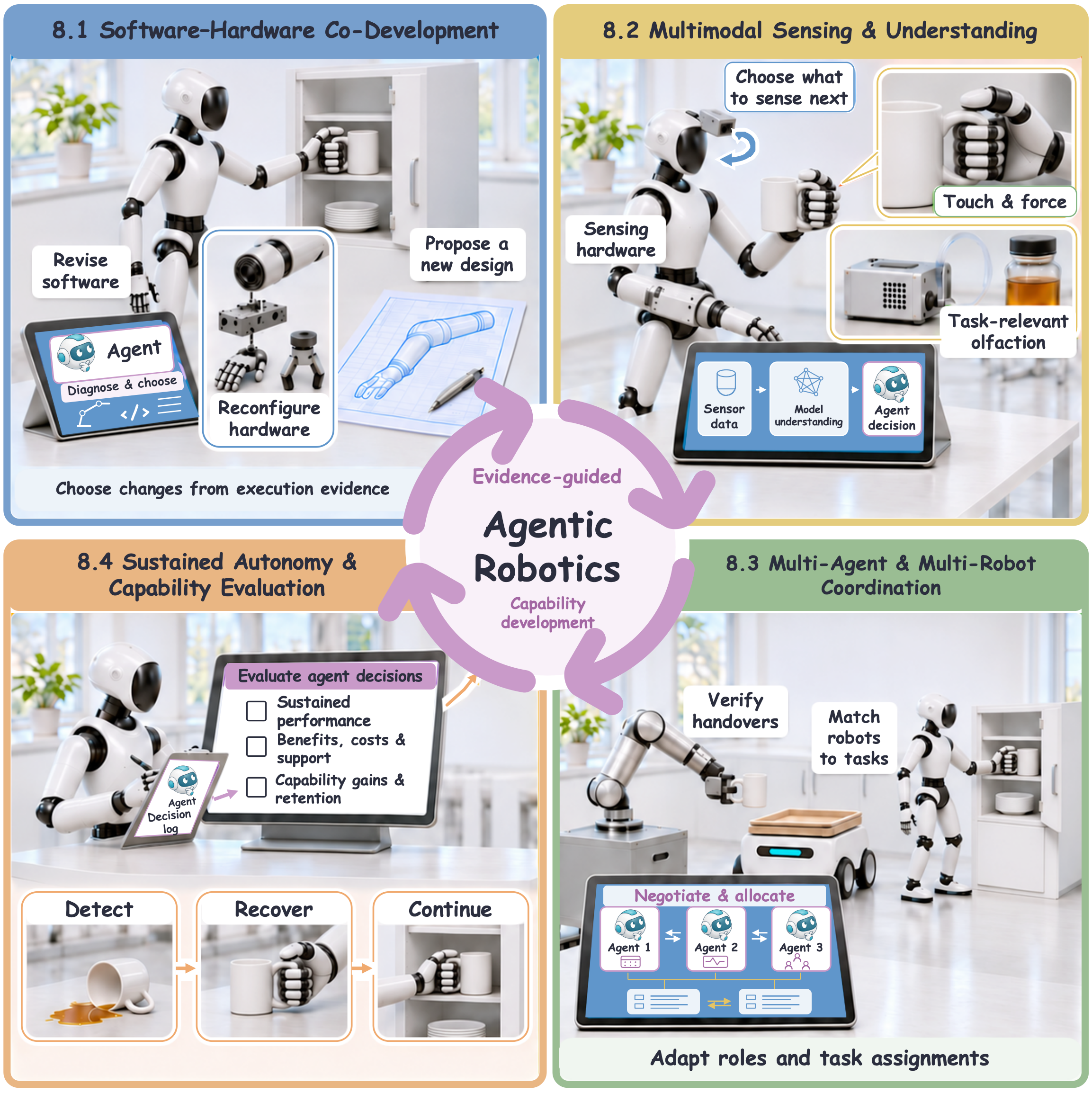}
\caption{Research priorities and evaluation for agentic robotics. Software--hardware co-development (Section~\ref{sec:discussion-improvement}) uses execution evidence to choose among software revision, hardware reconfiguration, and robot redesign. Multimodal sensing and understanding (Section~\ref{sec:discussion-multimodal}) connects task-relevant sensing hardware and data with model understanding, enabling agents to select further observations and guide physical action. Multi-agent and multi-robot coordination (Section~\ref{sec:discussion-coordination}) adapts agent roles, robot--task assignments, and task handovers across shared or separate goals, accounting for capability differences among heterogeneous robots. Evaluation (Section~\ref{sec:discussion-evaluation}) examines whether these decisions improve sustained performance and produce capability gains that are retained, while accounting for costs and external support. The cup-handling scenes connect these priorities to the example in Figure~\ref{fig:survey-overview}. The central loop represents evidence-guided capability development across the three research directions.}
\label{fig:future-directions}
\end{figure*}

Sections~\ref{sec:perception}--\ref{sec:feedback} review how robotic systems perceive, act, make decisions, and improve through feedback, while Section~\ref{sec:evaluation} examines the benchmarks and evidence used to assess these capabilities. Building on this synthesis, we propose three research priorities for agentic robotics: choosing software and hardware changes, connecting physical sensing with model understanding, and coordinating agents and robots across tasks. Figure~\ref{fig:future-directions} summarizes the three research priorities in Sections~\ref{sec:discussion-improvement}--\ref{sec:discussion-coordination} and the criteria for evaluating them in Section~\ref{sec:discussion-evaluation}. These criteria address sustained performance, costs and external support, and capability gains and retention.

\subsection{Agent-Directed Software--Hardware Co-Development}
\label{sec:discussion-improvement}
\label{sec:discussion-capabilities}

Software improvement can address many task failures, but its scope is bounded by the robot’s physical design. The body is therefore also a candidate for improvement when diagnostic evidence points to a physical limitation. A failed grasp, for example, may result from an inaccurate pose estimate, an infeasible approach motion, a control error, or a physical limitation of the gripper. Distinguishing among these causes requires targeted diagnostic trials, whose outcomes inform the choice among software revision, available hardware reconfiguration, and redesign. This choice connects the software changes reviewed in Section~\ref{sec:feedback} with the physical constraints on execution discussed in Section~\ref{sec:actionSpaces}.

A single diagnostic signal can admit multiple explanations. Slip during a short lift, for example, may reflect insufficient contact force, an unsuitable grasp configuration, or a control error. Testing a candidate intervention can help assess these explanations while evaluating a possible remedy. If an increase in commanded grasping force reduces slip, it provides evidence that a control adjustment can improve the grasp under the tested conditions while the comparison should also record whether hardware constraints remain a plausible explanation. Extending the learning harnesses in Section~\ref{sec:feedback-orchestration} to these trials would involve organizing diagnostic probes and intervention tests and managing any required resets. The resulting evidence would support comparisons among interventions based on execution outcomes, resource use, and required assistance.

Adapting to a changed body and choosing to change it are distinct capabilities. After an externally imposed tool replacement, the agent needs to check which models, controllers, and skill conditions remain valid and adapt those affected by the change. For autonomous reconfiguration, the agent must select a feasible physical change, such as a different tool or sensor placement, to address a diagnosed limitation. It must then arrange the change. In either case, changes in reach, loading, or collision geometry can invalidate the assumptions underlying existing skills, even when software interfaces remain unchanged. The agent must identify and revise the affected components and action descriptions, then verify execution with the modified body.

Design search offers methods for proposing new robot bodies, but implementing and validating those designs requires further work. RoboGrammar~\citep{sec8zhao2020robogrammar} explores morphology alongside control; RoboMoRe~\citep{sec8fang2025robomore} jointly optimizes morphology and reward; and Auto-Robotist~\citep{sec8wang2026autorobotist} retains reusable design knowledge from search. The latter two evaluate their design mechanisms in simulation. Bringing such methods into physical operation requires translating a diagnosed limitation into a design change, arranging fabrication and integration, and testing whether the resulting robot overcomes that limitation. The agent must compare redesign with software revision and available hardware reconfiguration in terms of expected benefits and implementation requirements. This comparison should account for manufacturing costs, integration and software adaptation, the risk of invalidating existing skills, and any required human assistance or external facilities.

\subsection{Multimodal Sensing and Understanding for Physical Action}
\label{sec:discussion-multimodal}

Camera images reveal only part of the physical conditions that determine successful action. Seeing a cup between the gripper fingers, for example, indicates its position, but grasp stability also depends on contact force and loading. Tactile, force, and proprioceptive signals provide complementary evidence about contact, loading, and robot configuration. Adding these sensors also raises a model-level question: how should their signals be represented and interpreted to assess the grasp and guide the next action? Developing these capabilities requires sensing hardware together with models and interfaces that make the resulting signals usable for decision-making.

The task determines which signals are useful and how they can inform the agent's decisions. For odor-source localization, olfaction provides chemical evidence that complements visual observations. \citet{hassan2024integrating} encode chemical-sensor readings as numerical prompt text alongside camera images to select odor-source navigation behaviors on a physical mobile robot. This example illustrates how an additional sensing modality can inform agent decisions through a text interface.

Beyond making sensor readings accessible to an agent, models need representations that connect physical signals with other modalities and robot actions. UniTouch~\citep{sec8yang2024unitouch} connects tactile representations with vision, language, and sound, while FuSe~\citep{sec8jones2025fuse} uses language grounding to incorporate heterogeneous sensors into generalist robot policies. These representations must also support task-level reasoning. For the information state in Section~\ref{sec:perception}, the agent needs to establish what judgment a reading supports. Recording its acquisition time helps assess whether it remains relevant to the intended action. When visual and contact evidence conflict, the agent must account for the unresolved state when deciding whether to revise its estimate, gather further evidence, or change the planned action.

Active sensing concerns which interaction will provide the evidence needed for the next decision. Building on active perception~\citep{sec8bajcsy2016active}, an agent can change its viewpoint, touch an object, or perform a limited physical probe. For olfactory tasks, it also needs to choose where and when to collect chemical measurements. Further contact evidence may help assess a grasp, whereas moving the camera may resolve an occlusion without disturbing the object. Selecting a sensing action requires weighing the information it is expected to provide against acquisition time and physical disturbance. Its outcome then informs whether further sensing is warranted or the available evidence is sufficient to act.

To guide an ongoing action, physical evidence must reach the relevant decision process while a response is still possible. Rapid contact changes may require local control responses, while task-level reasoning needs events such as contact loss to decide whether to continue, interrupt, or revise a plan. T-WLA \citep{zhou2026twla} uses incoming touch and cached visual context to correct the unexecuted portion of an action chunk between visual updates. Interfaces that translate sensor streams into these events need to preserve uncertainty and timing, allowing the system to judge whether the evidence still describes the current execution state. For the predictive models discussed in Section~\ref{sec:prediction}, the question is whether anticipating the consequences of sensing actions improves decisions within the available response time.

\subsection{Adaptive Multi-Agent and Multi-Robot Coordination}
\label{sec:discussion-coordination}

Agreement among reasoning agents is necessary for joint execution, which additionally requires executable assignments that respect physical workspace constraints. Changes in task requirements, robot capabilities, or execution conditions can invalidate existing roles and assignments. Software agents may collaborate on one task, divide interdependent tasks, or pursue independent goals in parallel. One agent may coordinate several robots, while several agents may support one robot; agent and robot counts therefore describe different aspects of a system. Building on Section~\ref{sec:hierarchy}, adaptive coordination concerns how these participants divide responsibility and revise their organization in response to execution evidence.

Role specialization and task allocation need to account for who can authorize physical actions and which resources those actions require. MALMM~\citep{singh2024malmm} separates planning, control-code generation, and supervision for manipulation. Such specialization requires a shared distinction between proposed plans, issued commands, and observed outcomes, with explicit responsibility for committing physical actions. Agents pursuing separate goals may also compete for the same robot, tool, or workspace. SMART-LLM~\citep{kannan2024smartllm} studies task decomposition, coalition formation, and allocation for robot teams. As goals or resource availability change, coalition membership and task assignments need to be reassessed against task dependencies and priorities. Reorganization can address these changes, but its benefits must justify the communication, synchronization, and arbitration it requires.

For multiple robots, task allocation must lead to physically compatible actions. RoCo~\citep{mandi2024roco} combines dialogue-based negotiation with motion-planning feedback and centralized multi-arm planning. Joint execution also requires attention to contact states and shared workspace constraints. At a task handover, the next robot needs evidence that the preceding action has established the conditions on which its own action depends. A delayed completion report or a conflict with local observations leaves those conditions uncertain. The system must account for that uncertainty when deciding whether to proceed, seek confirmation, or revise the joint plan.

Heterogeneous teams require assignments grounded in what each robot can currently observe and execute. EMOS~\citep{chen2025emos} derives capability descriptions from robot models and kinematic tools to guide planning and assignment in simulated heterogeneous teams. For physical deployment, these descriptions need to remain consistent with current robot states. The sensor, tool, loading, and body changes discussed in Sections~\ref{sec:discussion-improvement} and~\ref{sec:discussion-multimodal} can alter which tasks and handovers are feasible. Updating capability descriptions allows the team to reassess current assignments and determine whether roles or handover arrangements need to change.

\subsection{Evaluating Sustained Autonomy and Capability Development}
\label{sec:discussion-evaluation}

Evaluating sustained autonomy requires following a system from failure detection through recovery to subsequent task execution. LIBERO-RECOVER~\citep{c7recover} and RoboRecover~\citep{c8roborecover}, reviewed in Section~\ref{sec:evaluation}, assess continuation from supplied failure or deviation states. Extending these protocols to include autonomous failure detection during repeated task sequences would allow evaluation of when the system intervenes and whether its response restores progress. Detection errors, unnecessary interventions, and recovery outcomes should be related to task completion over the sequence. Reports should record resets, resource use, and human assistance, including help requested by the agent, and identify which interventions the system selects and completes autonomously.

The directions in Sections~\ref{sec:discussion-improvement}--\ref{sec:discussion-coordination} require comparisons that separate the effects of agent decisions from changes in available resources. For software--hardware development, software revision, available reconfiguration, and redesign should address the same limitation from comparable starting systems. Their costs include physical trials, hardware expenditure, software adaptation, and fabrication or integration assistance. For sensing, comparisons should distinguish adding a modality from improving the use of existing signals. Interpretation methods can be tested with the same modalities, and active sensing against fixed observation schedules with the same sensors and executors. Comparisons on the same robot team could test when changes in robot availability or task dependencies make reassignment worth its communication and handover costs. Across these comparisons, reports should account for computation, sensing time and disturbance, or communication and handover costs as appropriate. A specified operating horizon is needed to assess whether subsequent task gains justify the intervention costs.

Capability development should be tested on tasks or instances held apart from those used to select an update, while also checking previously acquired skills. Testing on held-out tasks assesses whether an update's benefits extend beyond the cases that guided its selection; retesting earlier skills measures retention and regression. After a body change, evaluations should distinguish recovery of previous performance from an expansion of the tasks the robot can perform. Design proposals, simulation results, and tested physical modifications support claims at different stages of development. Reports should identify the changes selected and completed by the system, the models or controllers supplied externally, and the fabrication or integration assistance required.

Evaluating the ability to produce further improvements requires testing the update mechanism itself. HyperAgents~\citep{c7hyperagents}, discussed in Section~\ref{sec:feedback-orchestration}, compares initial and evolved meta agents by their ability to produce better task agents within a fixed number of attempts; its robotics experiments concern reward design for a simulated Go2 robot. For sustained physical operation, original and revised mechanisms should start from comparable capabilities and receive matched trial and revision budgets. Subsequent gains on tasks held apart from mechanism selection would then provide evidence about their ability to generate useful updates. When agents can modify rewards or evaluators, final assessment should use a task criterion specified in advance and kept outside the system's permitted modifications.

\section{Conclusion}
\label{sec:conclusion}

This survey examines agentic robotics at the level of the complete robotic system. We organize recent work around four connected functions: perception and state representation; action interfaces and execution requirements; decision-making, planning, and coordination; and closed-loop feedback and capability improvement. This perspective connects the contributions of foundation models, classical robotic methods, and learned policies to the decisions a system makes and the physical behavior it can execute. We use benchmarks and reported results to assess these capabilities under their task demands, operating conditions, and external support.

Autonomous execution requires decisions grounded in current physical evidence and the robot's available capabilities. Recovery within a task, capability gains retained across tasks, and an improved ability to produce further gains require distinct evidence. The reviewed comparisons show that changes to system organization and execution support can improve performance, leave it unchanged, or reduce it, depending on the task and configuration. Comparisons that change several mechanisms together do not isolate their individual contributions. The numerical evidence synthesized here mainly concerns execution and organization in simulation.

Future research should advance agent-directed software--hardware co-development, connect physical sensing with model understanding, and enable adaptive multi-agent and multi-robot coordination. These directions require choices grounded in execution evidence: what to revise, which observations to acquire, and how to assign work as tasks and capabilities change. Progress should be assessed through sustained physical performance and capability gains retained on subsequent tasks, with explicit accounting for costs, external support, and losses in previously acquired capabilities.

\microtypesetup{activate=false}
\bibliographystyle{SageH}
\bibliography{references,refs_xuan,refs_chapter4,refs_chapter5,refs_chapter7,refs_chapter8,refs_intro_updates,refs_sec8_future}

\balance
\end{document}

%% file: sections/sec2_background.tex
\section{Background: Robot Autonomy, Foundation Models, and Embodied Interfaces}
\label{sec:background}

An agentic robot uses foundation models to decide what to do, but it still acts through its own skills, planners, and controllers. This section introduces the background for Sections~\ref{sec:perception}--\ref{sec:feedback}. Section~\ref{sec:autonomy-foundations} reviews classical concepts of robot autonomy, and Section~\ref{sec:fm-capabilities} describes what foundation models contribute to a robot and how their outputs reach its controllers.

\subsection{Foundations of Robot Autonomy}
\label{sec:autonomy-foundations}

An autonomous robot repeatedly observes its surroundings, decides what to do, and acts \citep{ingrand2017deliberation,kroemer2021review}. We review three classical ideas that later sections build on: acting from partial observations, planning and control, and skills. Table~\ref{tab:notation} lists the notation used across Sections~\ref{sec:background}--\ref{sec:feedback}.

\textbf{Partial observability, history, and belief.} Let $s_t$ denote the true task state and $o_t$ the sensor observation at time $t$. Under partial observability, $o_t$ reveals only part of $s_t$. The interaction history $h_t=(o_0,a_0,\dots,a_{t-1},o_t)$ records the observations and intervening actions available to the robot. A belief $b_t$ represents what the robot currently infers about the hidden state.


A partially observable Markov decision process (POMDP) formalizes this relationship \citep{kaelbling1998planning,lauri2023pomdp}. Because the true state $s_t$ is hidden, a policy selects an action $a_t$ from the interaction history $h_t$ and the goal $g$ as $a_t \sim \pi(\cdot\mid h_t,g)$. 
Formally, the belief can be denoted by:
\begin{align}
b_t(s) &= \Pr(s_t=s\mid h_t),\\
b_{t+1}(s') &\propto Z(o_{t+1}\mid s',a_t)
\sum_{s\in\mathcal{S}}T(s'\mid s,a_t)b_t(s),
\label{eq:belief}
\end{align}
where $\Pr$ denotes probability, $s$ and $s'$ are possible state values in the state space $\mathcal{S}$, $T$ is the transition model, and $Z$ is the observation model. The second line gives the belief update for a discrete state space. The belief is a sufficient statistic of the history for optimal decision-making in a POMDP \citep{kaelbling1998planning}.

Exact belief tracking is often intractable \citep{lauri2023pomdp}. Practical robotic systems therefore maintain an information state
$
m_t=u(m_{t-1},o_t,a_{t-1}),
$
where $u$ is the update function. The information state may be a map, scene graph, or memory buffer. Like a belief, $m_t$ summarizes information accumulated through interaction, but it need not be probabilistic or sufficient for optimal decision-making. Its usefulness depends on what $u$ retains. Section~\ref{sec:perception} discusses these representations. When a method derives a single current-state estimate from $b_t$ or $m_t$, we denote it by $\hat{s}_t$, distinguishing it from the hidden state $s_t$.



\textbf{Task planning, motion planning, and control.} Task planning selects and orders task steps, typically using operators with preconditions and effects \citep{fikes1971strips,ghallab2016acting}. Motion planning computes feasible robot motions for those steps \citep{lavalle2006planning}, while control executes the motions and responds to disturbances \citep{siciliano2016springer}. For placing a cup in a closed cabinet, opening the cabinet is a task-level choice, finding an arm path is a motion-planning problem, and regulating grip force is a control problem.

In practice, these levels are coupled. Task and motion planning (TAMP) checks symbolic choices for geometric feasibility, because a step that is valid symbolically may have no feasible motion \citep{garrett2021integrated}. Three-layer architectures run deliberation, sequencing, and feedback control as separate layers at the same time \citep{gat1998three}. A learned policy may cover one level or several \citep{bekris2024state}; an end-to-end policy, for example, takes a task-level instruction and directly outputs motor commands.

\textbf{Skills and options.} A skill is a reusable behavior that runs over many time steps, such as grasping a cup or opening a drawer. The options framework \citep{sutton1999between} formalizes a skill as an option $\omega=\langle\mathcal{I}_\omega,\pi_\omega,\beta_\omega\rangle$ consists of an initiation set $\mathcal{I}_\omega$, an option policy $\pi_\omega$, and a termination condition $\beta_\omega$. At the $k$th decision time $\tau_k$, a high-level rule $\mu$ selects an applicable option $\omega_k$ from the repertoire $\Omega$; its policy then produces primitive actions until the next decision time:
\begin{equation}
\begin{split}
\omega_k&\sim\mu(\cdot\mid h_{\tau_k},g),\qquad \omega_k\in\Omega,\\
a_t&\sim\pi_{\omega_k}(\cdot\mid h_t),\qquad \tau_k\le t<\tau_{k+1}.
\end{split}
\label{eq:two-level}
\end{equation}
Both levels condition on the history because the state is only partially observed; whether an option can start is checked with the current belief or state estimate.

The next decision may be triggered by option termination, an interrupting event, or expiry of a maximum decision interval:
\begin{equation}
\begin{split}
\tau_{k+1}=\min\bigl\{\,t>\tau_k:\ &\beta_{\omega_k}(h_{\tau_k:t})=1\\
&\lor\ e_t\in\mathcal{E}\ \lor\ t-\tau_k\ge\Delta\,\bigr\}.
\end{split}
\label{eq:trigger}
\end{equation}
Here, $h_{\tau_k:t}$ is the history since the option was invoked; $\beta_\omega$ is treated as a deterministic termination predicate; $e_t$ is an event detected at time $t$; $\mathcal{E}$ is the set of interrupting events; and $\Delta$ is the maximum interval between decisions.

The three triggers have different meanings. Termination ends the option by its own criterion, whether the option succeeded or failed. An event, such as a detected slip, interrupts the option from outside. When the interval expires, control returns to $\mu$ even if the option could continue; $\mu$ may select the same option again, which refreshes the decision, or switch to another one. None of these triggers changes the skills themselves; Section~\ref{sec:feedback} covers methods that update skills from experience.

\begin{table}[t]
\small\sf\centering
\setlength{\tabcolsep}{4pt}
\renewcommand{\arraystretch}{1.08}
\caption{Notation used across Sections~\ref{sec:background}--\ref{sec:feedback}. Specialized symbols are defined where they are used.
\label{tab:notation}}
\begin{tabularx}{\columnwidth}{@{}>{\raggedright\arraybackslash}p{0.30\columnwidth}>{\raggedright\arraybackslash}X@{}}
\toprule
Symbol & Meaning \\
\midrule
$t,k,j$ & Interaction-step index; high-level decision index; retained-update index \\
$g$, $\pi$ & Goal or instruction; policy \\
$s_t$, $o_t$, $a_t$ & True state, observation, primitive action at time $t$ \\
$h_t$ & Interaction history up to time $t$ \\
$b_t$, $m_t$, $\hat s_t$ & Belief; maintained information state; state estimate \\
$u$ & Information-state update \\
$q$, $c_t$ & Context selection or readout; context supplied for a decision \\
\addlinespace[2pt]
$\omega$, $\Omega$ & Option with initiation, policy, and termination $(\mathcal{I}_\omega,\pi_\omega,\beta_\omega)$; repertoire \\
$\mu$, $\tau_k$ & High-level rule; decision time \\
$\xi_k$ & Invocation parameters (continuous or discrete) \\
$e_t\in\mathcal{E}$, $\Delta$ & Interrupting event; maximum decision interval \\
$a_{t:t+H-1}$ & Action chunk of horizon $H$ \\
\addlinespace[2pt]
$\mathcal{V}$, $y_t$ & Execution assessment procedure; resulting assessment \\
\addlinespace[2pt]
$\Theta_j$ & Retained system state, such as skills, programs, model parameters, or learning-harness components \\
$U_j$, $\mathcal{D}_j$ & Retained-update procedure; experience and verification evidence for update $j$ \\
\bottomrule
\end{tabularx}
\end{table}


\subsection{Foundation Models and Robot Capabilities}
\label{sec:fm-capabilities}

Foundation models are large models trained on broad data that can be adapted to many tasks \citep{bommasani2021opportunities}. For a robot, they can describe scenes, write plans or programs, output actions, predict future observations, or judge task success \citep{yang2023foundation}. We describe what pretraining provides, how the main model families connect to a robot, and which roles they take in a robot system.


\textbf{Pretrained knowledge and task understanding.} Pretraining on language, images, and video gives a model general knowledge about objects, their typical uses and locations, and everyday activities. Such knowledge can support task interpretation, but it does not reveal the current scene or show what a particular robot can execute. The robot still needs current observations and embodiment-specific information to locate an object and assess reachability. An early evaluation of GPT-6 Astra as a manipulation policy illustrates this gap \citep{zhang2026unexpectedrobotpolicyearly}. Without robot-specific fine-tuning, Astra generated end-effector targets for a non-learned motion layer; it performed best on simulated tasks with high semantic demands and worst on tasks requiring precise or dynamic motion.

Transfer also depends on whether the pretraining data cover the target scenes and embodiments, and whether the sensor and action interfaces are compatible. Cross-embodiment data can support transfer between robots, as in Open X-Embodiment \citep{oneill2024open}, while broader scene coverage can improve performance on a single platform, as DROID shows \citep{khazatsky2024droid}. Adapting to new observation and action spaces remains a separate problem. Octo can be fine-tuned to new interfaces \citep{ghosh2024octo}, while Heterogeneous Pre-trained Transformers connect embodiment-specific input stems and task-specific output heads through a shared pretrained trunk \citep{wang2024hpt}. These evaluations use robots that appear in the training or fine-tuning data, so they do not test transfer to an entirely new robot.

\begin{table*}[t]
\small\sf\centering
\setlength{\tabcolsep}{4pt}
\renewcommand{\arraystretch}{1.15}
\caption{Model families and their interfaces to robotic systems. Inputs and outputs indicate typical uses; the families overlap. Examples illustrate the distinctions; API denotes application programming interface.\label{tab:model-roles}}
\begin{tabularx}{\textwidth}{@{}>{\raggedright\arraybackslash}p{2.3cm}>{\raggedright\arraybackslash}p{2.7cm}>{\raggedright\arraybackslash}p{2.8cm}>{\raggedright\arraybackslash}X>{\raggedright\arraybackslash}p{2.8cm}@{}}
\toprule
Family & Typical inputs & Typical outputs & Robot-interface considerations & Examples \\
\midrule
Language models & Instructions, context, skill or API descriptions & Subgoals, plans, programs, tool calls & Available capabilities and checks against the current scene & GPT-3 \citep{brown2020language} \\
Vision-language models & Images or video with text & Features, descriptions, spatial references, plans & Coordinate frames and the time of observation & CLIP \citep{radford2021clip}; PaLM-E \citep{pmlr-v202-driess23a} \\
Vision-language-action models & Observations, proprioception, instructions & Action tokens or continuous action chunks & Action representation, control rate, embodiment, and start/stop conditions & RT-2 \citep{pmlr-v229-zitkovich23a}; $\pi_0$ \citep{black2024pi0} \\
Video and predictive models & Past observations with text or action conditions & Future frames or latent states & Prediction horizon, action conditioning, and use in action selection & V-JEPA~2 \citep{assran2025v}; Cosmos \citep{agarwal2025cosmos} \\
Reward and evaluation models & Observations or videos with a task description & Success, progress, value, or reward estimates & Definition of success, calibration, and evaluation timing & GVL \citep{ma2025gvl}; RoboReward \citep{lee2026roboreward} \\
\bottomrule
\end{tabularx}
\end{table*}

\textbf{Model families and their interfaces.} Table~\ref{tab:model-roles} compares the outputs an agent can use and the interfaces needed to turn them into behavior. Language models and vision-language models (VLMs) produce text, programs, or spatial references for other components to ground and execute, whereas vision-language-action (VLA) models map observations and instructions directly to robot actions \citep{ma2026vlasurvey}. Widely used models of the first two families include the proprietary GPT series from OpenAI \citep{openai2026astraintro}, Claude from Anthropic \citep{anthropic2026opus55intro}, and Gemini from Google DeepMind \citep{google2026gemini4}, and the open-weight Llama from Meta \citep{grattafiori2024llama3}, Qwen from Alibaba \citep{yang2025qwen3}, DeepSeek \citep{deepseekai2026v4}, Kimi from Moonshot AI \citep{kimiteam2026k3}, and GLM from Zhipu AI \citep{glm5team2026glm5}. Predictive and evaluation models instead provide forecasts or success judgments that help the agent select and assess actions. For VLA models, the action representation determines how outputs are produced: action tokens \citep{pertsch2025fast} and continuous chunks \citep{black2024pi0} require different training and inference procedures (Section~\ref{sec:actionSpaces}). Their conditioning inputs, in turn, set how an agent can steer those outputs. For example, $\pi_{0.7}$ \citep{ai2026pi07} accepts subtask language, subgoal images, and episode metadata from a high-level policy, a human coach, or a world model.

\textbf{Roles in a robot system.} A model can take different roles in a robot system: describing the scene, producing a plan or program, proposing actions, selecting among candidates, predicting what an action would cause, or judging whether an executed action succeeded.

These roles cut across model families. A VLM can generate subgoals, as PaLM-E does \citep{pmlr-v202-driess23a}, or judge task success when prompted as a detector \citep{du2023vision}. The same evaluation role can instead be filled by a trained reward model such as RoboReward \citep{lee2026roboreward}. Likewise, scene information can come from a VLM description or from a specialized model such as SAM~2 \citep{ravi2025sam2}, which segments and tracks prompted objects.

An agent may therefore rely on one model for several roles or distribute them across components. Gemini Robotics 1.5 \citep{geminiroboticsteam2025gr15}, for example, assigns planning, tool calls, and success detection to an embodied-reasoning model that invokes a VLA through language instructions. In either arrangement, planners, policies, and controllers carry out the agent's decisions and report back.

\textbf{Connecting model outputs to robot execution.}\label{sec:interfaces}
Model outputs reach the robot through execution interfaces, which Section~\ref{sec:actionSpaces} discusses in detail. Robot software already provides such interfaces: a Robot Operating System (ROS) action \citep{macenski2022ros2} accepts a goal, reports feedback and a result, and can be canceled, and a behavior tree \citep{colledanchise2018behavior} reports whether each behavior is running, has succeeded, or has failed. Language models call software tools in a similar way \citep{schick2023toolformer,yao2023react}. The Model Context Protocol (MCP; \citealp{anthropic2024mcp}) standardizes how tools are listed and called, and ROS-MCP-Server \citep{varghese2025rosmcp} uses it to expose ROS topics and services to a language model. Once a behavior starts, control returns to the agent when the behavior terminates, an event interrupts it, or the maximum interval passes (Equation~\eqref{eq:trigger}). Upon control return, the agent reads the reported status before choosing its next step (Sections~\ref{sec:decision-making} and~\ref{sec:feedback}).

%% file: sections/sec3_perception.tex
\section{Perception and State Representation}
\label{sec:perception}

\begin{figure*}[!t]
\centering
\includegraphics[width=\textwidth]{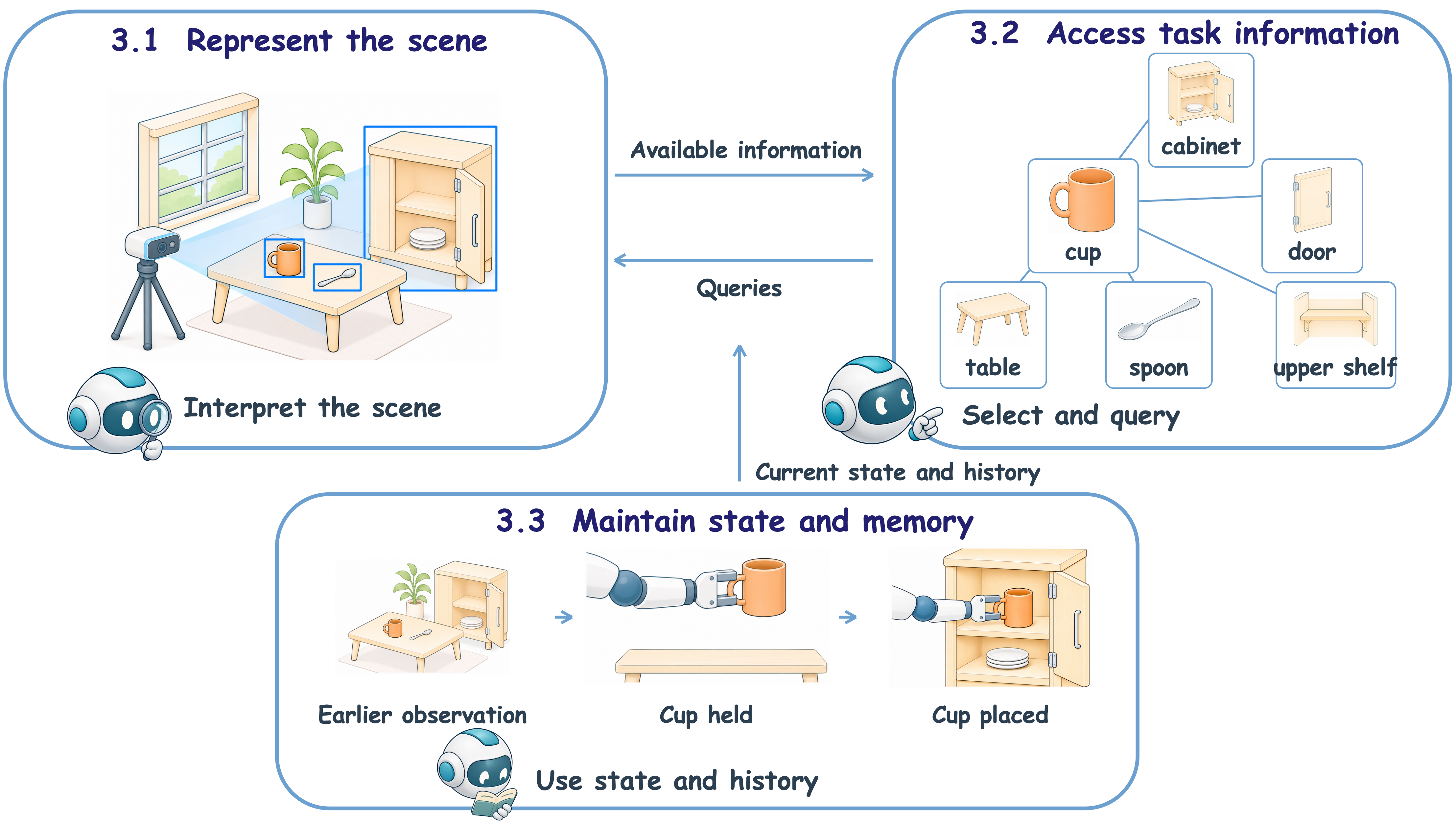}
\caption{Overview of perception and state representation. Scene representation (Section~\ref{sec:scene-representations}) makes environmental information available through observations, maps, and object-centered representations. Task-relevant information selection and access (Section~\ref{sec:task-relevant-state}) provide the information needed for a decision. State and memory maintenance (Section~\ref{sec:embodied-memory}) retains history and updates the current state as interaction proceeds. The cup-placement example illustrates these complementary functions; arrows indicate information exchange rather than a fixed processing pipeline.}
\label{fig:perception-logic}
\end{figure*}

This section describes how perception provides usable state information for an embodied agent. Figure~\ref{fig:perception-logic} outlines three functions: scene representation defines how environmental properties are encoded (Section~\ref{sec:scene-representations}); task-relevant selection and access determine which information reaches the agent and how missing details are obtained (Section~\ref{sec:task-relevant-state}); and state maintenance and environmental memory retain earlier evidence and update state estimates as the scene changes (Section~\ref{sec:embodied-memory}).

The information-state formulation in Section~\ref{sec:autonomy-foundations} describes how a robot maintains information through interaction:
\begin{equation}
m_t=u(m_{t-1},o_t,a_{t-1}).
\label{eq:perception-state-update}
\end{equation}
Here, $t$ indexes interaction steps, $o_t$ is the current observation, and $a_{t-1}$ is the preceding primitive action. The representations in Section~\ref{sec:scene-representations} determine what $m_t$ encodes, while $u$ combines these inputs with the previous information state $m_{t-1}$ to retain records and revise current estimates (Section~\ref{sec:embodied-memory}).

From this maintained information, a system constructs context for the current task:
\begin{equation}
c_t=q(m_t,g).
\label{eq:perception-context}
\end{equation}
Given the goal $g$, $q$ summarizes the selection and readout mechanisms in Section~\ref{sec:task-relevant-state}: it selects, retrieves, or organizes information from $m_t$ into the context $c_t$ available to a decision component. This context may contain images, structured records, or text. Task relevance may also shape representation construction. Queries may expose stored details without acquiring new observations, whereas newly acquired evidence can update $m_t$ through Equation~\eqref{eq:perception-state-update}.

\subsection{Scene Representation and Understanding}
\label{sec:scene-representations}

This section compares three ways to represent an environment: visual observations, metric and semantic maps, and explicit descriptions of objects and their relationships. We focus on what information each representation provides directly and what the agent must still infer.



\textbf{Visual context: interpreting camera observations.}
Camera observations provide visual information about the scene. Some systems also supply depth measurements. These inputs can help a model plan task steps or generate robot actions. PaLM-E uses sensor observations to generate language-based plans and skill instructions \citep{pmlr-v202-driess23a}, while RT-2 predicts robot action tokens from visual observations and language instructions \citep{pmlr-v229-zitkovich23a}. These examples illustrate how observations can support language-based task planning and condition robot action prediction. Both uses may coexist within a single system.

In systems that delegate execution to a policy or primitive library, observations can guide the model's choice of what to request next. Hi Robot uses a high-level vision-language model (VLM) to interpret multiple camera views together with user instructions and feedback, producing intermediate language commands for a low-level $\pi_0$ vision-language-action (VLA) policy. The VLA policy uses these commands, images, and robot state to generate actions \citep{shi2025hi}. Harness VLA uses an LLM agent to select and parameterize calls to analytic primitives and a frozen VLA primitive based on current color and depth (RGB-D) observations and robot state  \citep{zhang2026harnessvla}. The comparison therefore concerns which components can access observations, how those observations shape language commands or primitive calls, and how updated observations inform subsequent decisions.

Images retain appearance and local scene details, but the model must identify the information needed for its next decision. An image of a partly occluded cup does not directly provide its position in the robot's coordinate frame. If the next action requires a target position, the system must connect the object identified in the image or instruction to a location in physical space. SpatialHarness renders virtual views from a simulated scene kept in sync with the physical robot, giving a frozen policy a clearer view of task-critical alignments than the physical cameras alone provide \citep{wang2026spatialharness}.

\textbf{Spatial maps: locating objects in the environment.}
Metric and semantic maps associate descriptions of targets with locations in physical space. VLMaps links visual-language features to reconstructed space \citep{VLMaps}, while ConceptFusion supports open-set, multimodal queries over a three-dimensional map \citep{Conceptfusion}. These mechanisms can locate a language-described target in previously mapped areas, even when it is outside the current camera view. For the agent, the question is what map information it receives and how it can obtain further details.

MoMa-LLM connects its spatial representation to LLM planning through a structured textual encoding. It combines RGB-D observations with known camera poses and semantic labels to construct a dynamic 3D map and a hierarchical scene graph of rooms and objects. The LLM receives a compact description of the room–object structure, object states, distance categories, and unexplored areas, alongside recent actions and execution outcomes. It uses these inputs to select navigation, exploration, and manipulation commands, which are executed using the underlying spatial representation. As the robot explores or interacts with objects, the scene representation and action history are updated and re-encoded for subsequent LLM decisions \citep{MoMaLLM}.

The usefulness of these updated descriptions depends on the accuracy of the underlying scene representation. MoMa-LLM generally assumes ground-truth inputs for semantic masks, depth, localization, and handle detection in its reasoning-focused setup \citep{MoMaLLM}. The effectiveness of its textual encoding for planning and the system's robustness to errors in these inputs are therefore distinct evaluation questions.

\textbf{Object records: tracking identities and relations.} Object records identify individual objects and describe their relations. Room membership helps the agent decide where to search. Containment and holding indicate whether an object is inside a container or held by the robot. 

These relations help the agent decide what to do next, such as opening a container before grasping an object or placing an object that the robot is already holding. SG-Nav~\citep{sg-nav} uses an online object--group--room graph for language-model reasoning about where to search. Thea~\citep{wang2026thea} maintains persistent object references, coarse positions, container information, and relations such as containment and holding for subsequent queries and interaction. Hydra \citep{hydra}, ConceptGraphs \citep{Conceptgraphs}, and HOV-SG \citep{HOV-SG} provide related mechanisms for organizing environmental content through scene graphs.

These records preserve distinctions that a location alone does not express. A stable reference allows later queries to refer to the same cup. A holding relation records whether the robot holds it, while a separate location or support relation can indicate that it is on a shelf. These records must link observations to the correct object and contain enough detail for the task. A mistaken match can direct a later command to the wrong object, while a coarse record may omit geometry needed for manipulation. Across these forms, the comparison concerns what information is directly available and what the model must still infer. The next subsection considers which parts of this available information enter a particular task context.

\subsection{Task-Relevant Information Selection and Access}
\label{sec:task-relevant-state}
\label{sec:observation-interfaces}


Providing the full environmental representation can burden each reasoning step, but selecting only the named target can omit necessary information. Placing the cup may require information about the cabinet door and shelf even when the instruction names neither. We examine how observations and scene information are provided as the initial context, how task relevance guides information selection during representation construction and context preparation, and how queries and perception tools supply missing details. These functions can overlap: information may be selected before it is supplied or as the agent accesses a stored representation.

\textbf{Providing observations and scene information as the initial context.}
At each decision step, the system supplies context to the agent. This may include camera images, a structured description of the scene, or both. Here, initial context means the information supplied before any additional queries at that step. 

Hi Robot supplies multiple camera views to a high-level vision-language model \citep{shi2025hi}. Images retain appearance cues that structured records may omit, but the model must identify the objects and scene properties relevant to its next decision. Thea supplies current observations together with a compact world-state brief containing object references, coarse positions, confidence, and freshness, indicating how recently the information was observed or updated \citep{wang2026thea}. These fields make selected information directly available to the agent. The brief reduces the need to infer every property from images, but its records must refer to the correct objects and be updated as the scene changes. Details omitted from the brief can remain available through queries. This distinction between the supplied context and the underlying representation raises the question of which information to retain and which to include for a particular task.

\textbf{Selecting task-relevant information during representation construction and context preparation.}
Selection determines which information is retained or included in the agent's context. It can occur while constructing a representation or when preparing information from an existing representation for a task.

Clio uses a specified task set to select semantic concepts and spatial granularity during scene-graph construction \citep{Clio}. This focuses the representation on those tasks, but may omit details needed by later tasks. SCRUB, introduced in Taskography, prunes an existing graph for a modeled planning problem while preserving the required objects and dependencies \citep{agia2022taskography}. It reduces the information considered during planning. Clio shapes what is stored, while SCRUB selects from an existing graph. Both illustrate selection that can occur before information reaches the agent.

What is omitted matters as much as what is retained. Details excluded from the current context may remain available in the underlying map. Information never recorded there cannot be recovered simply by querying that map. The adequacy of the selected information also depends on the task. SCRUB's guarantees apply to its modeled planning domains; they do not establish that the selected graph contains geometry needed for manipulation outside those models \citep{agia2022taskography}.

\textbf{Obtaining missing information through queries and perception tools.}
When the current context is insufficient, the agent can retrieve stored details or call perception tools. Retrieval returns existing records, while perception tools derive information from observations.

Thea supports queries that use object references to retrieve object information and images \citep{wang2026thea}. Keeping these details outside the default context reduces its size, but obtaining them requires additional query steps. CaP-X exposes segmentation, visual pointing, and geometric processing to generated programs \citep{c5capx}. Instruct2Act \citep{c5instruct2act} and OK-Robot \citep{liu2024okrobot} also use modular perception in robot task execution. These calls return different information. A mask identifies an image region; a pose estimates position and orientation; a relation states, for example, whether one object is inside another. In Robo-Harness K1, the agent can request a depth measurement or a grasp candidate, use the result to choose a motion, and measure again after acting \citep{li2026roboharnessk1}.

The agent needs an output that matches the next action's requirements. An object mask alone does not determine a pose in the robot's coordinate frame, and a containment relation does not establish that a shelf has enough free space for placement. SayPlan lets the reasoning model expand relevant regions of a collapsed hierarchical scene graph \citep{rana2023sayplan}. Expansion reveals stored details that were hidden from the current view.

For example, a room can appear as a single node in an expandable tree view. Expanding it reveals its contents. This changes what the agent can inspect without adding new observations to the underlying representation. Query interfaces determine what information the agent can obtain, but the agent must still decide when to use them. EmbodiedRAG's feedback configuration supports further queries when planning reveals missing information \citep{EmbodiedRAGSceneGraph}.

Further queries help only if the agent recognizes what is missing. Knowing the cup's location may support a grasp request, while placement may also require the cabinet door state and available shelf space. The agent must therefore balance the cost of further queries against the risk of acting without necessary information. The required detail also depends on the execution interface.

A grasping skill may accept an object reference, while a motion command may require a target pose. Section~\ref{sec:actionSpaces} examines these interfaces. The next subsection examines how stored information is maintained as the scene changes.

\subsection{Dynamic State Maintenance and Environmental Memory}
\label{sec:embodied-memory}

When the cup leaves view, an earlier observation can help the agent decide where to search. That observation records where the cup was seen; it does not establish where the cup is now. We distinguish historical records from current state estimates. Historical records describe past observations and interactions, while current estimates describe what the system believes about the scene now. We examine how these records are constructed, how they are updated as new evidence arrives, and how they support planning and action.

\textbf{Constructing memory from observations and interactions.}
Memory construction selects what to retain and how to record it. A system may summarize observations in text, keep selected images, or link observations to object records. ReMEmbR stores video descriptions with observation locations and timestamps, supporting retrieval of what was observed, where, and when without supplying the full video history to the reasoning model \citep{ReMEmbR}. 3D-Mem retains selected multi-view snapshots of objects and their surroundings, preserving visual context for later spatial reasoning and exploration \citep{ThreeDMem}. RoboMemory combines summaries of previous interactions with an incrementally updated spatial knowledge graph, allowing its planner to use both task history and object relations \citep{RoboMemory}.

These choices involve different trade-offs. Descriptions can omit details or contain errors from the captioning process. Snapshots let the model inspect earlier images, but show only selected views. The model must still interpret what those images contain. Object records must link observations to the correct objects and keep their relations consistent as the scene changes. Repeated summarization can also remove details from earlier interactions. 

\textbf{Updating records and current state estimates.}
New evidence may add to the history or change the current state estimate. Seeing the cup in a new location can revise its estimated position without making the earlier observation an incorrect historical record. Systems therefore need rules for deciding how new evidence changes the current estimate and which historical records should be added, retained, or revised. Statler uses a world-state writer to record inferred changes, including changes to properties that are not directly observable. These updates support later reasoning, but incorrect updates can mislead subsequent decisions \citep{Statler}. DynaMem adds new spatial evidence and removes stored voxels when current depth observations contradict them. Voxels are the small volume elements used to represent a 3D map. This updates observed regions, but does not verify changes elsewhere \citep{DynaMem}. MoMa-LLM updates its spatial representation as exploration reveals new information. When room assignments change, it also adjusts room references in the action history to match the updated representation \citep{MoMaLLM}. Thea combines perceptual updates with action effects recorded when outcome evaluation judges that those effects occurred \citep{wang2026thea}.

A change in an object's state may also require revising related properties and relations. A grasp command states what the robot should do; a perception estimate describes the scene inferred from observations; outcome evaluation assesses whether the grasp succeeded. Updating the "holding" relation therefore requires more than recording the command. When a pickup is judged successful, the update should record that the robot holds the cup and revise its previous location or support relation. Thea follows this pattern, but reports cases in which failed actions were judged successful. Incorrect outcome judgments can therefore lead to incorrect state updates that require later correction \citep{wang2026thea}.

\textbf{Using memory for planning and action.}
Current observations may not provide all the information needed to continue a task. For the cup task, remembering where the cup was last seen can guide a later search. Records of earlier actions and their observed outcomes can help the agent assess task progress when the current image is ambiguous.

Different components may need different parts of this history: a local action policy may use recent observations to choose its next movement, while a task planner may need earlier actions and outcomes to determine how to continue. MemoryVLA retains perceptual and semantic information to supplement current policy input during manipulation \citep{MemoryVLA}. RoboMemory uses retained interaction history and object relations to support planning \citep{RoboMemory}. To plan its next manipulation, an agent can query analytic concept-centric memory for an object part’s current state and for changes attributed to earlier actions \citep{sun2026analyticmemory}. What is supplied from memory should therefore match the component's role: choosing a movement or deciding how to continue the task.

Across the section, the comparison concerns what information is retained, what the reasoning model can access for its current task, and how stored estimates are checked and revised over time. 
Action selection is discussed in Section~\ref{sec:decision-making}; outcome evaluation, recovery, and reusable capability improvement are discussed in Section~\ref{sec:feedback}.

%% file: sections/sec4_actions.tex
\section{Agentic Action Interfaces and Execution Requirements}
\label{sec:actionSpaces}\label{sec:action-spaces-interfaces}

\begin{figure*}[!t]
\centering
\includegraphics[width=\textwidth]{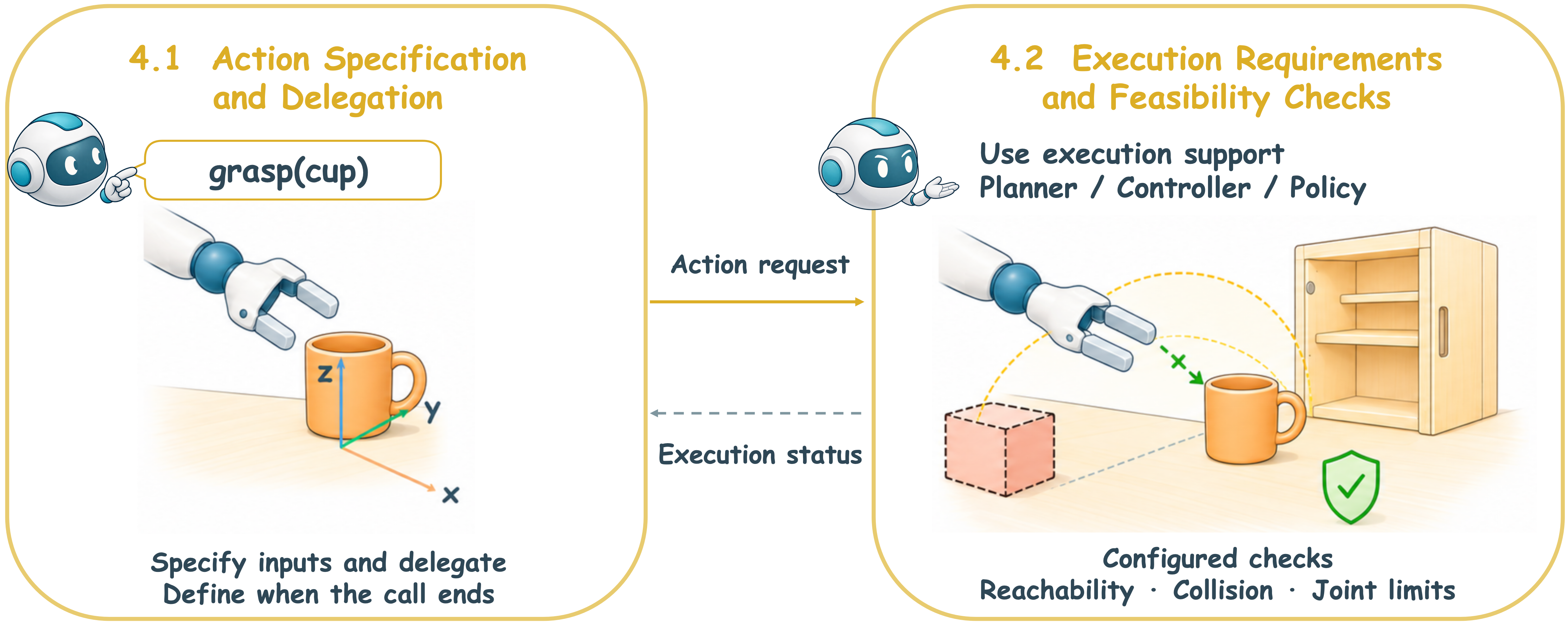}
\caption{Action specification, delegation, and execution support in agentic robotics. Left (Section~\ref{sec:action-specification-delegation}): action interfaces determine the division of work between the agent and the invoked capability, defining the behavior covered by a call, the inputs the agent must specify, and when control returns. Right (Section~\ref{sec:action-grounding-feasibility}): execution modules turn requests into robot motion through motion generation, parameter solving, and configured checks or runtime restrictions. The returned status reports what has been accepted, checked, or completed, with its meaning bounded by the conditions actually assessed. The arrows show functional relationships.}
\label{fig:action-overview}
\end{figure*}

An agent uses action interfaces to call the robot's skills, controllers, or policies. Each interface defines the requested behavior, the inputs the agent must specify, and the details handled by the execution system. Figure~\ref{fig:action-overview} illustrates this division of work with a grasping example. Section~\ref{sec:action-specification-delegation} explains how agents specify actions and delegate execution. Section~\ref{sec:action-grounding-feasibility} describes how execution modules generate motion, resolve parameters, and check constraints.

The option-level formulation in Section~\ref{sec:autonomy-foundations} connects an action interface to execution. For a parameterized option, the invoked policy produces primitive actions as
\begin{equation}
a_t\sim\pi_{\omega_k}(\cdot\mid h_t,\xi_k),\qquad
\tau_k\le t<\tau_{k+1}.
\label{eq:parameterized-execution}
\end{equation}
Here, $k$ indexes high-level decisions, $\omega_k$ is the invoked option, and $\xi_k$ contains its continuous or discrete invocation parameters. The interface determines which details $\xi_k$ must supply and which are delegated to the invoked capability (Section~\ref{sec:action-specification-delegation}). The policy $\pi_{\omega_k}$ produces primitive actions $a_t$ from the interaction history $h_t$ and parameters $\xi_k$; the motion generation, parameter solving, and configured checks supporting this mapping are examined in Section~\ref{sec:action-grounding-feasibility}. The symbol $\sim$ denotes policy sampling; deterministic policies are included as point-mass distributions. The interval from $\tau_k$ to $\tau_{k+1}$ spans one invocation and need not match a single action-prediction chunk. Initiation, termination, and the events returning control follow Equations~\eqref{eq:two-level}--\eqref{eq:trigger}.


\subsection{Action Specification and Delegation}
\label{sec:action-specification-delegation}\label{sec:action-abstraction}\label{sec:embodied-action-interfaces}

Consider the illustrative task of grasping a cup: one interface may accept an object identifier and resolve the grasp pose internally, whereas another requires the explicit target pose as an input. Named operations do similar work: in MoMa-LLM \citep{MoMaLLM}, object- and room-associated behaviors let the agent request navigation or interaction without specifying explicit motion trajectories.

\textbf{Callable behaviors and invocation scope.} Language can describe available skills, condition policy execution, or guide tool use. SayCan \citep{ahn2022can} presents callable skills through natural-language descriptions. Hi Robot \citep{shi2025hi} passes intermediate language commands to a low-level VLA policy, while CLIPort \citep{c5cliport} and RT-H \citep{c5rth} also use language to guide manipulation; in these cases, the executable behavior is bounded by the receiving policy's capabilities. In Thea \citep{wang2026thea}, \texttt{SKILL.md} resources explain how to use robot tools, while the tools carry out the requested actions.

Interface scope determines which procedural steps are handled within a single invocation and which the agent must organize. Sequential controller composition \citep{burridge1999sequential} and SkiROS2 \citep{mayr2023skiros2} support reusable behavior organization, but the interface still decides how much of a procedure is exposed as one call. DexHoldem \citep{chen2026dexholdem} shows several steps grouped into one call: its agent-level \texttt{view\_card(L)} packages picking up, inspecting, and putting down a card, while its router manages intermediate transitions without querying the main agent at each step. CaP-X \citep{c5capx} makes this division configurable: high-level macros handle parts of the procedure internally, whereas lower-level primitives require the generated program to supply more of the sequencing. In single-turn comparisons across abstraction levels, CaP-Bench \citep{c5capx} reports higher task success with human-designed abstractions. Agent comparisons therefore need to control for this procedural support: choosing a precomposed behavior and composing it from its constituent steps place different demands on the agent, even for the same physical task.

Prediction horizon, invocation termination, and the timing of task-level decisions are separate concerns. ACT \citep{c5act} predicts a chunk of actions at a time, and real-time chunking \citep{black2025rtc} addresses how predicted chunks are executed under inference delay. One policy invocation can span several prediction cycles: in Harness VLA \citep{zhang2026harnessvla}, a prompt specifies the requested behavior, and the invocation ends when its stopping condition is met or its budget for action chunks runs out. A chunk boundary therefore need not coincide with a new task-level decision; Section~\ref{sec:decision-making} discusses how those decisions are scheduled.

\begin{table*}[!t]
\centering\small
\caption{Execution support for agentic action. Rows compare a specified request, the mechanism processing it, and its consequence for execution. The listed conditions are local to the indicated mechanism or configuration.}
\label{tab:action-check-evidence}
\renewcommand{\arraystretch}{1.16}
\begin{tabularx}{\textwidth}{@{}p{0.18\textwidth}XXX@{}}
\toprule
\textbf{System or configuration} & \textbf{Request supplied} & \textbf{Processing or check} & \textbf{Consequence for execution}\\
\midrule
LLM$^3$ \citep{wang2024llm3} & Actions with model-selected continuous parameters & Motion planner checks the proposed motion; collision and reachability diagnoses & Supplies a trajectory or identifies problems with the proposal under the scene model.\\
PRoC3S \citep{curtis2024proc3s} & Program with open continuous parameters & Solver searches under encoded constraints & Supplies parameter values when a satisfying solution is found.\\
Thea \citep{wang2026thea}: base motion & Navigation or translation call & Runtime refreshes clearance measurements & Blocks navigation or limits translation according to clearance.\\
Harness VLA \citep{zhang2026harnessvla}: \texttt{vla\_act} & Language subtask, stop predicate, chunk budget & Wrapper checks termination conditions during policy execution & Ends the invocation at its stop condition or budget; final success requires a separate assessment.\\
CaP-X \citep{c5capx}: reported LIBERO-PRO setup & Generated code using configured APIs & Perception and control services; no collision-aware planner in this setup & Executes through the configured services without that planning check.\\
\bottomrule
\end{tabularx}
\end{table*}

\textbf{Target objects and spatial specification.} Identifying the intended object and forming a valid tool call are two distinct requirements. Persistent object references let Thea \citep{wang2026thea} reuse the same reference across queries and action calls, while ETP’s argument schemas \citep{c5etp} define the fields and formats its tools accept. Middleware such as ROS 2 \citep{macenski2022ros2} can carry such references between components. Perceptual object association determines which physical entity a reference identifies, while schema validation checks the call format.

Spatial inputs determine which geometric choices the caller must make before execution. A grasp request that specifies only the object leaves pose selection to the skill, whereas an interface that accepts a position without requiring an orientation leaves orientation to predefined conventions or the execution module. Specifying a pose or path requires the caller to provide additional geometric information about the intended action. These inputs leave different aspects of motion to the execution module: HAMSTER \citep{li2025hamster} passes a coarse 2D end-effector path to a downstream controller that uses 3D scene information to generate motion, whereas Harness VLA’s RoboTwin interface \citep{zhang2026harnessvla} requires world-frame end-effector targets and an \texttt{arm} argument selecting the left arm, right arm, or a bimanual mode.

Visual interfaces allow spatial inputs to be specified through selection or editing. PIVOT \citep{c5pivot} lets the agent select and iteratively refine candidates on an annotated observation image, whereas VIA \citep{c5visualinterface} lets the agent edit virtual gripper targets in a reconstructed 3D scene and uses a downstream controller to execute the resulting waypoints. VIA’s reported manipulation evaluation was conducted in simulation.

\phantomsection\label{sec:information-assistance-actions}
When a required input cannot be established, the agent may need additional perception or clarification \citep{c5capx,ren2023knowno}; Sections~\ref{sec:perception} and~\ref{sec:decision-making} discuss information acquisition and the timing of requests for assistance.

\textbf{Connecting perception to action.} Passing a target from perception to control requires preserving its physical meaning across interfaces. CaP-X’s Contact-GraspNet interface \citep{c5capx} returns grasp poses in the camera frame; the caller transforms them into the frame expected by the action interface and applies any required tool center point (TCP) adjustment.

Generated programs can make this data flow explicit. Code as Policies \citep{liang2023code} and Instruct2Act \citep{c5instruct2act} use perception outputs as inputs to later control calls. In Agent as Policy, a coding agent reads camera images and writes control programs during execution, then revises its commands in light of the robot’s observed response \citep{jia2026agentaspolicy}. These connections have to respect each API’s semantics and coordinate conventions, and the target information has to remain valid at the moment the action is issued.

\subsection{Execution Requirements and Feasibility Checks}
\label{sec:action-grounding-feasibility}

Execution modules turn action requests into robot motion, resolve parameters, and apply the checks enabled in the current configuration. Their feedback indicates whether the checked conditions are met. When initiation is checked using a point estimate, $\hat{s}_{\tau_k}\in\mathcal{I}_{\omega_k}(\xi_k)$ denotes that modeled check. Table~\ref{tab:action-check-evidence} compares representative requests, the mechanisms that process them, and their consequences for execution.

\textbf{Motion generation and platform execution.} Harness VLA \citep{zhang2026harnessvla} exposes different execution mechanisms: \texttt{move\_to} passes Cartesian targets to a control or motion-planning backend, while \texttt{vla\_act} invokes a frozen VLA policy for local behavior involving complex contact. When execution relies on a learned policy, its predictions must be converted into commands accepted by the robot’s control interface. In DexHoldem \citep{chen2026dexholdem}, these commands are joint-position targets for the UR10e–ShadowHand platform, executed in chunks.

\textbf{Parameter solving and feasibility.} Task and motion planning \citep{garrett2021integrated} grounds action parameters in the physical constraints imposed by the robot and scene. Systems divide this responsibility based on whether the model proposes parameter values directly or delegates them to a solver. In LLM\textsuperscript{3} \citep{wang2024llm3}, the model proposes action parameters while a motion planner searches for a collision-free trajectory, explicitly distinguishing goal-pose collisions from inverse kinematics (IK) failures. Conversely, PRoC3S \citep{curtis2024proc3s} generates a program skeleton with undetermined continuous parameters, leaving a constraint solver to search for values satisfying the modeled conditions. In either paradigm, search failure indicates only that no feasible solution was identified within the current model and computational budget.

\textbf{Runtime motion restrictions.} Some constraints are enforced using observations available at request time. Thea \citep{wang2026thea} updates local clearance measurements before base motion, blocks navigation requests when no immediate direction is admissible, and limits direct translations to the admissible distance. The runtime enforces these rules on the current base-motion request, independent of how the model interprets a warning. During simulated execution, the Auriga harness monitors stage tolerances and grasp bounds, halts motion when a check fails, and returns the measured deviation to the agent \citep{wang2026auriga}.

\textbf{Execution checks and task outcomes.} CaP-X \citep{c5capx} supports integration with planning and inverse kinematics services, but its reported LIBERO-PRO configuration lacks motion planning that accounts for collisions. Claims about execution support therefore need to say which checks were enabled in the evaluated configuration.

\phantomsection\label{sec:execution-return}
Harness VLA \citep{zhang2026harnessvla} uses different conditions for primitive termination and for final benchmark success. A reachability check, for example, only concerns the target pose; whether later contact or grasping succeeds still requires observations of execution. Returned status should identify which condition was actually assessed and distinguish request acceptance, a passed check, primitive termination, and task completion. Section~\ref{sec:feedback} discusses how agents use this feedback to assess outcomes and recover from failure.

%% file: sections/sec5_decision.tex
\section{Agentic Decision-Making, Planning and Coordination}
\label{sec:decision}
\label{sec:decision-making}

\begin{figure*}[!t]
\centering
\includegraphics[width=\textwidth]{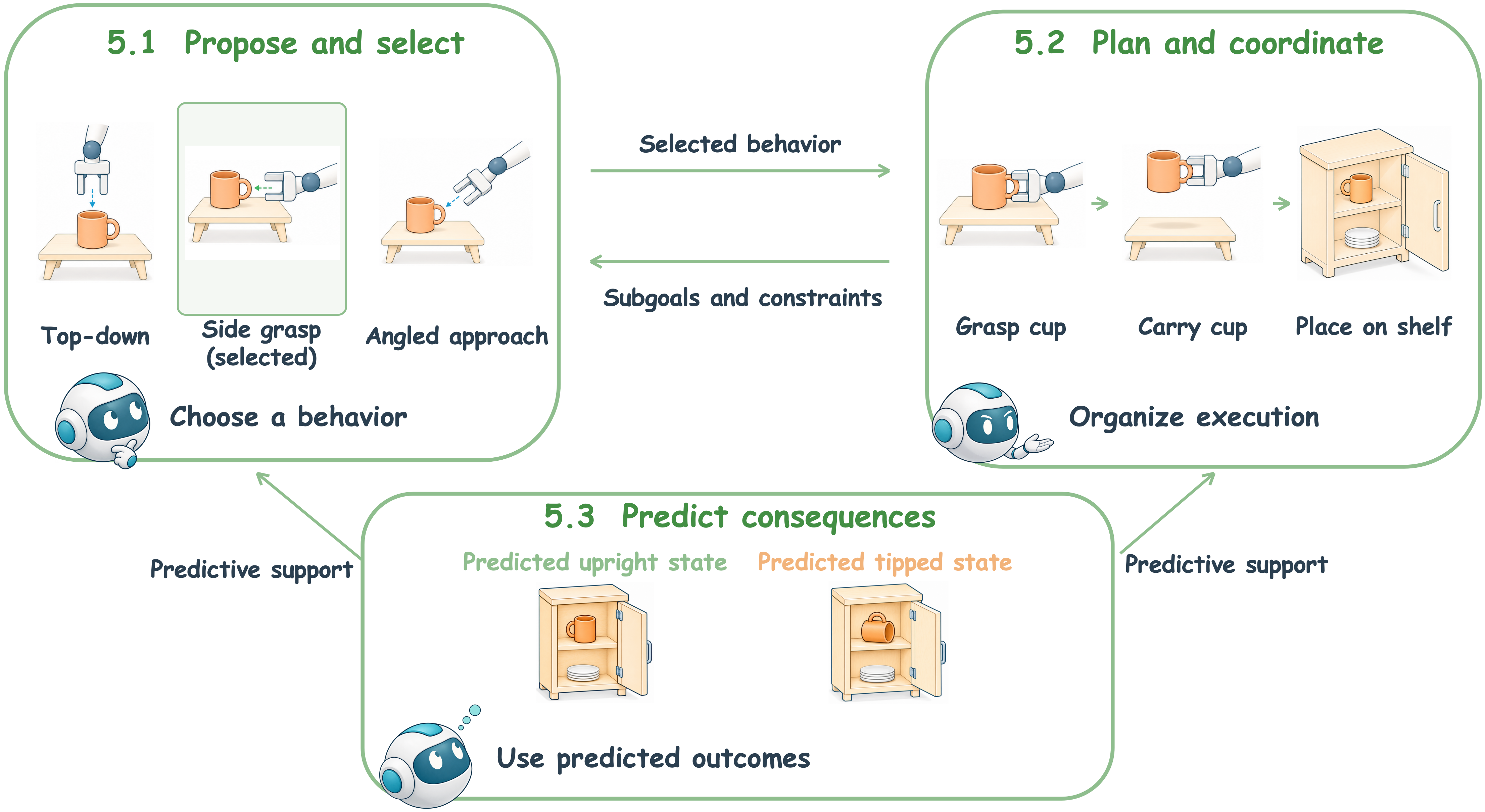}
\caption{Overview of agentic decision-making, planning, and coordination. Action proposal and selection (Section~\ref{sec:selection}) generate candidates and determine which behavior to execute. Planning and coordination (Section~\ref{sec:hierarchy}) organize these choices across task steps, execution schedules, and partners. Predictive models (Section~\ref{sec:prediction}) support both through generated futures used as plans, predicted-consequence scores, or prediction-informed action generation. Selected behaviors inform planning, while subgoals and constraints shape subsequent choices. The arrows show functional relationships, not a required architecture or execution order.}
\label{fig:decision-overview}
\end{figure*}

Using what it perceives (Section~\ref{sec:perception}) and the actions available to it (Section~\ref{sec:actionSpaces}), an agent must decide what the robot does next. This section covers three parts of that decision: choosing the next action (Section~\ref{sec:selection}), planning multi-step tasks and coordinating their execution with other components, robots, and people (Section~\ref{sec:hierarchy}), and using predictive models to inform both (Section~\ref{sec:prediction}). Figure~\ref{fig:decision-overview} shows how the three parts relate.

The decision rule in Equation~\eqref{eq:two-level} conditions on interaction history. In practical implementations, the decision component often receives the task context from Equation~\eqref{eq:perception-context}. For systems organized around parameterized option calls, we write
\begin{equation}
(\omega_k,\xi_k)\sim\mu(\cdot\mid c_{\tau_k},g),\qquad \omega_k\in\Omega.
\label{eq:context-conditioned-decision}
\end{equation}
Here, $\mu$ describes the choice of a behavior and its invocation parameters $(\omega_k,\xi_k)$ (Section~\ref{sec:selection}), with $\Omega$ denoting the available option repertoire. The context $c_{\tau_k}=q(m_{\tau_k},g)$ is the information provided at decision time $\tau_k$; it is not assumed to preserve all information in $h_{\tau_k}$. Planning and coordination organize and revise successive invocations across task steps and participants (Section~\ref{sec:hierarchy}), while predictive models can support these choices by generating plans, evaluating consequences, or guiding action generation (Section~\ref{sec:prediction}). The chosen invocation is executed through Equation~\eqref{eq:parameterized-execution}. This option-level expression is a functional abstraction: direct policies and program-generating systems need not expose a separate selector or an explicit skill library.


\begin{figure*}[t]
\centering
\includegraphics[width=0.90\textwidth]{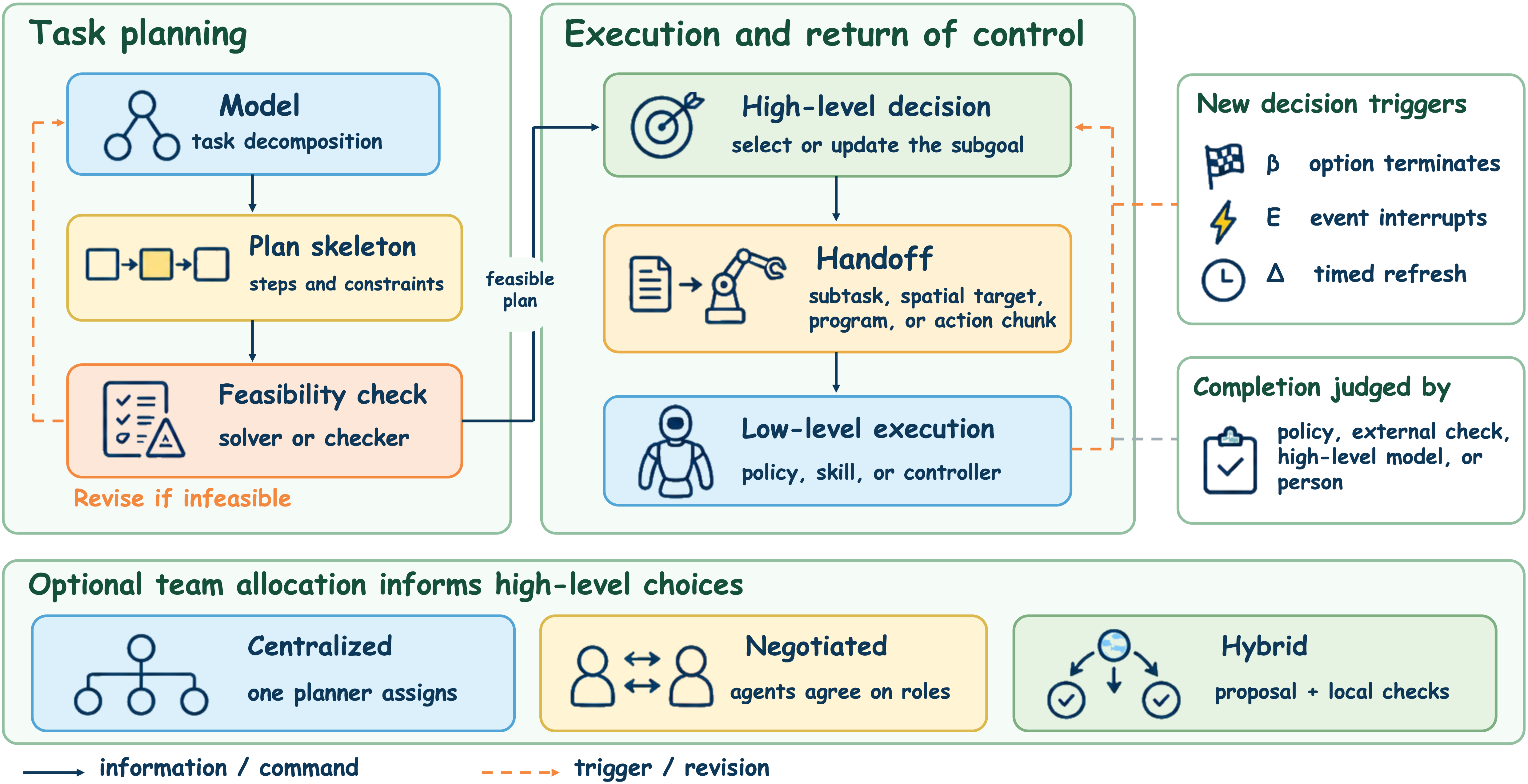}
\caption{How the agent organizes multi-step execution. Decomposition and feasibility checks inform its choices, while the runtime loop determines what is handed to lower levels, when control returns, and who judges completion. Dashed arrows mark failed checks and decision triggers: termination ($\beta$), interruption ($E$), or interval expiry ($\Delta$; ``timed refresh''), following Equation~\eqref{eq:trigger}. When the agent refreshes its decision, it may choose to continue the current skill. Team allocation, when needed, is centralized, negotiated, or hybrid.}
\label{fig:orchestration}
\end{figure*}

\begin{table*}[t]
\small\sf\centering
\setlength{\tabcolsep}{4pt}
\renewcommand{\arraystretch}{1.12}
\caption{Control handoffs between decision levels. In the Interface column, Lang denotes language, Spatial denotes a spatial target, Tool denotes a tool call, and Chunk denotes an action chunk. The triggers follow Equation~\eqref{eq:trigger}: T denotes behavior termination ($\beta_\omega$), E denotes interruption, and P denotes interval expiry ($\Delta$, fixed or predicted). O denotes one decision per episode. Completion evidence is listed separately from return triggers. In the Evaluation column, Sim denotes simulation, Real denotes a physical robot, Mani denotes manipulation, and MoMa denotes mobile manipulation. The superscript $^{q}$ indicates that only qualitative demonstrations or videos are reported. The abbreviation n.r. means not reported, and -- indicates that the reported loop has no high-level completion judgment.\label{tab:orchestration}}
\begin{tabularx}{\textwidth}{@{}>{\raggedright\arraybackslash}p{2.5cm}>{\raggedright\arraybackslash}X>{\raggedright\arraybackslash}p{2.1cm}>{\raggedright\arraybackslash}X>{\raggedright\arraybackslash}p{2.3cm}>{\raggedright\arraybackslash}p{3.0cm}@{}}
\toprule
System & Interface & Trigger & Completion evidence & Invocation (reported) & Evaluation \\
\midrule
Inner Monologue \citep{huang2022inner} & Lang: named skill & T (skill completion) & External (success detector); human & Per skill & Sim-Mani; Real-Mani; Real-MoMa \\
Hi~Robot \citep{shi2025hi} & Lang: subtask & P; E (user interjection) & High-level inference & About 1 Hz & Real-Mani; Real-MoMa (3 platforms) \\
HAMSTER \citep{li2025hamster} & Spatial: 2D path & O & -- & Once per episode & Real-Mani; Sim-Mani \\
OneTwoVLA \citep{lin2025onetwovla} & Lang reasoning; Chunk & T (learned switch) & High-level inference & Learned & Real-Mani \\
Hi-VLA \citep{hu2026orchestrating} & Lang: subgoal & T; P (fixed or predicted) & VLM detection; timers are not success evidence & 8 s default; varied & Sim-Mani; Real-Mani (1 task) \\
CoPAL \citep{joublin2024copal} & Lang: steps and commands & E (inconsistency) & External (rule and simulation checks) & n.r. & Sim-Mani; Real-Mani$^{q}$ \\
MALMM \citep{singh2024malmm} & Tool: generated waypoint code & T (after each step) & High-level inference (planner) & Per step & Sim-Mani; Real-Mani \\
Pigey \citep{galanti2026pigey} & Tool: TAMP or VLA call & T (tool-call result) & Sensor flags plus agent visual inference & n.r. & Sim-Mani; Real-Mani (30 probes) \\
\bottomrule
\end{tabularx}
\end{table*}

\subsection{Action Proposal and Selection}
\label{sec:selection}
At each decision point, an agent can choose an available skill, write a program or planning objective, or let a learned policy output actions directly. When it has several candidates, it must also judge which one to execute, or ask for help when it cannot decide.

\textbf{Selecting skills by task relevance and feasibility.} An agent can choose among available skills by considering both their relevance to the task and their likelihood of successful execution. SayCan \citep{ahn2022can} multiplies a language-model score for task relevance by a learned value estimate of skill success, then selects the highest-scoring skill. Removing the value estimate reduces human-rated plan success in its mock kitchen from 84\% to 67\%. SayCanPay \citep{hazra2024saycanpay} extends this approach with a learned long-horizon payoff, while Grounded Decoding \citep{huang2023grounded} combines language and grounding scores during token generation. CLASP uses skill descriptions that specify parameters and preconditions to select a skill, bind its arguments, and compose skills when a single one does not suffice \citep{knauer2026clasp}.

\textbf{Generating programs or planning objectives.} A foundation model can generate a program that controls the robot or an objective that guides a planner. Code as Policies \citep{liang2023code} generates programs that use perception results to invoke control APIs and adapt behavior to observed conditions. These programs can directly implement policies without a separate candidate-ranking stage. VoxPoser \citep{huang2023voxposer} instead generates 3D value maps that specify objectives for a model-predictive planner.

\textbf{Generating actions directly.} Vision-language-action policies such as RT-2 \citep{pmlr-v229-zitkovich23a}, OpenVLA \citep{kim2024openvla}, and $\pi_0$ \citep{black2024pi0} generate actions or action chunks directly from observations and task instructions. Their performance depends on their training data and on the inputs they receive. Intermediate outputs, such as the subtask predicted by $\pi_{0.5}$ \citep{intelligence2025pi05visionlanguageactionmodelopenworld} or the plan in Embodied Chain-of-Thought (ECoT; \citealp{zawalski2024ecot}), can be inspected or corrected by a person or another model, but they do not guarantee that the action is feasible.
A policy can also generate multiple candidates for evaluation before execution. V-GPS \citep{nakamoto2024vgps} and RoboMonkey \citep{kwok2025robomonkey} use separate evaluators, Hume \citep{song2025hume} uses a value head, and FOREWARN \citep{wu2025forewarn} evaluates predicted outcomes (Section~\ref{sec:prediction}). Generating more candidates increases computation and helps only when the policy produces useful actions and the evaluator ranks them correctly.

\textbf{Deferring when selection is uncertain.} If an agent is unsure which action to take, it can ask for help before proceeding. KnowNo \citep{ren2023knowno} acts autonomously when conformal prediction returns a single-option set; the coverage level controls the trade-off between help requests and selection errors. Its guarantee assumes exchangeable scenarios and covers whether the correct option is in the set, not whether execution succeeds. Introspective Planning (IntroPlan; \citealp{liang2024introspective}) uses retrieved plans to inform uncertainty estimates, while Uncertainty-aware Policy Steering (UPS; \citealp{yuan2026ups}) calibrates a verifier over sampled actions. These methods decide whether the agent should act on its proposal or ask first.


\subsection{Task Planning and Execution Coordination}
\label{sec:hierarchy}

Many tasks require several steps. This subsection covers how agents build plans that the robot can execute, revise them while the robot acts, and coordinate with other robots and people (Figure~\ref{fig:orchestration}).

\textbf{Constructing feasible task plans.}\label{sec:decomposition}
A feasible task plan breaks the goal into steps that the robot can carry out \citep{garrett2021integrated}. Foundation models can describe the planning problem and leave the search for a step sequence to a classical planner. LLM+P \citep{liu2023llmp} expresses the task in the Planning Domain Definition Language (PDDL) under a hand-written domain model. SayPlan \citep{rana2023sayplan} reduces the search space through selective scene-graph expansion. The planner guarantees that the plan is consistent with the symbolic model, but not that the robot can physically execute it.

The plan must also respect physical constraints, such as reachability and collision avoidance. PRoC3S \citep{curtis2024proc3s} generates a program with unresolved continuous parameters that a solver determines, while ReKep \citep{huang2025rekep} expresses task requirements as keypoint constraints. Planning can also alternate between proposals and checks: LLM$^3$ \citep{wang2024llm3} revises proposals after motion-planning failures. These checks are only as reliable as the scene representation and the constraints included in the problem.

Each action must also leave the robot able to complete the remaining steps. A grasp may be executable but unsuitable for the intended placement. Text2Motion \citep{lin2023text2motion} uses learned dynamics and skill-value estimates to assess such dependencies across steps.

\textbf{Coordinating planning and execution.}\label{sec:handoffs}
A high-level planner sets subgoals or instructions for the robot and decides when to update them (Section~\ref{sec:actionSpaces}). Updates may follow behavior termination, an interrupting event, or a scheduled interval (Equation~\eqref{eq:trigger}). Inner Monologue \citep{huang2022inner} returns control after each skill, whereas Hi~Robot \citep{shi2025hi} updates periodically or in response to a user interjection. OneTwoVLA \citep{lin2025onetwovla} learns when to switch between reasoning and action generation, including before a subgoal is complete.

Execution feedback helps determine whether to continue or revise the plan. ProgPrompt \citep{singh2023progprompt} checks preconditions before each step, while LLM-Planner \citep{song2023llmplanner} revises its plan after subgoal failure or timeout. Both systems check and revise the plan while it runs instead of verifying the whole sequence in advance. At each update, the agent can continue the current behavior, change its instruction, or move to the next subgoal. Moving on requires evidence that the subgoal is complete; an expired timer or a returned tool call does not provide this evidence. Section~\ref{sec:feedback} covers how agents interpret this evidence and recover from failure.

Table~\ref{tab:orchestration} compares update mechanisms, timing, and completion assessment across systems. Systems also differ in which component judges completion: in MALMM \citep{singh2024malmm}, the planner assesses outcomes while a supervisor assigns work to the other components. The low-level policy can also update on its own, faster schedule: real-time chunking \citep{black2025rtc} computes the next action chunk while the current one executes, and HiRT \citep{zhang2024hirt} asynchronously refreshes the vision-language features used by a fast policy. These updates keep the robot moving smoothly without changing the task plan.

\textbf{Coordinating robots and people.}\label{sec:coordination}
Robots working together must divide the task according to their capabilities and progress. SMART-LLM \citep{kannan2024smartllm} combines task decomposition and allocation in a generated program. RoCo \citep{mandi2024roco} lets robots negotiate plans through dialogue and revise them after motion-planning feedback; its negotiated plans are executed through a centralized multi-arm motion planner. For heterogeneous teams, COHERENT \citep{liu2025coherent} updates assignments using robot feedback.

Working with people also requires the robot to track what its partner is doing and adjust its next action. CoELA \citep{zhang2024coela} supports cooperation through communication, while MOSAIC \citep{wang2024mosaic} uses verbal reports to track a partner's progress. Learning to Assist \citep{xu2026learnassist} addresses a robot handing over an object before the person is ready, which happens when an action chunk continues past a change in the task that the robot has not observed; it corrects this by steering the policy at inference time. Human participation can also take the form of clarification, as in KnowNo (Section~\ref{sec:selection}), or approval before execution, as in ProVox \citep{grannen2025provox}. PARTNR \citep{chang2025partnr} provides a benchmark for evaluating human-robot collaboration under spatial, temporal, and agent-capability constraints.

\subsection{Predictive Models for Action Evaluation and Planning}
\label{sec:prediction}

A predictive model, often called a world model, forecasts what will happen if the robot acts. Agents use these forecasts in three ways: generating a desired future and deriving actions from it, comparing the predicted results of candidate actions, and guiding a policy while it generates actions.

\textbf{Planning with generated futures.} An agent can first generate a video of the desired behavior and then derive actions from it. Follow Your Pose \citep{ma2024follow} generates human motion videos conditioned on a pose sequence, while UniPi \citep{du2023unipi} uses inverse dynamics to infer robot actions from changes between generated frames; its control experiments are conducted in simulation. VLP \citep{du2024vlp} ranks video plans by estimated progress, while RIGVid \citep{patel2026rigvid} rejects videos that violate the instruction. SuSIE \citep{black2024susie} instead passes a subgoal image to a goal-conditioned policy. These methods work only when the robot can physically achieve the generated future.

\textbf{Evaluating candidate actions.} An agent can predict what candidate actions would do and compare their expected outcomes. DINO-WM \citep{zhou2024dinowm} and V-JEPA~2-AC \citep{assran2025v} predict how actions change a learned representation of the scene, then search for the action sequence whose predicted result is closest to the representation of a goal image. With model-predictive control (MPC), the robot executes the first part of the sequence and replans from new observations. Hierarchical World Models (HWM; \citealp{zhang2026hwm}) use longer-term predictions as subgoals. World Action Planner uses imagined rollouts to identify potential failures and revise the overall plan, then compares the predicted outcomes of nearby actions to choose the next move \citep{zhang2026worldactionplanner}.

Predictions can also help select among actions generated by a pretrained policy (Section~\ref{sec:selection}). Cosmos Policy \citep{kim2026cosmos} predicts a future state and value for each action chunk, then executes the highest-valued candidate. FOREWARN \citep{wu2025forewarn} uses a VLM to judge descriptions of predicted outcomes. EmbodiedSkills compares candidates from a frozen policy using both their predicted outcome scores and the disagreement among independently trained world action models \citep{wang2026embodiedskills}.

\textbf{Using prediction during action generation.} A policy can also use future predictions while generating actions, without a separate ranking step. Unified World Models (UWM; \citealp{zhu2025uwm}) and DreamZero \citep{ye2026dreamzero} jointly generate actions and future observations. Video Prediction Policy \citep{hu2025vpp} uses predictive video features to generate actions, while Seer \citep{tian2025seer} conditions action tokens on visual-foresight tokens. These methods use prediction during both training and inference. Candidate selection can still be added by comparing multiple generated rollouts \citep{zhao2026geobon,zhang2026wcd}.

Prediction is useful when it improves task success. Prediction errors grow with the horizon, and comparing more candidates gives scoring errors more chances to pick a wrong action. Comparisons with direct execution should therefore also report the extra computation spent on prediction and search.

%% file: sections/sec6_feedback.tex
\section{Closed-Loop Feedback and Capability Improvement}
\label{sec:feedback}

\begin{figure*}[!t]
\centering
\includegraphics[width=\textwidth]{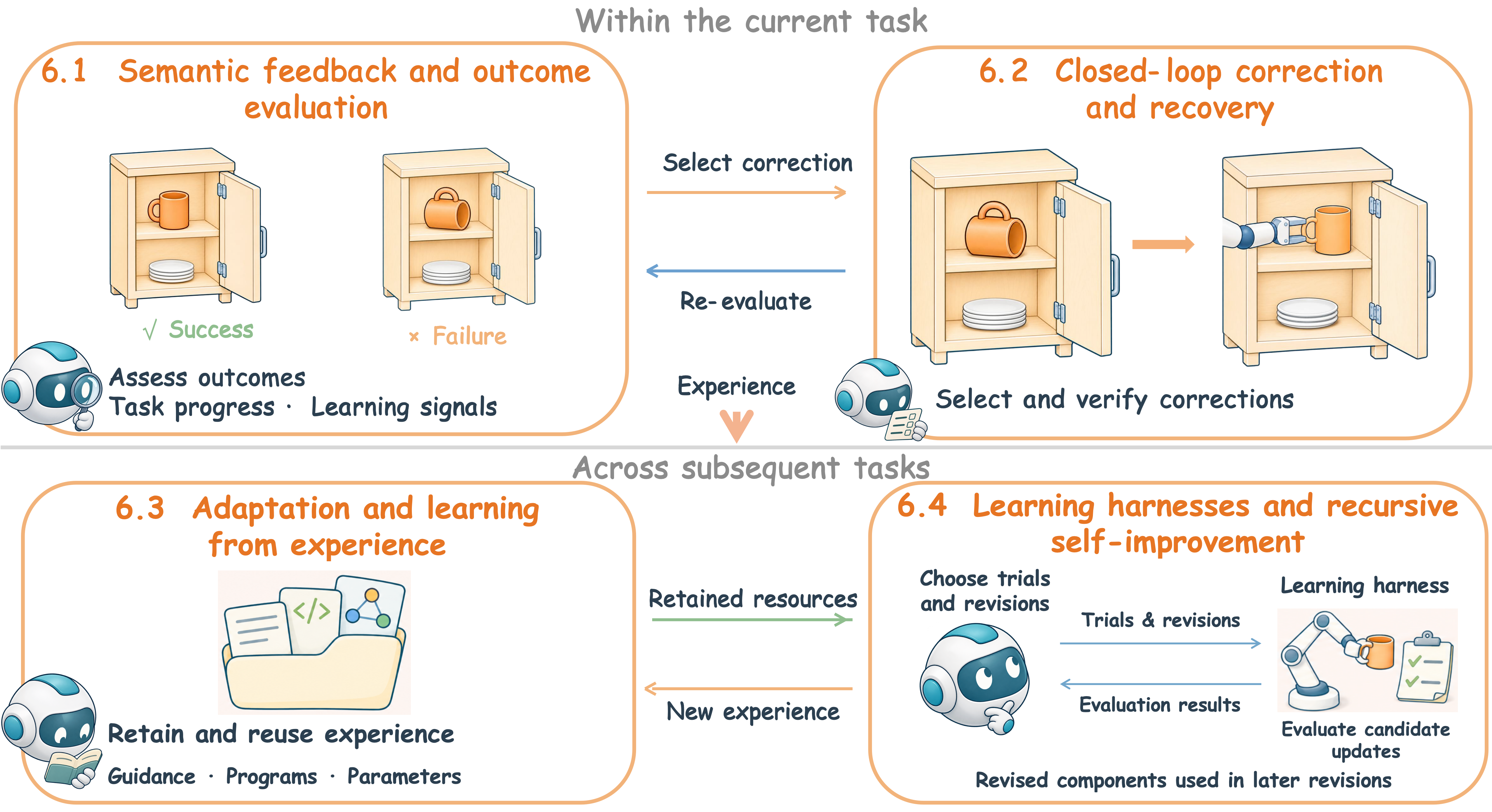}
\caption{Overview of closed-loop feedback and capability improvement. Within the current task, outcome evaluation (Section~\ref{sec:feedback-evaluation}) informs correction and recovery (Section~\ref{sec:feedback-recovery}), followed by reassessment. Across subsequent tasks, experience may be retained as guidance, program code, or model parameters (Section~\ref{sec:feedback-learning}). Agents and learning harnesses organize further trials and evaluate candidate revisions (Section~\ref{sec:feedback-orchestration}); retained revisions can inform subsequent improvement attempts.}
\label{fig:c7-overview}
\end{figure*}

Feedback helps robots adjust their behavior based on observed outcomes \citep{astrom2008feedback,ingrand2017deliberation}. This section describes how agents use feedback to assess execution, recover from failures, retain useful experience, and organize further learning. Foundation models support these decisions by interpreting execution feedback. Figure~\ref{fig:c7-overview} shows how these functions connect within a task and across later tasks.

We distinguish primitive execution steps $t$, high-level decision times $\tau_k$, and persistent capability-update rounds $j$, which need not coincide. During an invocation, execution evidence can be assessed as
\begin{equation}
y_t=\mathcal{V}(h_{\tau_k:t},g,\omega_k,\xi_k).
\label{eq:execution-assessment}
\end{equation}
Here, $h_{\tau_k:t}$ contains the observations and actions available from the invocation's start through time $t$. Outcome evaluation (Section~\ref{sec:feedback-evaluation}) is represented by $\mathcal{V}$, which uses this evidence, the goal $g$, and the invoked behavior $(\omega_k,\xi_k)$ to produce an assessment $y_t$, such as a completion judgment, progress estimate, or failure diagnosis. This assessment need not be correct or binary and can inform correction and recovery (Section~\ref{sec:feedback-recovery}) under Equations~\eqref{eq:two-level}--\eqref{eq:trigger}.

Changes retained beyond the current invocation are represented by
\begin{equation}
\Theta_{j+1}=U_j(\Theta_j,\mathcal{D}_j).
\label{eq:feedback-update}
\end{equation}
The state $\Theta_j$ may include skills, programs, model parameters, or learning-harness components; Section~\ref{sec:feedback-learning} examines what experience modifies and retains. The procedure $U_j$ uses experience and verification evidence $\mathcal{D}_j$, which may include earlier traces. Section~\ref{sec:feedback-orchestration} examines how further practice and verification are organized to supply this evidence and select revisions. The procedure $U_j$ may leave $\Theta_j$ unchanged when no candidate update is adopted. Unlike the information update $u$ in Equation~\eqref{eq:perception-state-update}, $U_j$ describes retained capability changes, which may occur during practice or between task trials.

\subsection{Semantic Feedback and Outcome Evaluation}
\label{sec:feedback-evaluation}

Execution feedback helps agents judge task outcomes and guide learning. We discuss three uses: verifying task completion, estimating progress and how well behavior matches the task, and learning from preferences and rewards. Table~\ref{tab:c7-evaluation} compares the feedback signals and observations used by representative methods.

\noindent\textbf{Verifying task completion.}
Completion checks help an agent decide whether a task step has achieved its goal and whether to move on. Inner Monologue \citep{huang2022inner} passes skill success signals and scene descriptions to a planner. Manipulate-Anything \citep{c7pub_c5manipulateanything} generates verification conditions and selects views to check them, while Thea \citep{wang2026thea} compares camera observations with the expected outcomes of a tool call. After a skill ends, the agent must check its outcome against the task requirements. Grasping a cup and placing it stably, for example, require different checks.

\noindent\textbf{Estimating progress and task correspondence.}
Progress estimates indicate how far the robot has advanced toward its goal. Task correspondence measures how closely observed behavior matches a task description or demonstration. LIV \citep{c7liv} supports dense rewards through language--image value representations. RoboCLIP \citep{c7roboclip} uses the match between an episode video and a task specification as a terminal reward. GVL \citep{ma2025gvl} estimates progress across multiple frames and also supports success detection.
Using a progress score to judge completion requires a decision rule tested against the task's requirements. Feedback must also be available when the decision is made. A score that uses later video frames cannot directly guide an earlier intervention.

\noindent\textbf{Evaluating attempts with preferences and rewards.}
Preferences compare task attempts, while reward scores express how well an attempt meets the task requirements. A preference indicates which attempt is better, even when neither completes the task. RL-VLM-F \citep{c7rlvlmf} learns rewards from VLM preferences, while Offline RL-VLM-F \citep{c7offlinevlmf} learns rewards from preferences over collected data. RoboReward \citep{lee2026roboreward} includes graded outcomes for unsuccessful and partially completed attempts, while Robometer \citep{c7robometer} combines progress and success supervision with trajectory comparisons.

\begin{table*}[t]
\centering\small
\setlength{\tabcolsep}{4pt}\renewcommand{\arraystretch}{1.12}
\caption{Feedback semantics, observation support, and use. Evaluators are included alongside agent systems to identify the signals available for decision-making. The listed uses are not mutually exclusive; outcome, progress, and preference signals may support multiple decisions or learning objectives.}
\label{tab:c7-evaluation}
\begin{tabularx}{\textwidth}{@{}>{\raggedright\arraybackslash}p{0.19\textwidth}*{3}{>{\raggedright\arraybackslash}X}@{}}
\toprule
Work & Feedback & Observation support & Use \\
\midrule
Inner Monologue \citep{huang2022inner} & Skill outcomes; scene descriptions & Returned grounding-module outputs & Revise skill choices \\
Manipulate-Anything \citep{c7pub_c5manipulateanything} & Subtask verification & Selected views of execution & Verify and retry \\
Thea \citep{wang2026thea} & Tool-outcome judgments & Post-conditions and camera observations & Continue or reconsider execution \\
LIV \citep{c7liv} & Language--image value representation & Images conditioned on language or image goals & Representation and reward learning for control \\
RoboCLIP \citep{c7roboclip} & Episode--task correspondence & Episode video; text or demonstration-video task specification & Terminal learning reward \\
GVL \citep{ma2025gvl} & Task progress & Joint frame context; initial anchor & Progress estimation; success detection; policy learning \\
RL-VLM-F \citep{c7rlvlmf} & Pairwise preferences & Observed images & Fit reward; train policy \\
Offline RL-VLM-F \citep{c7offlinevlmf} & Preferences on collected data & Image judgments; point-cloud learning in physical dressing & Offline reward/policy training \\
RoboReward \citep{lee2026roboreward} & Graded task outcomes & Task-conditioned videos with relabeled/clipped examples & Reward prediction; policy learning \\
Robometer \citep{c7robometer} & Progress; success; trajectory preferences & Within/across-trajectory observations & Learn a reward model for downstream evaluation and control \\
\bottomrule
\end{tabularx}
\end{table*}



\subsection{Closed-Loop Correction and Recovery}
\label{sec:feedback-recovery}

Agents use feedback to correct execution and recover from failures. Correction adjusts ongoing behavior in response to deviations or instructions; recovery restores the conditions needed to continue after failure or interruption. These functions can overlap. We discuss how agents diagnose problems, choose what to change, and carry out repairs (Table~\ref{tab:c7-recovery}).

\noindent\textbf{Diagnosing execution problems.}
Diagnosis helps an agent choose a repair, such as relocalizing an object or changing an unstable grasp. REFLECT \citep{c7pub_REFLECT} summarizes multisensory experience to locate failures and plan corrections. AHA \citep{c7aha} learns visual failure detection and reasoning from perturbed simulated demonstrations and provides diagnostic feedback to manipulation frameworks. Its integration experiments assess how this feedback helps recovery, while the receiving framework determines the executable response. If a repair fails, new observations can guide a revised diagnosis or a different action. A successful retry supports the usefulness of a repair; confirming the diagnosed cause requires further evidence.

\noindent\textbf{Choosing what to change.}
An agent can adjust motion, revise a program or plan, or prepare the physical conditions needed for execution. The available action interfaces determine which changes it can make (Section~\ref{sec:embodied-action-interfaces}). RACER \citep{c7racer} uses a VLM supervisor to give language guidance to an actor trained with recovery trajectories. DROC \citep{c7droc} interprets human corrections to revise plans or skill programs. RACER relies on a trained corrective-action interface, while DROC depends on human feedback for planning and code changes.

The detail in execution feedback affects how precisely an agent can revise a program. CaP-Agent0 \citep{c5capx} uses interactive execution feedback to revise programs; ASPIRE \citep{c7aspire} uses primitive return states and visual traces for targeted repair and re-execution. Harness VLA \citep{zhang2026harnessvla} changes the conditions under which a frozen VLA is called: a coding agent re-grounds targets and uses analytic primitives to prepare the scene for another attempt. Recovery can thus revise behavior or prepare for execution without updating the policy. For example, changing a grasp trajectory addresses a different problem from opening a cabinet or removing an obstruction. When the robot lacks an action that addresses the diagnosed problem, it may need to gather more observations, request assistance, or stop.

\noindent\textbf{Timing repairs and checking recovery.}
A repair must be feasible from the robot's current state and arrive in time to change its behavior. SteerAct \citep{zhou2026steeract} selects a feature from successful and failed reference rollouts, then uses its activation pattern during execution to determine when and how strongly to steer the frozen action model. CheckVLA \citep{c7checkvla} uses action-conditioned predictions to check execution and accounts for inference latency when revising the remaining actions. Actions already executed remain unchanged. In RoboCasa365 simulation, CheckVLA achieves an average success rate of 36.1\%, compared with 27.6\% for periodic replanning, under a common training recipe and a matched invocation budget. This comparison evaluates the full monitoring and repair system.

Before resuming execution, the agent needs to check that the robot and scene are ready. A failed attempt may leave an object displaced, the gripper closed, or the next motion obstructed. Recovery may therefore require another observation and preparatory actions. ASPIRE \citep{c7aspire} re-executes repaired programs, while Harness VLA \citep{zhang2026harnessvla} restores the conditions for policy execution. Outcome checks (Section~\ref{sec:feedback-evaluation}) then determine whether the required conditions hold. A renewed grasp, for example, enables transport; completing placement requires a further check. If the conditions remain unmet, new feedback can guide another correction.

Execution checks also help an agent decide whether to retry, seek assistance, or stop. DexHoldem \citep{chen2026dexholdem} connects execution verification to subsequent primitive execution, recovery retries, and requests for human assistance in physical tabletop interaction. Across 33 rollouts, each covering one poker hand, it reports verified completion of 12 of 34 dispatched retries. Controlled comparisons are still needed to measure recovery's contribution to overall task completion. If verification remains inconclusive or the retry budget is exhausted, requesting assistance or stopping may be appropriate.

Effective recovery depends on the robot's current state, available observations, and executable actions. The agent uses these to choose a repair and revises its decisions as new outcomes arrive. Section~\ref{sec:feedback-learning} describes how experience from these interactions can be retained for later tasks.

\begin{table*}[t]
\centering\small
\setlength{\tabcolsep}{4pt}\renewcommand{\arraystretch}{1.12}
\caption{Diagnostic support and intervention scopes for correction and recovery. Entries identify the feedback guiding an intervention, what can be changed, and the supporting execution mechanism; diagnostic components need not implement the complete loop themselves. Some systems also retain experience for later use (Table~\ref{tab:c7-updates}).}
\label{tab:c7-recovery}
\begin{tabularx}{\textwidth}{@{}>{\raggedright\arraybackslash}p{0.19\textwidth}*{3}{>{\raggedright\arraybackslash}X}@{}}
\toprule
Work & Diagnostic/corrective input & Intervention target & Role in correction or recovery \\
\midrule
REFLECT \citep{c7pub_REFLECT} & Multisensory summaries & Task plan & Failure localization and corrective planning \\
AHA \citep{c7aha} & Learned visual failure reasoning & Downstream framework's response & Diagnostic feedback for recovery \\
RACER \citep{c7racer} & Supervisor language; observations & Corrective motion & Trained language-conditioned actor \\
DROC \citep{c7droc} & Human language corrections & Plan or skill program & Planning and robot-code revision \\
CaP-Agent0 \citep{c5capx} & Interactive execution feedback & Robot program & Behavior revision through available robot APIs \\
ASPIRE \citep{c7aspire} & Primitive returns; visual traces & Program fragments & Targeted repair and re-execution \\
CheckVLA \citep{c7checkvla} & Action-conditioned predictions; observations & Still-executable action suffix & Execution-time checking and latency-aware correction \\
Harness VLA \citep{zhang2026harnessvla} & Observed state; execution guidance & VLA invocation conditions & Re-staging and retrying a frozen VLA \\
DexHoldem \citep{chen2026dexholdem} & Parsed tabletop state; execution verification & Primitive continuation/retry; assistance requests & Physical evaluation of continuation, recovery, and escalation \\
\bottomrule
\end{tabularx}
\end{table*}

\subsection{Adaptation and Learning from Experience}
\label{sec:feedback-learning}

\begin{figure*}[t]
\centering
\includegraphics[width=\textwidth]{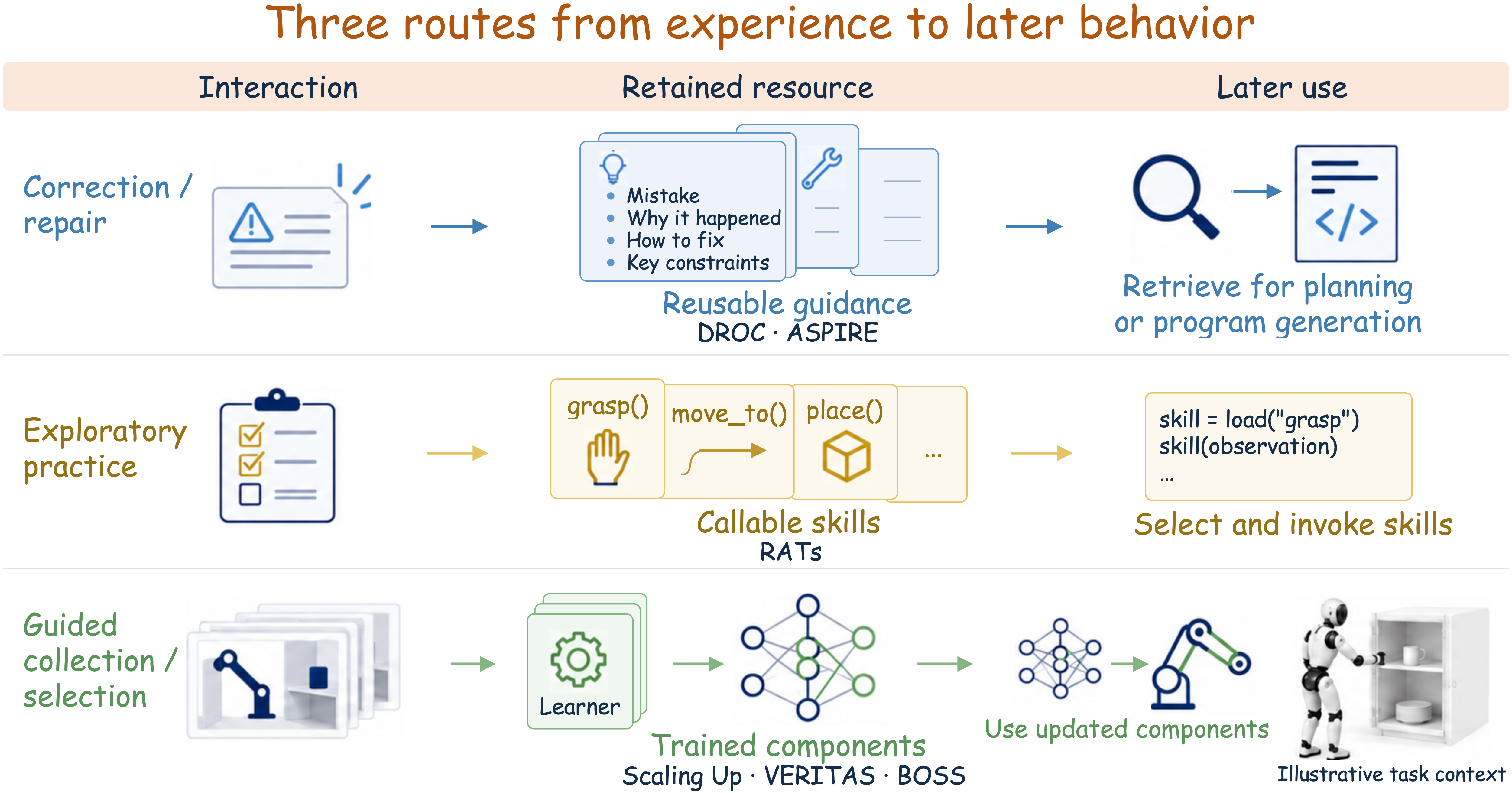}
\caption{Three forms of retained content connecting experience to later behavior: guidance, reusable robot programs stored as source code, and updated model parameters. DROC \citep{c7droc} and ASPIRE \citep{c7aspire} retain guidance from corrections or repairs; RATs \citep{c7rats} retains reusable code skills; Scaling Up \citep{c7scalingup}, VERITAS \citep{c7veritas}, and BOSS \citep{c7boss} use collected or practiced behavior to train policies. The examples highlight particular retained content within each system; multiple forms can be combined. In the figure, callable skills denote retained program code, and trained components denote models with updated parameters. Their value depends on subsequent use and performance.}
\label{fig:c7-loops}
\end{figure*}

Agents can use experience from corrections, successful execution, and practice to guide later behavior. We discuss three ways to retain this experience: guidance, reusable programs, and updated model parameters (Figure~\ref{fig:c7-loops}; Table~\ref{tab:c7-updates}). A system can combine these forms.

\noindent\textbf{Reusing guidance and programs.}
Agents can reuse guidance and programs from earlier interactions. DROC \citep{c7droc} extracts knowledge from corrections for later planning and skill composition. ASPIRE \citep{c7aspire} turns validated program repairs into guidance for future code generation, including when to apply each repair. RATs \citep{c7rats} stores code skills acquired through interaction for later program generation and execution. Guidance helps the agent construct new plans or programs. Stored programs can be executed again or used as examples for generating new code.

Before reusing guidance or a program, the agent needs to check that it fits the current task and scene. A correction tied to one object pose or a program requiring an unobstructed approach may fail when those conditions change. RATs keeps its library fixed during downstream tests, while its execution system still verifies outcomes and allows bounded retries. This separates library acquisition from adaptation during execution. Section~\ref{sec:embodied-memory} covers access to retained information, and Section~\ref{sec:embodied-action-interfaces} describes interfaces for invoking executable skills.

\noindent\textbf{Updating model parameters.}
Policies can learn from experience through parameter updates. Outcome judgments and rewards (Section~\ref{sec:feedback-evaluation}) help select training data or guide these updates. Scaling Up \citep{c7scalingup} uses language to guide data collection and checks task success before policy distillation. VERITAS \citep{c7veritas} trains policies on successful rollouts guided by a visual verifier that selects action candidates. BOSS \citep{c7boss} trains policies on experience from skill chains and their composite descriptions to expand the behaviors they can execute. Its physical experiments still rely on human resets and primitive success detection.

Experience can update coding models and world models as well as action policies. CaP-RL \citep{c5capx} trains the coding model from execution rewards, changing the programs it subsequently generates. Its reported training uses privileged S1 simulation APIs with ground truth masks and object poses, while evaluation includes S2 perception without privileged state access and physical cube manipulation. VLAW \citep{c7vlaw} updates a world model with physical rollout data, generates synthetic experience with that model, and uses a reward model to select data for further VLA training.

Expert feedback and human guidance can also support policy updates. DAgger \citep{c7dagger}, HG-DAgger \citep{c7hgdagger}, and ThriftyDAgger \citep{c7thrifty} use expert feedback for imitation learning. HIL-SERL \citep{c7hilserl} and RECAP \citep{intelligence2025pi06vlalearnsexperience} combine experience with human guidance for policy learning.

Retained experience needs to be tested on later tasks. Guidance, programs, and parameter updates may depend on the conditions in which they were acquired, and a change that helps one task may impair another. Evaluation should therefore track both gains and regressions (Section~\ref{sec:c8-feedback}).



\begin{table*}[t]
\centering\small
\setlength{\tabcolsep}{4pt}
\renewcommand{\arraystretch}{1.12}
\caption{Experience sources and update mechanisms, retained content, and subsequent use. The table distinguishes guidance, robot programs stored as source code, and model parameters while recording combinations where relevant. A system may retain several forms of content.}
\label{tab:c7-updates}
\begin{tabularx}{\textwidth}{@{}>{\raggedright\arraybackslash}p{0.18\textwidth}*{3}{>{\raggedright\arraybackslash}X}@{}}
\toprule
Work & Experience source and update mechanism & Retained content & Subsequent use \\
\midrule
DROC \citep{c7droc} & Interpret and extract human corrections & Correction knowledge & Retrieve for planning and skill composition \\
ASPIRE \citep{c7aspire} & Repair programs and consolidate validated fixes & Repair guidance with applicability conditions & Guide future robot programs \\
RATs \citep{c7rats} & Acquire and curate code skills from interaction & Library of skill programs stored as source code & Reuse in program generation and execution; library frozen in tests \\
Scaling Up \citep{c7scalingup} & Guide collection with language; check outcomes; distill & Diffusion policy parameters & Generate robot actions \\
VERITAS \citep{c7veritas} & Select actions with a verifier; train on successful rollouts & Updated action policy parameters & Execute learned behavior \\
BOSS \citep{c7boss} & Practice skill chains with LLM guidance; train with RL & Policy parameters and composite skill descriptions & Execute and further compose learned behaviors \\
CaP-RL \citep{c5capx} & Train from execution rewards using privileged S1 APIs & Coding model parameters & Generate later programs \\
VLAW \citep{c7vlaw} & Adapt models using real and synthetic experience & Updated world model, reward model, and VLA parameters & Generate and select synthetic training data; produce robot actions \\
\bottomrule
\end{tabularx}
\end{table*}


\subsection{Learning Harnesses and Recursive Self-Improvement}
\label{sec:feedback-orchestration}

Section~\ref{sec:feedback-learning} covered how experience changes the guidance, programs, and models used in later behavior.
This section discusses practice selection, trial coordination through learning harnesses, and recursive self-improvement (Figure~\ref{fig:c7-overview}).

A \emph{learning harness} provides the software and execution support for learning experiments. It connects data collection, environment reset, training or code revision, evaluation, and update adoption. The harness can remain fixed, and individual steps may rely on agent decisions, fixed procedures, or human intervention. The examples below cover parts of this workflow or recursive systems, with varying levels of autonomy.

In \emph{recursive self-improvement} (RSI), a system revises its own components and uses them to generate, implement, or evaluate further changes \citep{c7dgm,c7hyperagents}. These components can include agent code, tools, or revision procedures, while foundation-model weights may remain fixed. A fixed training loop may repeatedly update a task policy while leaving the revision process unchanged. Table~\ref{tab:c7-orchestration} compares learning workflows and recursive revision mechanisms.



\begin{table*}[t]
\centering\small
\setlength{\tabcolsep}{4pt}\renewcommand{\arraystretch}{1.12}
\caption{Decisions and dependencies across successive experiments and revisions. The table identifies what the agent decides, the evidence guiding its next choice, and the connection to subsequent rounds. Collection and training workflows are included alongside recursive revision mechanisms.}
\label{tab:c7-orchestration}
\begin{tabularx}{\textwidth}{@{}>{\raggedright\arraybackslash}p{0.18\textwidth}*{3}{>{\raggedright\arraybackslash}X}@{}}
\toprule
Work & Agent decision & Evidence guiding the next choice & Connection to subsequent rounds \\
\midrule
AutoRT \citep{ahn2024autort} & Propose and filter collection tasks & Scene observations and feasibility checks & Further task proposals use the observed scene and available capabilities \\
RoboClaw \citep{li2026roboclawagenticframeworkscalable} & Coordinate collection and reset & Observations and execution feedback & Forward and reset policies support further collection trials \\
Eureka \citep{c7eureka}; DrEureka \citep{c7dreureka} & Revise rewards; configure domain randomization & Policy training results; reward evaluation & Reward revisions and generated configurations guide further training \\
HARBOR \citep{c7harbor} & Choose repairs and training trials & Failed stage checks and trial outcomes & Recorded trial experience informs later tuning choices \\
ENPIRE \citep{c7enpire} & Test policy and algorithm revisions & Physical rollouts and automated verification & Revised implementations are tested through fixed environment APIs \\
EmbodiRSI \citep{c7embodirsi} & Target the next data collection & Failures of the current policy & Updated policies generate evidence for further acquisition \\
COAP \citep{c7coap} & Develop and select code branches & Recorded states and validation episodes & Accepted robot code becomes the next development baseline \\
OpenETA \citep{chen2026eta} & Assess candidates for admission & Contract checks, isolated review, task replay, and held-out paired evaluation & Admission gates baseline changes; no candidate passed all reported criteria \\
PhysicalRSI \citep{physicalrsi2026} & Propose and select harness revisions & Execution evidence and candidate evaluation & The selected harness is inherited by the next agent version \\
SICA \citep{c7sica}; DGM \citep{c7dgm} & Select and modify coding agents & Software task evaluation and agent archives & Modified agents can undertake later modifications \\
HyperAgents \citep{c7hyperagents} & Revise task and meta agents & Task results and histories of generated agents & Updated meta agents produce further agent revisions \\
\bottomrule
\end{tabularx}
\end{table*}

\noindent\textbf{Choosing further practice.}
Agents choose practice tasks based on their current skills and environment. AutoRT \citep{ahn2024autort} proposes and filters physical data collection tasks using scene understanding and available robot capabilities. RATs \citep{c7rats} and BOSS \citep{c7boss} use available skills to guide further practice. BOSS also uses the executed skill chain to guide its next proposals. Section~\ref{sec:feedback-learning} describes the resulting code and policy updates.

Execution failures can guide the next round of practice. EmbodiRSI \citep{c7embodirsi} uses diagnosed failures to guide corrective trajectory generation and configure subsequent data collection. Under a fixed learning formulation, the policy learns from this experience and produces new rollouts that inform the next collection round. Practice selection can therefore draw on feasibility, skill expansion, or observed weaknesses without directly estimating learning progress.

\noindent\textbf{Coordinating trials and revisions.}
A learning harness turns an agent's proposed experiment into an executable trial. For physical experiments, this includes restoring usable initial conditions and recording outcomes for comparison. RoboClaw \citep{li2026roboclawagenticframeworkscalable} coordinates collection through a VLM controller and forward and reset policies. Reset and learning mechanisms such as Leave no Trace \citep{c7leavenotrace}, RoboFuME \citep{c7robofume}, and SERL \citep{c7serl} support repeated trials. These mechanisms prepare the next trial; Section~\ref{sec:feedback-recovery} covers recovery within the current task. Evaluation should distinguish reset failures from policy failures and account for human operation or intervention.

Trial results help agents compare and revise candidate changes. Real2Worlds2Real uses failures in simulated rollouts to guide further data generation and policy refinement, and retains validated procedures for later tasks \citep{gu2026real2worlds2real}. Eureka \citep{c7eureka} uses policy training results to revise reward programs; DrEureka \citep{c7dreureka} also develops domain randomization configurations. HARBOR \citep{c7harbor} coordinates stage commands, training trials, and checks, using failed checks for local repair and trial outcomes to choose further revisions. ENPIRE \citep{c7enpire} establishes reset and verification APIs with human assessment. It then holds these interfaces fixed while coding agents inspect physical rollouts and revise policies, algorithms, and training code.

Failed integration checks may require implementation repairs before a trial can proceed. Poor task performance can instead guide the next learning experiment. HARBOR may require user intervention when its repair budget is exhausted \citep{c7harbor}.

Validation episodes can guide the choice of code revisions. In COAP \citep{c7coap}, coding agents diagnose failures from recorded states, explore alternative branches, and select changes using validation episodes. Accepted code becomes the baseline for further development of task programs and a shared library. The reported experiments are in simulation. This process uses execution evidence to choose revisions; Section~\ref{sec:feedback-learning} covers retaining reusable code.

Candidate updates can be checked before adoption. OpenETA \citep{chen2026eta} uses contract checks, isolated review, replay on the original task, and paired evaluation on held-out cases. No candidate passed all promotion criteria in its reported studies. The results therefore demonstrate the admission procedure, with no evidence of a capability gain from an accepted update.

\noindent\textbf{Recursive self-improvement.}
Recursive self-improvement requires revised components to take part in later self-modification. The PhysicalRSI \citep{physicalrsi2026} project describes changes to an execution harness containing routing, skills, tools, and control code. An agent proposes alternatives, evaluates them, and inherits the selected harness. SHAPER \citep{c7shaper} revises procedural guidance and context construction around frozen planning and execution models. Evaluation must distinguish the effects of these changes on task performance from their role in producing or assessing further modifications.

An updated agent can itself make further changes. SICA \citep{c7sica} and DGM \citep{c7dgm} provide software examples in which modified coding agents can perform later revision attempts. HyperAgents \citep{c7hyperagents} makes both the task agent and the meta agent that generates revisions editable. It compares initial and evolved meta agents on their ability to produce better task agents within a fixed number of attempts, including transfer to another domain. Its robotics experiments concern reward design for a simulated Go2 robot; the main experiments retain a fixed procedure for selecting parent agents.

Robot studies also need to track what changes across revision rounds. RegenHarness \citep{c7regenharness} reports physical deployment cases and proposes an RSI evaluation protocol. Gains across successive improvement rounds still require separate evidence. A study of 123 rounds in RoboCasa \citep{c7rsi123} records local changes without solving its principal fridge task, while model changes and human repairs complicate attribution. Outcomes used to choose a revision should be distinguished from subsequent tests of its effects.

RSI evaluation must assess the recursive mechanism, task performance, and the ability to produce further improvements. Reports should identify inherited changes and their role in later revisions, along with human interventions and changes to the underlying models. Claims of greater improvement ability require comparing the original and revised improvement mechanisms from comparable starting systems. Match interaction and revision budgets, and test on outcomes not used to select candidates (Section~\ref{sec:c8-feedback}).

%% file: sections/sec7_evaluation.tex
\section{Benchmarks and Evaluation}
\label{sec:evaluation}

This section reviews how benchmarks evaluate perception, embodied action, decision-making, and feedback (Secs.~\ref{sec:c8-information}, \ref{sec:c8-execution}, \ref{sec:c8-organization}, \ref{sec:c8-feedback}). We investigate which capabilities each protocol tests and which parts of the task are handled by the evaluation setup. Section~\ref{sec:c8-results} then compares model performance across tasks and examines how configuration changes affect results within individual studies.

The four subsections cover complementary evaluation perspectives, so a benchmark may appear in more than one. For each benchmark, we describe the information and action capabilities available to the system, the decisions it makes, and the outcomes. Table~\ref{tab:c8-benchmark-overview} summarizes representative task benchmarks and evaluations of agentic capabilities across these perspectives.



\begin{table*}[!t]
\centering\small
\setlength{\tabcolsep}{4pt}\renewcommand{\arraystretch}{1.12}
\caption{Representative benchmarks across four functional perspectives.}\label{tab:c8-benchmark-overview}
\begin{tabularx}{\textwidth}{@{}>{\raggedright\arraybackslash}p{0.32\textwidth}*{4}{>{\centering\arraybackslash}X}@{}}
\toprule
Benchmark & Perception and state representation & Action and embodied interaction & Decision-making, planning and coordination & Closed-loop feedback and capability improvement\\
\midrule
RLBench \citep{c8pub_c5rlbench} & \raisebox{-1.4pt}{\includegraphics[width=9.5pt,height=9.5pt]{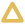}} & \raisebox{-1.4pt}{\includegraphics[width=9.5pt,height=9.5pt]{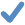}} & \raisebox{-1.4pt}{\includegraphics[width=9.5pt,height=9.5pt]{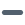}} & \raisebox{-1.4pt}{\includegraphics[width=9.5pt,height=9.5pt]{figures/coverage-dash.pdf}}\\
CALVIN \citep{c8pub_c5calvin} & \raisebox{-1.4pt}{\includegraphics[width=9.5pt,height=9.5pt]{figures/coverage-triangle.pdf}} & \raisebox{-1.4pt}{\includegraphics[width=9.5pt,height=9.5pt]{figures/coverage-check.pdf}} & \raisebox{-1.4pt}{\includegraphics[width=9.5pt,height=9.5pt]{figures/coverage-dash.pdf}} & \raisebox{-1.4pt}{\includegraphics[width=9.5pt,height=9.5pt]{figures/coverage-dash.pdf}}\\
LIBERO \citep{c8pub_c5libero} & \raisebox{-1.4pt}{\includegraphics[width=9.5pt,height=9.5pt]{figures/coverage-triangle.pdf}} & \raisebox{-1.4pt}{\includegraphics[width=9.5pt,height=9.5pt]{figures/coverage-check.pdf}} & \raisebox{-1.4pt}{\includegraphics[width=9.5pt,height=9.5pt]{figures/coverage-triangle.pdf}} & \raisebox{-1.4pt}{\includegraphics[width=9.5pt,height=9.5pt]{figures/coverage-check.pdf}}\\
BEHAVIOR-1K \citep{c5behavior} & \raisebox{-1.4pt}{\includegraphics[width=9.5pt,height=9.5pt]{figures/coverage-triangle.pdf}} & \raisebox{-1.4pt}{\includegraphics[width=9.5pt,height=9.5pt]{figures/coverage-check.pdf}} & \raisebox{-1.4pt}{\includegraphics[width=9.5pt,height=9.5pt]{figures/coverage-triangle.pdf}} & \raisebox{-1.4pt}{\includegraphics[width=9.5pt,height=9.5pt]{figures/coverage-dash.pdf}}\\
RoboDojo \citep{c8robodojo} & \raisebox{-1.4pt}{\includegraphics[width=9.5pt,height=9.5pt]{figures/coverage-check.pdf}} & \raisebox{-1.4pt}{\includegraphics[width=9.5pt,height=9.5pt]{figures/coverage-check.pdf}} & \raisebox{-1.4pt}{\includegraphics[width=9.5pt,height=9.5pt]{figures/coverage-triangle.pdf}} & \raisebox{-1.4pt}{\includegraphics[width=9.5pt,height=9.5pt]{figures/coverage-dash.pdf}}\\
THE COLOSSEUM \citep{c8pub_c5colosseum} & \raisebox{-1.4pt}{\includegraphics[width=9.5pt,height=9.5pt]{figures/coverage-triangle.pdf}} & \raisebox{-1.4pt}{\includegraphics[width=9.5pt,height=9.5pt]{figures/coverage-check.pdf}} & \raisebox{-1.4pt}{\includegraphics[width=9.5pt,height=9.5pt]{figures/coverage-dash.pdf}} & \raisebox{-1.4pt}{\includegraphics[width=9.5pt,height=9.5pt]{figures/coverage-dash.pdf}}\\
OpenEQA \citep{OpenEQA} & \raisebox{-1.4pt}{\includegraphics[width=9.5pt,height=9.5pt]{figures/coverage-check.pdf}} & \raisebox{-1.4pt}{\includegraphics[width=9.5pt,height=9.5pt]{figures/coverage-triangle.pdf}} & \raisebox{-1.4pt}{\includegraphics[width=9.5pt,height=9.5pt]{figures/coverage-triangle.pdf}} & \raisebox{-1.4pt}{\includegraphics[width=9.5pt,height=9.5pt]{figures/coverage-dash.pdf}}\\
GOAT-Bench \citep{c8pub_GOATBench} & \raisebox{-1.4pt}{\includegraphics[width=9.5pt,height=9.5pt]{figures/coverage-check.pdf}} & \raisebox{-1.4pt}{\includegraphics[width=9.5pt,height=9.5pt]{figures/coverage-check.pdf}} & \raisebox{-1.4pt}{\includegraphics[width=9.5pt,height=9.5pt]{figures/coverage-triangle.pdf}} & \raisebox{-1.4pt}{\includegraphics[width=9.5pt,height=9.5pt]{figures/coverage-dash.pdf}}\\
RoboMME \citep{c8pub_RoboMME} & \raisebox{-1.4pt}{\includegraphics[width=9.5pt,height=9.5pt]{figures/coverage-check.pdf}} & \raisebox{-1.4pt}{\includegraphics[width=9.5pt,height=9.5pt]{figures/coverage-check.pdf}} & \raisebox{-1.4pt}{\includegraphics[width=9.5pt,height=9.5pt]{figures/coverage-dash.pdf}} & \raisebox{-1.4pt}{\includegraphics[width=9.5pt,height=9.5pt]{figures/coverage-dash.pdf}}\\
EmbodiedBench \citep{yang2025embodiedbench} & \raisebox{-1.4pt}{\includegraphics[width=9.5pt,height=9.5pt]{figures/coverage-check.pdf}} & \raisebox{-1.4pt}{\includegraphics[width=9.5pt,height=9.5pt]{figures/coverage-check.pdf}} & \raisebox{-1.4pt}{\includegraphics[width=9.5pt,height=9.5pt]{figures/coverage-check.pdf}} & \raisebox{-1.4pt}{\includegraphics[width=9.5pt,height=9.5pt]{figures/coverage-triangle.pdf}}\\
Embodied Agent Interface \citep{li2024eai} & \raisebox{-1.4pt}{\includegraphics[width=9.5pt,height=9.5pt]{figures/coverage-triangle.pdf}} & \raisebox{-1.4pt}{\includegraphics[width=9.5pt,height=9.5pt]{figures/coverage-dash.pdf}} & \raisebox{-1.4pt}{\includegraphics[width=9.5pt,height=9.5pt]{figures/coverage-check.pdf}} & \raisebox{-1.4pt}{\includegraphics[width=9.5pt,height=9.5pt]{figures/coverage-dash.pdf}}\\
PARTNR \citep{chang2025partnr} & \raisebox{-1.4pt}{\includegraphics[width=9.5pt,height=9.5pt]{figures/coverage-check.pdf}} & \raisebox{-1.4pt}{\includegraphics[width=9.5pt,height=9.5pt]{figures/coverage-check.pdf}} & \raisebox{-1.4pt}{\includegraphics[width=9.5pt,height=9.5pt]{figures/coverage-check.pdf}} & \raisebox{-1.4pt}{\includegraphics[width=9.5pt,height=9.5pt]{figures/coverage-triangle.pdf}}\\
RoCoBench \citep{mandi2024roco} & \raisebox{-1.4pt}{\includegraphics[width=9.5pt,height=9.5pt]{figures/coverage-check.pdf}} & \raisebox{-1.4pt}{\includegraphics[width=9.5pt,height=9.5pt]{figures/coverage-check.pdf}} & \raisebox{-1.4pt}{\includegraphics[width=9.5pt,height=9.5pt]{figures/coverage-check.pdf}} & \raisebox{-1.4pt}{\includegraphics[width=9.5pt,height=9.5pt]{figures/coverage-triangle.pdf}}\\
CaP-Bench \citep{c5capx} & \raisebox{-1.4pt}{\includegraphics[width=9.5pt,height=9.5pt]{figures/coverage-triangle.pdf}} & \raisebox{-1.4pt}{\includegraphics[width=9.5pt,height=9.5pt]{figures/coverage-check.pdf}} & \raisebox{-1.4pt}{\includegraphics[width=9.5pt,height=9.5pt]{figures/coverage-triangle.pdf}} & \raisebox{-1.4pt}{\includegraphics[width=9.5pt,height=9.5pt]{figures/coverage-triangle.pdf}}\\
RobotWorld \citep{c8robotworld} & \raisebox{-1.4pt}{\includegraphics[width=9.5pt,height=9.5pt]{figures/coverage-triangle.pdf}} & \raisebox{-1.4pt}{\includegraphics[width=9.5pt,height=9.5pt]{figures/coverage-check.pdf}} & \raisebox{-1.4pt}{\includegraphics[width=9.5pt,height=9.5pt]{figures/coverage-triangle.pdf}} & \raisebox{-1.4pt}{\includegraphics[width=9.5pt,height=9.5pt]{figures/coverage-triangle.pdf}}\\
IS-Bench \citep{c8isbench} & \raisebox{-1.4pt}{\includegraphics[width=9.5pt,height=9.5pt]{figures/coverage-check.pdf}} & \raisebox{-1.4pt}{\includegraphics[width=9.5pt,height=9.5pt]{figures/coverage-check.pdf}} & \raisebox{-1.4pt}{\includegraphics[width=9.5pt,height=9.5pt]{figures/coverage-check.pdf}} & \raisebox{-1.4pt}{\includegraphics[width=9.5pt,height=9.5pt]{figures/coverage-check.pdf}}\\
REPAIR-Bench \citep{c8repair} & \raisebox{-1.4pt}{\includegraphics[width=9.5pt,height=9.5pt]{figures/coverage-check.pdf}} & \raisebox{-1.4pt}{\includegraphics[width=9.5pt,height=9.5pt]{figures/coverage-dash.pdf}} & \raisebox{-1.4pt}{\includegraphics[width=9.5pt,height=9.5pt]{figures/coverage-triangle.pdf}} & \raisebox{-1.4pt}{\includegraphics[width=9.5pt,height=9.5pt]{figures/coverage-check.pdf}}\\
LIBERO-RECOVER \citep{c7recover} & \raisebox{-1.4pt}{\includegraphics[width=9.5pt,height=9.5pt]{figures/coverage-triangle.pdf}} & \raisebox{-1.4pt}{\includegraphics[width=9.5pt,height=9.5pt]{figures/coverage-check.pdf}} & \raisebox{-1.4pt}{\includegraphics[width=9.5pt,height=9.5pt]{figures/coverage-dash.pdf}} & \raisebox{-1.4pt}{\includegraphics[width=9.5pt,height=9.5pt]{figures/coverage-check.pdf}}\\
Embodied First Exam (EFE) \citep{c8efe} & \raisebox{-1.4pt}{\includegraphics[width=9.5pt,height=9.5pt]{figures/coverage-check.pdf}} & \raisebox{-1.4pt}{\includegraphics[width=9.5pt,height=9.5pt]{figures/coverage-check.pdf}} & \raisebox{-1.4pt}{\includegraphics[width=9.5pt,height=9.5pt]{figures/coverage-check.pdf}} & \raisebox{-1.4pt}{\includegraphics[width=9.5pt,height=9.5pt]{figures/coverage-check.pdf}}\\
\bottomrule
\end{tabularx}
\par\smallskip\begin{minipage}{\textwidth}\footnotesize
\raisebox{-1.1pt}{\includegraphics[width=7.5pt,height=7.5pt]{figures/coverage-check.pdf}}: explicit evaluation of relevant capabilities; \raisebox{-1.1pt}{\includegraphics[width=7.5pt,height=7.5pt]{figures/coverage-triangle.pdf}}: supporting or indirect coverage grounded in task design or interaction; \raisebox{-1.1pt}{\includegraphics[width=7.5pt,height=7.5pt]{figures/coverage-dash.pdf}}: no explicit coverage in the reviewed scope. Symbols summarize benchmark-level coverage under this survey's framework, not exhaustive assessment of every capability within a perspective. Coverage within a row may come from different tasks or evaluation settings; LIBERO includes lifelong learning evaluation, and EFE is considered at suite level. The selection is representative rather than exhaustive.
\end{minipage}
\end{table*}

\subsection{Perception and State Representation}
\label{sec:c8-information}

Perception benchmarks differ in whether observations are supplied to the system or the system must choose how to obtain them. This difference affects what task performance can reveal about perception. ScanQA \citep{azuma2022scanqa}, SQA3D \citep{c8pub_ma2022sqa3d}, and OpenEQA's episodic-memory setting \citep{OpenEQA} assess answers from supplied scenes or histories, whereas OpenEQA's active setting and ESI-Bench \citep{c8esi} expose observation or action choices before answering. Answer quality in the latter settings must be interpreted alongside exploration budgets. R2R \citep{c8r2r} and REVERIE \citep{c8reverie} connect observation use to navigation and target grounding through simulated interaction with real-scene imagery. Together, these settings distinguish interpreting evidence from obtaining and using it during behavior.

Other benchmarks test whether information obtained earlier in an episode supports subsequent decisions. GOAT-Bench \citep{c8pub_GOATBench} provides successive navigation goals; RoboMME \citep{c8pub_RoboMME} separates counting, object permanence, reference resolution, and imitation; and MIKASA-Robo-VLA \citep{c8mikasavla} compares tasks with hidden cues against control conditions with visible cues. RMBench \citep{RMBench}, MEMOBench \citep{MEMOBench}, and RoboMemArena \citep{RoboMemArena} provide complementary tests of memory demands, behavior at the level of individual operations, and stage completion. MEMOBench's operation rates use simulator checks rather than inspection of internal memory, while RoboMemArena also includes physical evaluation.

For the information mechanisms reviewed in Section~\ref{sec:perception}, the relevant question is whether acquired or retained information changes subsequent choices when action interfaces and available execution capabilities are comparable. Long episodes alone need not require memory. History access, action chunking, and success timing can change the interpretation: MIKASA's released protocol credits success reached at any episode step, which differs from requiring success at termination \citep{c8mikasavla}. Task outcomes measure performance under these conditions; identifying the contribution of information requires matched comparisons that vary its availability or retention. Intermediate checks help separate stale or missing information from execution failure.

\begin{table*}[!t]
\centering\small
\setlength{\tabcolsep}{5pt}\renewcommand{\arraystretch}{1.16}
\caption{Interaction conditions in EFE's shared task protocol \citep{c8efe}. The modes specify evaluation conditions rather than levels of agentic capability.}\label{tab:c8-efe-protocols}
\begin{tabularx}{\textwidth}{@{}>{\raggedright\arraybackslash}p{0.14\textwidth}>{\raggedright\arraybackslash}X>{\raggedright\arraybackslash}p{0.13\textwidth}>{\raggedright\arraybackslash}X>{\raggedright\arraybackslash}X@{}}
\toprule
Mode & Observation access & Reset allowance & World progression & Grading evidence\\
\midrule
Privileged & Full simulator state & Unrestricted & Agent-controlled simulation and planning & Independent replay of submitted trajectory or policy\\
\addlinespace[3pt]
Standard & Robot observations and declared sensors; no privileged scene state & Task-declared budget & Advances during action or timed-observation calls & Trusted record of the live episode\\
\addlinespace[3pt]
Standard-\newline realtime & Same observation boundary as standard & Task-declared budget & Continues at $1/S$ speed after execution starts, including during deliberation & Trusted record of the live episode\\
\bottomrule
\end{tabularx}
\par\smallskip\begin{minipage}{\textwidth}\footnotesize
All modes are subject to a limit on elapsed real time specified by the task; $S$ is the declared slowdown factor. Mode availability is task-specific. Benchmarks retaining a custom runtime are outside this summary of the shared protocol and are reported separately.
\end{minipage}
\end{table*}

\subsection{Action and Embodied Interaction}
\label{sec:c8-execution}

This subsection examines physical task performance under specified observation and control interfaces, with attention to the behavior required and the execution support provided. RLBench \citep{c8pub_c5rlbench} and ManiSkill3 \citep{c8pub_c5maniskill3} allow configured observations, action modes, or controllers; CALVIN \citep{c8pub_c5calvin} supplies successive language goals, while VIMA-Bench \citep{c8pub_c5vima} varies multimodal task specifications and generalization splits. Their task success, chain completion, or results for each evaluation split therefore describe different execution demands. Identifying the supplied goals, primitives, and controller is essential to understanding what the system itself must select and realize.

Task families expose complementary physical demands. RoboCasa \citep{c8pub_c5robocasa}, RoboCasa365 \citep{c8pub_c5robocasa365} and RoboTwin 2.0 \citep{c5robotwin2} cover household and dual-arm manipulation, while ClevrSkills \citep{c8pub_c5clevrskills} tests compositional manipulation and BEHAVIOR-1K \citep{c5behavior} provides a broad activity specification whose evaluated subset must be stated. RoboDojo \citep{c8robodojo} organizes tests around different capability requirements; DuoBench \citep{c8duobench} and RoboEval \citep{c8roboeval} assess bimanual execution and progress. HomeRobot OVMM \citep{c8homerobot} and BiGym \citep{c8bigym} add mobile manipulation, and DexArt \citep{c8dexart}, FurnitureBench \citep{c8furniture}, and FMB \citep{c8fmb} introduce dexterity or the composition of skills involving extensive physical contact. Stage and skill completion can reveal progress hidden by binary task success. Execution support remains consequential: simulated grasp attachment in HomeRobot or fixed support for the lower body in BiGym changes the control problem. RobotWorld~\citep{c8robotworld} extends the comparison across manipulation, locomotion, driving, and aerial control with task-specific interfaces and budgets.

Embodied First Exam (EFE) extends this analysis to coding agents that write and execute robot programs in frozen simulation tasks \citep{c8efe}. Table~\ref{tab:c8-efe-protocols} summarizes three interaction modes, distinguishing access to simulator state, reset opportunities, world progression, and the evidence used for grading \citep{c8efe}.

These settings connect the interfaces discussed in Section~\ref{sec:interfaces} and the scheduling mechanisms in Section~\ref{sec:handoffs} to measurable execution demands. Observation access, reset opportunities, and timing define the conditions of these mode comparisons.

Robustness and physical validation test whether these capabilities remain useful under changed conditions. THE COLOSSEUM \citep{c8pub_c5colosseum} , LIBERO-PRO \citep{c5liberopro} , and LIBERO-Plus \citep{c8liberoplus}   report performance under specified perturbations; observing the resulting choices is additionally needed to identify adaptive capability selection. SIMPLER \citep{c8pub_c5simpler} and REALM \citep{c8realm} assess simulation as a proxy for physical evaluation, while RoboArena \citep{c8roboarena} ranks paired physical executions by human preference. Proxy agreement, preference, and absolute success provide different evidence. Physical comparisons should preserve trial counts, uncertainty, timing, retries, safety stops, and human assistance; sim--real differences require matched tasks, policies, interfaces, and scoring rules. These conditions connect execution scores to the capability choices discussed in Section~\ref{sec:actionSpaces}.

\subsection{Decision-Making, Planning and Coordination}
\label{sec:c8-organization}

This subsection examines how systems select and organize behavior across task steps and, where applicable, across participants. Planning evaluation depends on how much task structure a system constructs; coordination evaluation additionally considers dependencies between decision and execution levels or between participants. Embodied Agent Interface \citep{li2024eai} separates goal interpretation, decomposition, sequencing, and transition modeling, while EmbodiedBench \citep{yang2025embodiedbench} varies the support provided through skills and atomic actions. ALFRED \citep{c8pub_c5alfred} distinguishes instructions that specify goals from those that specify individual steps, and ALFWorld \citep{c8alfworld} separates text interaction from embodied transfer. EmbodiedEval \citep{c8embodiedeval} and VLABench \citep{c8pub_c5vlabench} broaden the task and reasoning settings. Embodied Playground~\citep{c8embodiedplayground} evaluates long-horizon planning through simulated robot tasks whose success requires goal completion, rule compliance, and physical validity. Module correctness and plan agreement locate intermediate capabilities; executed completion shows whether they suffice with the available executor.

Adaptive organization also requires revising choices as conditions change. CaP-Bench \citep{c5capx} varies state access, primitive abstraction, and coding with feedback; DynAfford \citep{c8dynafford} changes preconditions within supplied goals and subgoals. Assessing the quality of the initial plan, task completion after conditions change, and the cost of revising the plan separately makes this adaptation observable. WorldArena \citep{shang2026worldarena} evaluates world models through policy evaluation and planning, while RoboWM-Bench \citep{c8robowm} connects predicted manipulation to simulated execution through video-to-action conversion. These downstream outcomes assess prediction together with its grounding pipeline; visual plausibility alone cannot establish usefulness for action. RoboCerebra~\citep{c8robocerebra} similarly distinguishes agreement with a reference plan from completion of key state transitions during manipulation. Section~\ref{sec:c8-results} covers configuration comparisons that connect such organization to execution outcomes.

Execution scheduling concerns when control returns to the decision level and whether the current behavior should continue or end. Evaluation should relate these choices to task completion, decision delays and unnecessary interruptions. EFE's interaction modes (Table~\ref{tab:c8-efe-protocols}) specify whether the world advances during deliberation, making timing part of the execution conditions. Section~\ref{sec:c8-results} compares
the evaluated configurations.

Multi-robot and human--robot coordination introduces dependencies on other participants. PARTNR \citep{chang2025partnr} and RoCoBench \citep{mandi2024roco} evaluate collaboration and coordination in a shared workspace, and Habitat-MAS \citep{chen2025emos} emphasizes allocation across heterogeneous embodiments. TEACh \citep{c8pub_c5teach} distinguishes supplied dialogue history from interactive completion; CVDN's Navigation from Dialog History \citep{c8cvdn} evaluates use of prior assistance; and NIABench \citep{c8nia} tests action selection and timing alongside a scripted human sequence. Team success must consequently be interpreted with partner capabilities, information access, and supplied plans. Communication costs and progress measures complement completion, while comparisons with controlled partner conditions help distinguish coordination benefits from stronger individual capabilities.

\subsection{Closed-Loop Feedback and Capability Improvement}
\label{sec:c8-feedback}

Feedback evaluation distinguishes judging an outcome, executing a correction, and retaining changes for future tasks. RoboVQA \citep{c8pub_RoboVQA} and RoboRMBench \citep{c7robormbench} assess recorded activities or reward judgments, while REPAIR-Bench \citep{c8repair} adds failure detection and prediction of preferences among recovery strategies. REPAIR's recovery prediction receives a failure label and candidate strategies, so its preference agreement measures strategy selection rather than executed repair. SafeManip \citep{c8safemanip} and IS-Bench \citep{c8isbench} connect safety to simulated execution through grounded constraints and risk-mitigation checks; supplied reminders and evaluator information must be distinguished from the robot's own monitoring. EFE's Embodied Safety Exam includes paired hazardous and benign requests and reports safety judgment, incident rates, and task capability separately \citep{c8efe}, so that low incident rates can be interpreted alongside the ability to act.

Recovery benchmarks make correction observable, but differ in how it begins. LIBERO-RECOVER \citep{c7recover} and RoboRecover \citep{c8roborecover} test continuation from supplied failure or deviation states, leaving failure detection outside that comparison. PLanAR \citep{c8agenticlab} compares checker configurations under physical disturbances, while CheckVLA \citep{c7checkvla} compares periodic replanning, guidance triggered by verification, and memory. These settings distinguish recovery from a supplied state from intervention initiated by the system; task outcomes must be interpreted alongside the detection process and external support.

Persistent improvement requires evidence beyond correction within one episode. LIBERO \citep{c8pub_c5libero}, Continual World \citep{c8continualworld}, and RoboCasa365's lifelong setting \citep{c8pub_c5robocasa365} assess learning and retention across task sequences, with outcomes conditioned on task order, replay access, and update budgets. Connecting these results to the learning mechanisms reviewed in Section~\ref{sec:feedback} requires distinguishing retained task gains from the agent's role in selecting experience, initiating updates, and adopting changes.

Development feedback used to revise or select an update must be distinguished from final testing. The selected version and any permitted adaptation procedure should be fixed before testing on tasks or instances not used for candidate selection; feedback allowed within that fixed test protocol remains available. An isolated outcome evaluator does not by itself make a score independent of development.

EFE's roboEngineer provides a concrete separation within a task involving controller adaptation \citep{c8efe}. An agent develops a controller in a public nominal simulator, receives feedback from at most three deployments to a hidden dynamical system, and selects a policy to be evaluated on private episodes of the same system reserved for final testing. This separates adaptation feedback from final assessment, while testing new episodes rather than transfer to unseen dynamical systems.

For the recursive improvement discussed in Section~\ref{sec:feedback-orchestration}, later task gains and improved ability to produce further gains require separate evidence. Section~\ref{sec:discussion-evaluation} develops the corresponding evaluation requirements.

\subsection{Comparative Results and Evaluation Gaps}
\label{sec:c8-results}

We first compare performance across models and task demands within each benchmark, then examine configuration differences within individual studies. These comparisons relate reported outcomes to the execution capabilities and support available to each system. The synthesis assesses what this evidence establishes about autonomous execution and capability improvement.
The selected reports provide explicit task breakdowns or executor comparisons; the RoboDojo board supplies a dated cross-system snapshot. All values are reported by the original sources. A row represents its complete evaluated configuration, including the checkpoint, tools, history, and budget. Comparisons remain within the original protocols; unreported entries are not zeros, and cross-system scores do not isolate component effects. Versions, source tables, and the board's 29 October 2026 retrieval date accompany the results.

\subsubsection{Performance across Models and Tasks}
Performance varies across both models and task demands. Cross-domain execution, capability-specific system profiles, and comparisons between planning and manipulation reveal differences that aggregate scores can obscure.

\begin{table*}[!t]
\centering\small
\setlength{\tabcolsep}{3.2pt}\renewcommand{\arraystretch}{1.15}
\caption{RobotWorld: execution across task domains, token use, estimated API costs and cumulative elapsed time. Outcomes are successful tasks / evaluated tasks; Overall also reports the success percentage.}\label{tab:c8-crossstudy}\label{tab:c8-system-profiles}
\begin{tabular}{@{}lrrrrrrrrr@{}}
\toprule
Model & Manip. & Mobile manip. & Locomotion & Driving & Aerial & Overall & \shortstack{Total\\tokens} & \shortstack{Est. total API\\cost (USD)} & \shortstack{Cumulative\\elapsed time (h)}\\
\midrule
GPT-6 Astra &9/38&4/20&0/11&2/11&1/4&16/84 (19.0\%)&941M&9,912.80&54.6\\
Opus 5.5 &7/38&0/20&1/11&2/11&3/4&13/84 (15.5\%)&1.63B&2,916.43&140.3\\
Kimi K3 &0/38&1/20&0/11&1/11&0/4&2/84 (2.4\%)&2.00B$^{\dagger}$&624.09$^{\dagger}$&163.8\\
DeepSeek V4.1 Flash &0/38&0/20&0/11&0/11&1/4&1/84 (1.2\%)&907M&34.49&35.7\\
Gemini 3.8 Flash &1/38&0/20&0/11&0/11&0/4&1/84 (1.2\%)&303M&66.68&28.7\\
\bottomrule
\end{tabular}
\par\smallskip\begin{minipage}{\textwidth}\footnotesize
RobotWorld v1, Tables 3 and 14 \citep{c8robotworld}: one retained episode per model and task, direct robot-tool calls, and task-specific interfaces and budgets. Non-action budget tiers were calibrated from Astra records; some outcomes were adjudicated at budget boundaries. M/B denote million/billion tokens; totals include input (with cached input) and output, without repeated cumulative updates. Costs use reported list prices, not invoices. $^{\dagger}$Kimi token and cost totals extrapolate 57 valid records to 84 tasks; other models have 84. Missing or zero-flagged usage is excluded from means. Resources include failed runs and may extend beyond retrospective scoring boundaries. Time sums all 84 recorded durations, including setup, waiting and execution, rather than the duration of a parallel campaign.
\end{minipage}
\end{table*}

\begin{table*}[!t]
\centering\footnotesize
\definecolor{dojoAgent}{HTML}{245EA8}
\definecolor{dojoAgentWam}{HTML}{7043A0}
\definecolor{dojoAgentVla}{HTML}{A04E14}
\definecolor{dojoVla}{HTML}{1F7560}
\definecolor{dojoWam}{HTML}{8C4D66}
\definecolor{dojoVlaWam}{HTML}{6A6518}
\definecolor{dojoVlmWam}{HTML}{385B75}
\definecolor{dojoSR}{HTML}{558BC4}
\setlength{\tabcolsep}{2pt}\setlength{\fboxsep}{2pt}\renewcommand{\arraystretch}{1.22}
\caption{RoboDojo-Sim: system profiles ordered by Average Score. Each task group retains Score and SR (\%); blue shading encodes Score in all six groups on a shared 0--100 scale.}\label{tab:c8-robodojo-profiles}
\begin{tabular}{@{}ll@{\hspace{6pt}}r@{\hspace{3pt}}r@{\hspace{18pt}}r@{\hspace{3pt}}r@{\hspace{18pt}}r@{\hspace{3pt}}r@{\hspace{18pt}}r@{\hspace{3pt}}r@{\hspace{18pt}}r@{\hspace{3pt}}r@{\hspace{18pt}}r@{\hspace{3pt}}r@{}}\toprule
 & & \multicolumn{2}{c}{Average} & \multicolumn{2}{c}{General.} & \multicolumn{2}{c}{Precision} & \multicolumn{2}{c}{Long} & \multicolumn{2}{c}{Memory} & \multicolumn{2}{c}{Open}\\
\cmidrule(lr){3-4}\cmidrule(lr){5-6}\cmidrule(lr){7-8}\cmidrule(lr){9-10}\cmidrule(lr){11-12}\cmidrule(lr){13-14}
System & Type & Score & SR & Score & SR & Score & SR & Score & SR & Score & SR & Score & SR\\\midrule
\raisebox{-0.8mm}{\includegraphics[width=3.5mm,height=3.5mm]{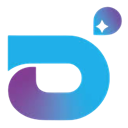}}\hspace{2pt}GO-3 Nano Preview & \colorbox{dojoAgentWam!12}{\textcolor{dojoAgentWam}{\scriptsize Agent + WAM}} & \cellcolor{dojoSR!44.5680}61.90 & 56.42 & \cellcolor{dojoSR!26.2440}36.45 & 33.66 & \cellcolor{dojoSR!56.6280}78.65 & 76.00 & \cellcolor{dojoSR!49.1112}68.21 & 53.50 & \cellcolor{dojoSR!46.4400}64.50 & 63.67 & \cellcolor{dojoSR!44.4312}61.71 & 55.25\\
\raisebox{-0.8mm}{\includegraphics[width=3.5mm,height=3.5mm]{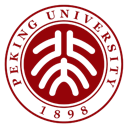}}\hspace{2pt}PyRUA-Lean & \colorbox{dojoAgent!12}{\textcolor{dojoAgent}{\scriptsize Agent}} & \cellcolor{dojoSR!39.7872}55.26 & 50.35 & \cellcolor{dojoSR!34.7544}48.27 & 43.17 & \cellcolor{dojoSR!32.2344}44.77 & 38.25 & \cellcolor{dojoSR!39.0960}54.30 & 43.75 & \cellcolor{dojoSR!49.2480}68.40 & 68.33 & \cellcolor{dojoSR!43.6032}60.56 & 58.25\\
\raisebox{-0.8mm}{\includegraphics[width=3.5mm,height=3.5mm]{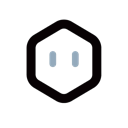}}\hspace{2pt}HexaShell-Preview & \colorbox{dojoAgent!12}{\textcolor{dojoAgent}{\scriptsize Agent}} & \cellcolor{dojoSR!38.2896}53.18 & 46.85 & \cellcolor{dojoSR!28.2384}39.22 & 34.50 & \cellcolor{dojoSR!29.7000}41.25 & 32.00 & \cellcolor{dojoSR!30.8952}42.91 & 31.50 & \cellcolor{dojoSR!60.8616}84.53 & 81.00 & \cellcolor{dojoSR!41.7384}57.97 & 55.25\\
\raisebox{-0.8mm}{\includegraphics[width=3.5mm,height=3.5mm]{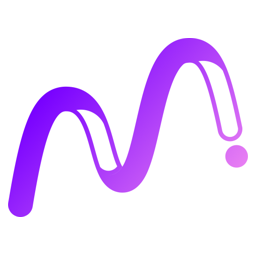}}\hspace{2pt}WorldScape-R$^{2}$SI & \colorbox{dojoAgentWam!12}{\textcolor{dojoAgentWam}{\scriptsize Agent + WAM}} & \cellcolor{dojoSR!33.5376}46.58 & 39.35 & \cellcolor{dojoSR!30.9240}42.95 & 35.83 & \cellcolor{dojoSR!16.9272}23.51 & 15.50 & \cellcolor{dojoSR!37.0656}51.48 & 34.25 & \cellcolor{dojoSR!47.3616}65.78 & 63.67 & \cellcolor{dojoSR!35.3952}49.16 & 47.50\\
\raisebox{-0.8mm}{\includegraphics[width=3.5mm,height=3.5mm]{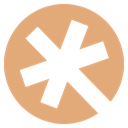}}\hspace{2pt}EmbodiedEvo & \colorbox{dojoAgent!12}{\textcolor{dojoAgent}{\scriptsize Agent}} & \cellcolor{dojoSR!32.4576}45.08 & 38.95 & \cellcolor{dojoSR!25.3584}35.22 & 30.83 & \cellcolor{dojoSR!18.1080}25.15 & 16.25 & \cellcolor{dojoSR!24.3576}33.83 & 21.75 & \cellcolor{dojoSR!57.1680}79.40 & 76.67 & \cellcolor{dojoSR!37.3032}51.81 & 49.25\\
\raisebox{-0.8mm}{\includegraphics[width=3.5mm,height=3.5mm]{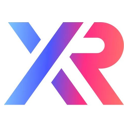}}\hspace{2pt}WALL-OSS-1.0-lite & \colorbox{dojoVla!12}{\textcolor{dojoVla}{\scriptsize VLA}} & \cellcolor{dojoSR!30.6648}42.59 & 34.97 & \cellcolor{dojoSR!33.5952}46.66 & 37.94 & \cellcolor{dojoSR!27.4752}38.16 & 29.08 & \cellcolor{dojoSR!35.9568}49.94 & 32.25 & \cellcolor{dojoSR!41.3424}57.42 & 56.00 & \cellcolor{dojoSR!14.9472}20.76 & 19.58\\
\raisebox{-0.8mm}{\includegraphics[width=3.5mm,height=3.5mm]{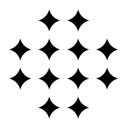}}\hspace{2pt}VPP2 & \colorbox{dojoWam!12}{\textcolor{dojoWam}{\scriptsize WAM}} & \cellcolor{dojoSR!28.2672}39.26 & 32.26 & \cellcolor{dojoSR!26.1936}36.38 & 29.50 & \cellcolor{dojoSR!30.9816}43.03 & 35.17 & \cellcolor{dojoSR!40.5216}56.28 & 37.00 & \cellcolor{dojoSR!40.3272}56.01 & 55.56 & \cellcolor{dojoSR!3.3192}4.61 & 4.08\\
\raisebox{-0.8mm}{\includegraphics[width=3.5mm,height=3.5mm]{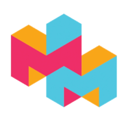}}\hspace{2pt}PhysicalRSI & \colorbox{dojoAgentVla!12}{\textcolor{dojoAgentVla}{\scriptsize Agent + VLA}} & \cellcolor{dojoSR!26.1144}36.27 & 31.38 & \cellcolor{dojoSR!15.3360}21.30 & 15.62 & \cellcolor{dojoSR!27.3960}38.05 & 32.50 & \cellcolor{dojoSR!33.3864}46.37 & 37.33 & \cellcolor{dojoSR!33.6528}46.74 & 46.56 & \cellcolor{dojoSR!20.7936}28.88 & 24.92\\
\raisebox{-0.8mm}{\includegraphics[width=3.5mm,height=3.5mm]{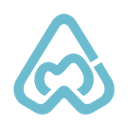}}\hspace{2pt}Awomo-0.5 & \colorbox{dojoWam!12}{\textcolor{dojoWam}{\scriptsize WAM}} & \cellcolor{dojoSR!25.4448}35.34 & 29.64 & \cellcolor{dojoSR!21.8232}30.31 & 23.84 & \cellcolor{dojoSR!28.5480}39.65 & 33.50 & \cellcolor{dojoSR!29.0592}40.36 & 26.75 & \cellcolor{dojoSR!30.4344}42.27 & 41.22 & \cellcolor{dojoSR!17.3448}24.09 & 22.92\\
\raisebox{-0.8mm}{\includegraphics[width=3.5mm,height=3.5mm]{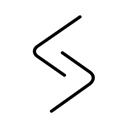}}\hspace{2pt}Simate-beta & \colorbox{dojoVla!12}{\textcolor{dojoVla}{\scriptsize VLA}} & \cellcolor{dojoSR!24.4440}33.95 & 27.96 & \cellcolor{dojoSR!25.2648}35.09 & 27.95 & \cellcolor{dojoSR!24.7320}34.35 & 26.92 & \cellcolor{dojoSR!41.6448}57.84 & 43.42 & \cellcolor{dojoSR!23.9976}33.33 & 33.00 & \cellcolor{dojoSR!6.5664}9.12 & 8.50\\
\raisebox{-0.8mm}{\includegraphics[width=3.5mm,height=3.5mm]{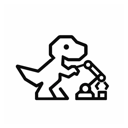}}\hspace{2pt}Rex-M1 Preview & \colorbox{dojoVla!12}{\textcolor{dojoVla}{\scriptsize VLA}} & \cellcolor{dojoSR!24.3288}33.79 & 27.44 & \cellcolor{dojoSR!24.5160}34.05 & 26.05 & \cellcolor{dojoSR!23.6304}32.82 & 23.83 & \cellcolor{dojoSR!33.7608}46.89 & 33.00 & \cellcolor{dojoSR!34.9416}48.53 & 48.22 & \cellcolor{dojoSR!4.8024}6.67 & 6.08\\
\raisebox{-0.8mm}{\includegraphics[width=3.5mm,height=3.5mm]{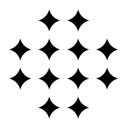}}\hspace{2pt}VPP2-Preview & \colorbox{dojoWam!12}{\textcolor{dojoWam}{\scriptsize WAM}} & \cellcolor{dojoSR!22.6080}31.40 & 25.62 & \cellcolor{dojoSR!22.5648}31.34 & 24.62 & \cellcolor{dojoSR!24.3000}33.75 & 25.58 & \cellcolor{dojoSR!25.4232}35.31 & 22.17 & \cellcolor{dojoSR!36.5616}50.78 & 50.33 & \cellcolor{dojoSR!4.1976}5.83 & 5.42\\
\raisebox{-0.8mm}{\includegraphics[width=3.5mm,height=3.5mm]{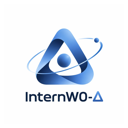}}\hspace{2pt}InternW0-$\Delta$ & \colorbox{dojoWam!12}{\textcolor{dojoWam}{\scriptsize WAM}} & \cellcolor{dojoSR!22.1544}30.77 & 23.91 & \cellcolor{dojoSR!21.6648}30.09 & 22.78 & \cellcolor{dojoSR!23.0256}31.98 & 23.25 & \cellcolor{dojoSR!33.3288}46.29 & 29.33 & \cellcolor{dojoSR!24.9624}34.67 & 34.00 & \cellcolor{dojoSR!7.8048}10.84 & 10.17\\
\raisebox{-0.8mm}{\includegraphics[width=3.5mm,height=3.5mm]{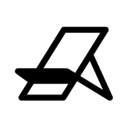}}\hspace{2pt}Liber-0 Preview & \colorbox{dojoVlmWam!12}{\textcolor{dojoVlmWam}{\scriptsize VLM + WAM}} & \cellcolor{dojoSR!22.1328}30.74 & 25.52 & \cellcolor{dojoSR!17.9928}24.99 & 18.61 & \cellcolor{dojoSR!27.5616}38.28 & 33.17 & \cellcolor{dojoSR!33.1056}45.98 & 33.17 & \cellcolor{dojoSR!27.1944}37.77 & 37.33 & \cellcolor{dojoSR!4.8096}6.68 & 5.33\\
\raisebox{-0.8mm}{\includegraphics[width=3.5mm,height=3.5mm]{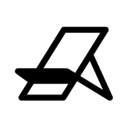}}\hspace{2pt}Liber-0 Lite & \colorbox{dojoWam!12}{\textcolor{dojoWam}{\scriptsize WAM}} & \cellcolor{dojoSR!21.0528}29.24 & 24.23 & \cellcolor{dojoSR!18.2592}25.36 & 19.11 & \cellcolor{dojoSR!26.3520}36.60 & 31.08 & \cellcolor{dojoSR!32.7600}45.50 & 32.92 & \cellcolor{dojoSR!25.6896}35.68 & 35.44 & \cellcolor{dojoSR!2.2032}3.06 & 2.58\\
\raisebox{-0.8mm}{\includegraphics[width=3.5mm,height=3.5mm]{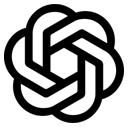}}\hspace{2pt}GPT-6-Astra & \colorbox{dojoAgent!12}{\textcolor{dojoAgent}{\scriptsize Agent}} & \cellcolor{dojoSR!20.8584}28.97 & 22.48 & \cellcolor{dojoSR!24.0192}33.36 & 30.50 & \cellcolor{dojoSR!9.1080}12.65 & 4.00 & \cellcolor{dojoSR!15.4440}21.45 & 8.25 & \cellcolor{dojoSR!30.9888}43.04 & 38.67 & \cellcolor{dojoSR!24.7392}34.36 & 31.00\\
\raisebox{-0.8mm}{\includegraphics[width=3.5mm,height=3.5mm]{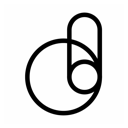}}\hspace{2pt}DM0.5 & \colorbox{dojoVla!12}{\textcolor{dojoVla}{\scriptsize VLA}} & \cellcolor{dojoSR!17.9280}24.90 & 19.34 & \cellcolor{dojoSR!11.3544}15.77 & 10.95 & \cellcolor{dojoSR!17.8704}24.82 & 16.75 & \cellcolor{dojoSR!24.2640}33.70 & 19.50 & \cellcolor{dojoSR!34.3728}47.74 & 47.44 & \cellcolor{dojoSR!1.7496}2.43 & 2.08\\
\raisebox{-0.8mm}{\includegraphics[width=3.5mm,height=3.5mm]{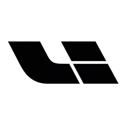}}\hspace{2pt}ME-Brain-1.0 & \colorbox{dojoVlaWam!12}{\textcolor{dojoVlaWam}{\scriptsize VLA + WAM}} & \cellcolor{dojoSR!15.6024}21.67 & 15.99 & \cellcolor{dojoSR!15.7032}21.81 & 15.34 & \cellcolor{dojoSR!17.4456}24.23 & 15.92 & \cellcolor{dojoSR!23.7312}32.96 & 20.58 & \cellcolor{dojoSR!16.9416}23.53 & 22.78 & \cellcolor{dojoSR!4.1976}5.83 & 5.33\\
\raisebox{-0.8mm}{\includegraphics[width=3.5mm,height=3.5mm]{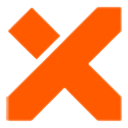}}\hspace{2pt}GalaxeaVLA (G0.5) & \colorbox{dojoVla!12}{\textcolor{dojoVla}{\scriptsize VLA}} & \cellcolor{dojoSR!14.5656}20.23 & 14.88 & \cellcolor{dojoSR!13.2912}18.46 & 12.83 & \cellcolor{dojoSR!20.3400}28.25 & 20.42 & \cellcolor{dojoSR!31.7664}44.12 & 32.25 & \cellcolor{dojoSR!6.1992}8.61 & 7.33 & \cellcolor{dojoSR!1.2456}1.73 & 1.58\\
\raisebox{-0.8mm}{\includegraphics[width=3.5mm,height=3.5mm]{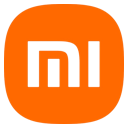}}\hspace{2pt}Xiaomi-Robotics-1 & \colorbox{dojoVla!12}{\textcolor{dojoVla}{\scriptsize VLA}} & \cellcolor{dojoSR!14.4504}20.07 & 13.93 & \cellcolor{dojoSR!16.9488}23.54 & 17.00 & \cellcolor{dojoSR!19.2168}26.69 & 18.83 & \cellcolor{dojoSR!27.6408}38.39 & 23.67 & \cellcolor{dojoSR!5.6232}7.81 & 6.56 & \cellcolor{dojoSR!2.8368}3.94 & 3.58\\
\raisebox{-0.8mm}{\includegraphics[width=3.5mm,height=3.5mm]{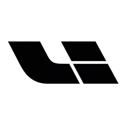}}\hspace{2pt}MachEmbodied-U0 & \colorbox{dojoWam!12}{\textcolor{dojoWam}{\scriptsize WAM}} & \cellcolor{dojoSR!12.8016}17.78 & 11.51 & \cellcolor{dojoSR!11.5704}16.07 & 8.78 & \cellcolor{dojoSR!17.4240}24.20 & 16.67 & \cellcolor{dojoSR!26.3952}36.66 & 22.25 & \cellcolor{dojoSR!5.8752}8.16 & 6.67 & \cellcolor{dojoSR!2.7216}3.78 & 3.17\\
\raisebox{-0.8mm}{\includegraphics[width=3.5mm,height=3.5mm]{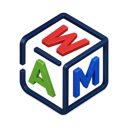}}\hspace{2pt}OpenWAM-$\alpha$ & \colorbox{dojoWam!12}{\textcolor{dojoWam}{\scriptsize WAM}} & \cellcolor{dojoSR!12.3696}17.18 & 11.92 & \cellcolor{dojoSR!14.9112}20.71 & 14.83 & \cellcolor{dojoSR!13.2840}18.45 & 9.25 & \cellcolor{dojoSR!25.1496}34.93 & 25.33 & \cellcolor{dojoSR!7.4952}10.41 & 9.11 & \cellcolor{dojoSR!1.0152}1.41 & 1.08\\
\raisebox{-0.8mm}{\includegraphics[width=3.5mm,height=3.5mm]{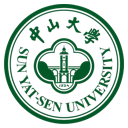}}\hspace{2pt}EWAM & \colorbox{dojoWam!12}{\textcolor{dojoWam}{\scriptsize WAM}} & \cellcolor{dojoSR!10.9440}15.20 & 9.37 & \cellcolor{dojoSR!10.8864}15.12 & 9.67 & \cellcolor{dojoSR!15.9480}22.15 & 12.33 & \cellcolor{dojoSR!21.7656}30.23 & 18.08 & \cellcolor{dojoSR!5.3208}7.39 & 6.00 & \cellcolor{dojoSR!0.8064}1.12 & 0.75\\
\raisebox{-0.8mm}{\includegraphics[width=3.5mm,height=3.5mm]{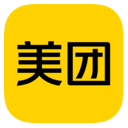}}\hspace{2pt}Meituan-Robotics-0 & \colorbox{dojoVla!12}{\textcolor{dojoVla}{\scriptsize VLA}} & \cellcolor{dojoSR!10.7640}14.95 & 9.53 & \cellcolor{dojoSR!9.9000}13.75 & 8.17 & \cellcolor{dojoSR!12.0744}16.77 & 7.75 & \cellcolor{dojoSR!21.3192}29.61 & 18.58 & \cellcolor{dojoSR!7.2432}10.06 & 8.89 & \cellcolor{dojoSR!3.2688}4.54 & 4.25\\
\raisebox{-0.8mm}{\includegraphics[width=3.5mm,height=3.5mm]{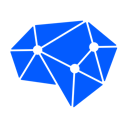}}\hspace{2pt}HWM-CogWAM & \colorbox{dojoWam!12}{\textcolor{dojoWam}{\scriptsize WAM}} & \cellcolor{dojoSR!10.3248}14.34 & 10.34 & \cellcolor{dojoSR!9.2016}12.78 & 9.28 & \cellcolor{dojoSR!14.7888}20.54 & 14.58 & \cellcolor{dojoSR!19.6920}27.35 & 18.17 & \cellcolor{dojoSR!5.8896}8.18 & 7.00 & \cellcolor{dojoSR!2.0592}2.86 & 2.67\\
\raisebox{-0.8mm}{\includegraphics[width=3.5mm,height=3.5mm]{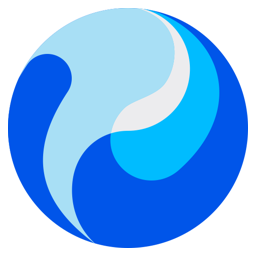}}\hspace{2pt}Hy-Embodied-0.5-VLA & \colorbox{dojoVla!12}{\textcolor{dojoVla}{\scriptsize VLA}} & \cellcolor{dojoSR!9.4104}13.07 & 8.80 & \cellcolor{dojoSR!8.4816}11.78 & 8.39 & \cellcolor{dojoSR!9.9432}13.81 & 8.00 & \cellcolor{dojoSR!18.5328}25.74 & 14.92 & \cellcolor{dojoSR!9.6264}13.37 & 12.11 & \cellcolor{dojoSR!0.4680}0.65 & 0.58\\
\raisebox{-0.8mm}{\includegraphics[width=3.5mm,height=3.5mm]{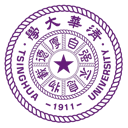}}\hspace{2pt}KinRT & \colorbox{dojoVla!12}{\textcolor{dojoVla}{\scriptsize VLA}} & \cellcolor{dojoSR!9.3744}13.02 & 8.80 & \cellcolor{dojoSR!10.0944}14.02 & 8.61 & \cellcolor{dojoSR!11.2680}15.65 & 9.92 & \cellcolor{dojoSR!19.0080}26.40 & 18.08 & \cellcolor{dojoSR!3.4704}4.82 & 3.56 & \cellcolor{dojoSR!3.0456}4.23 & 3.83\\
\raisebox{-0.8mm}{\includegraphics[width=3.5mm,height=3.5mm]{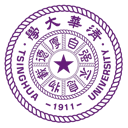}}\hspace{2pt}SimpleMemVLA & \colorbox{dojoVla!12}{\textcolor{dojoVla}{\scriptsize VLA}} & \cellcolor{dojoSR!9.0576}12.58 & 9.27 & \cellcolor{dojoSR!4.5792}6.36 & 3.95 & \cellcolor{dojoSR!5.3424}7.42 & 2.92 & \cellcolor{dojoSR!10.4976}14.58 & 5.50 & \cellcolor{dojoSR!24.2712}33.71 & 33.22 & \cellcolor{dojoSR!0.6120}0.85 & 0.75\\
\raisebox{-0.8mm}{\includegraphics[width=3.5mm,height=3.5mm]{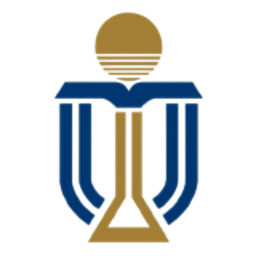}}\hspace{2pt}Spatial Forcing & \colorbox{dojoVla!12}{\textcolor{dojoVla}{\scriptsize VLA}} & \cellcolor{dojoSR!8.9136}12.38 & 8.04 & \cellcolor{dojoSR!10.1664}14.12 & 9.34 & \cellcolor{dojoSR!12.4704}17.32 & 10.58 & \cellcolor{dojoSR!16.7472}23.26 & 14.58 & \cellcolor{dojoSR!3.9096}5.43 & 4.11 & \cellcolor{dojoSR!1.2816}1.78 & 1.58\\
\raisebox{-0.8mm}{\includegraphics[width=3.5mm,height=3.5mm]{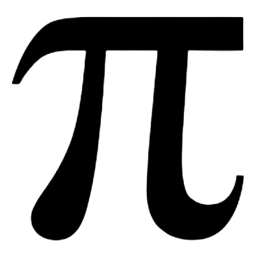}}\hspace{2pt}Pi-05 & \colorbox{dojoVla!12}{\textcolor{dojoVla}{\scriptsize VLA}} & \cellcolor{dojoSR!8.2368}11.44 & 6.93 & \cellcolor{dojoSR!9.6336}13.38 & 8.17 & \cellcolor{dojoSR!8.9280}12.40 & 5.50 & \cellcolor{dojoSR!16.9488}23.54 & 14.67 & \cellcolor{dojoSR!4.2408}5.89 & 4.67 & \cellcolor{dojoSR!1.4256}1.98 & 1.67\\
\raisebox{-0.8mm}{\includegraphics[width=3.5mm,height=3.5mm]{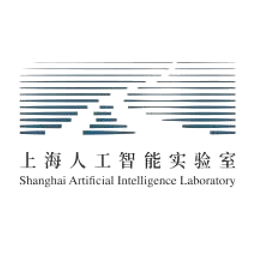}}\hspace{2pt}InternVLA-A1.5 & \colorbox{dojoVla!12}{\textcolor{dojoVla}{\scriptsize VLA}} & \cellcolor{dojoSR!8.0280}11.15 & 7.14 & \cellcolor{dojoSR!7.4520}10.35 & 6.83 & \cellcolor{dojoSR!10.9656}15.23 & 10.17 & \cellcolor{dojoSR!17.1360}23.80 & 13.75 & \cellcolor{dojoSR!3.5496}4.93 & 3.56 & \cellcolor{dojoSR!1.0296}1.43 & 1.42\\
\bottomrule\end{tabular}
\par\smallskip\begin{minipage}{\textwidth}\footnotesize
Official board updated 28 October 2026, accessed 29 October 2026 \citep{c8robodojo}. All entries through InternVLA-A1.5 (Average Score 11.15) are ordered together by Average Score, including Agent entries listed separately on the website. Paired columns report progress Score and task success rate (SR). Blue intensity increases with Score using the same mapping in all six groups; SR cells remain unshaded. Generalization averages standard and randomized settings; Average retains the source macro aggregate. Marks and types follow the board; training and observation/control support vary across submissions. The protocol specifies three simulation seeds, hidden-layout verification and a fixed system across tasks \citep{c8robodojo}; this snapshot neither confirms reevaluation of every entry under that protocol nor provides uncertainty for each row. Astra: 42 tasks, 50 episodes/task, medium effort, 100 calls/episode \citep{zhang2026unexpectedrobotpolicyearly}.
\end{minipage}
\end{table*}

\begin{table*}[!t]\centering\small
\setlength{\tabcolsep}{4pt}
\caption{EAA: performance of ten large models acting as agents on shared planning and manipulation tasks, with resource statistics for planning.}\label{tab:c8-agent-models}
\begin{tabular}{@{}lrrrrrr@{}}\toprule
 & \multicolumn{2}{c}{Task outcomes (\%)} & \multicolumn{4}{c}{Planning resources}\\
\cmidrule(lr){2-3}\cmidrule(lr){4-7}
Model & Planning & Manipulation & \shortstack{Median\\time (s)} & IQR (s) & \shortstack{Failed\\tokens (\%)} & \shortstack{Usage\\records}\\\midrule
GPT-6 Astra & 80.3 & 41.8 & 231.7 & 153.2--361.8 & 55.5 & 157/157\\
GPT-6 Sol & 44.6 & 19.8 & 280.7 & 199.9--427.1 & 76.0 & 157/157\\
Claude Fable 5.1 & 63.7 & 24.7 & 109.5 & 62.0--201.4 & 64.9 & 157/157\\
Qwen 3.8 Max & 36.9 & 9.3 & 300.0 & 216.8--506.7 & 81.2 & 145/157\\
Gemini 3.8 Flash & 48.4 & 18.7 & 61.8 & 28.6--175.9 & 77.1 & 157/157\\
Qwen 3.5 397B-A17B & 19.1 & 9.9 & 106.5 & 75.8--166.7 & 91.9 & 157/157\\
Qwen 3.5 27B & 19.1 & 11.0 & 94.5 & 60.8--146.8 & 94.3 & 157/157\\
Claude Opus 5.5 & 48.4 & 47.8 & 135.6 & 105.3--211.0 & 74.9 & 157/157\\
DeepSeek-V4.1-Flash-1017 & 31.2 & 20.9 & 57.1 & 36.1--78.5 & 83.8 & 157/157\\
Gemini 4 Argon & 70.1 & 44.0 & 159.5 & 51.5--612.1 & 64.0 & 157/157\\
\bottomrule\end{tabular}
\par\smallskip\begin{minipage}{\textwidth}\footnotesize
Source: Tables 2 and A42 of the EAA paper (arXiv version 3); model identifiers follow Table A5 \citep{c8eaa}. Planning uses 157 assignments; manipulation uses 182 binary assignments, excluding one continuous-reward task. Each has one episode per model and assignment. Nine missing Fable ManiSkill records receive no success credit in the reported total. Interfaces, observation schedules and budgets follow the evaluated configurations (Appendix E.1), not uniform control conditions. Planning time is recorded elapsed time; IQR is the 25th--75th percentile interval. Qwen HumanCLAW times include setup, retries and teardown. Failed tokens is the share of recorded input and output tokens spent in unsuccessful planning episodes; usage records state coverage. It is neither a token count nor a monetary cost.
\end{minipage}\end{table*}

\textbf{Cross-domain execution.} RobotWorld shows that the reported overall ordering does not hold uniformly across physical task domains (Table~\ref{tab:c8-system-profiles}). Astra completes more tasks overall, but fewer aerial tasks than Opus 5.5. Because RobotWorld retains one episode per model--task pair, these counts describe the evaluated task coverage rather than reliability across repeated trials. They locate differences in execution without identifying whether perception, planning, or control produced them. Token use, estimated API costs, and cumulative elapsed time give different views of resource use. Astra uses fewer tokens and less cumulative time than Opus 5.5, but incurs higher estimated API costs \citep{c8robotworld}. Together with task outcomes, these measures characterize execution under the reported interfaces and budgets.

\textbf{Profiles across task demands.} RoboDojo shows that a system's overall advantage need not hold across generalization, precision, long-horizon, memory, and open-vocabulary conditions (Table~\ref{tab:c8-robodojo-profiles}). For the organization of embodied behavior discussed in Section~\ref{sec:decision}, these profiles describe relative strengths under the evaluated task conditions. Score captures partial progress, while SR measures task completion. Table~\ref{tab:c8-robodojo-profiles} is ordered by Average Score, and shading shows Score across task dimensions; the comparisons below use SR to examine task completion. Among Agent configurations, PyRUA-Lean exceeds HexaShell-Preview in four dimensions but has lower memory success. GO-3 Nano Preview, labeled Agent + WAM, exceeds both in precision and long-horizon success, while PyRUA-Lean retains higher generalization and open-vocabulary success.

The VLA and WAM profiles also vary by task: WALL-OSS-1.0-lite has higher Average SR than VPP2, but lower precision and long-horizon success. This variation motivates reading system results alongside the execution capabilities discussed in Section~\ref{sec:actionSpaces}. The board labels describe submitted configurations, not levels of autonomy. Training, observations, and runtime support differ across these configurations..

\textbf{Large models acting as agents.} The relative performance of the ten VLM agents in Embodied Agent Arena (EAA) differs between its shared planning and manipulation task sets (Table~\ref{tab:c8-agent-models}) \citep{c8eaa}. Astra achieves the highest planning success, whereas Opus 5.5 leads manipulation when tasks are weighted equally. Equal weighting of the thirteen manipulation sources instead places Gemini 4 Argon above Opus 5.5, showing that the observed ordering depends on the task mixture. In its interactive tasks, models select operations, execute code or actions, and use returned observations to continue toward the goal. These outcomes characterize each model within its evaluated interaction configuration.

Planning resources complement these outcomes. The table reports median elapsed time and the share of recorded tokens consumed in unsuccessful planning episodes \citep{c8eaa}. Short duration can reflect early failure; the token share describes where recorded consumption occurs, not whether that computation was useful. These statistics and RobotWorld's aggregate resource totals have different scopes and are interpreted within their respective benchmarks. Because these cross-system results reflect complete configurations, the following within-study comparisons examine how changes in execution support or organization relate to outcomes under specified task conditions.

\subsubsection{Within-Study Configuration Comparisons}
Within-study comparisons relate performance changes to modifications in execution support or system organization. Table~\ref{tab:c8-organization-results} summarizes policy assistance, hierarchical organization, and online supervision; each comparison is interpreted within its own task conditions and outcome definition. Each block preserves the selected comparison's reported condition families, including unchanged and negative outcomes.

\begin{table*}[!t]
\centering\small
\setlength{\tabcolsep}{6pt}\renewcommand{\arraystretch}{1.32}
\caption{Within-study configuration comparisons. Explicit column headings identify the configurations compared in each study. Numerical entries are source-reported percentages; comparisons apply within a study and task condition.}\label{tab:c8-organization-results}
\definecolor{shiftUp}{HTML}{EEA064}
\definecolor{shiftDown}{HTML}{79AD78}
\definecolor{shiftEqual}{HTML}{53616E}
\begin{tabularx}{\textwidth}{@{}>{\raggedright\arraybackslash}p{0.31\textwidth}*{3}{>{\centering\arraybackslash}X}@{}}
\toprule
\multicolumn{4}{@{}l}{\textbf{A. Policy assistance [G]}}\\
Task condition & Astra Direct & Astra + $\pi_{0.5}$ & \shortstack{Performance\\shift}\\
\midrule
RoboDojo, local ten-task subset &26.0&48.0 & \raisebox{-1.4pt}{\includegraphics[width=9.5pt,height=9.5pt]{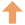}}\\
RoboCasa, atomic seen &28.0&52.0 & \raisebox{-1.4pt}{\includegraphics[width=9.5pt,height=9.5pt]{figures/shift-up.pdf}}\\
RoboCasa, composite seen &16.0&28.0 & \raisebox{-1.4pt}{\includegraphics[width=9.5pt,height=9.5pt]{figures/shift-up.pdf}}\\
RoboCasa, composite unseen &56.0&36.0 & \raisebox{-1.4pt}{\includegraphics[width=9.5pt,height=9.5pt]{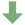}}\\
\midrule
\multicolumn{4}{@{}l}{\textbf{B. High-level organization [H]}}\\
Task condition & Flat VLA & Best Hierarchy & \shortstack{Performance\\shift}\\
\midrule
Short horizon &$69.63\pm1.07$&$78.22\pm0.91$ & \raisebox{-1.4pt}{\includegraphics[width=9.5pt,height=9.5pt]{figures/shift-up.pdf}}\\
Long horizon &$25.30\pm1.22$&$67.08\pm1.38$ & \raisebox{-1.4pt}{\includegraphics[width=9.5pt,height=9.5pt]{figures/shift-up.pdf}}\\
Reasoning &$50.90\pm1.20$&$80.89\pm1.17$ & \raisebox{-1.4pt}{\includegraphics[width=9.5pt,height=9.5pt]{figures/shift-up.pdf}}\\
\midrule
\multicolumn{4}{@{}l}{\textbf{C. Online supervision [R]}}\\
Task condition & Planner + OpenVLA & Hierarchical framework & \shortstack{Performance\\shift}\\
\midrule
Random disturbance &18.63&18.63 & \raisebox{-1.4pt}{\includegraphics[width=9.5pt,height=9.5pt]{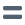}}\\
Observation mismatch &19.45&19.18 & \raisebox{-1.4pt}{\includegraphics[width=9.5pt,height=9.5pt]{figures/shift-down.pdf}}\\
Memory exploration &8.04&9.06 & \raisebox{-1.4pt}{\includegraphics[width=9.5pt,height=9.5pt]{figures/shift-up.pdf}}\\
Memory execution &16.69&17.83 & \raisebox{-1.4pt}{\includegraphics[width=9.5pt,height=9.5pt]{figures/shift-up.pdf}}\\
Mixed conditions &11.48&13.21 & \raisebox{-1.4pt}{\includegraphics[width=9.5pt,height=9.5pt]{figures/shift-up.pdf}}\\
Ideal conditions &21.92&21.10 & \raisebox{-1.4pt}{\includegraphics[width=9.5pt,height=9.5pt]{figures/shift-down.pdf}}\\
\bottomrule
\end{tabularx}
\par\smallskip\begin{minipage}{\textwidth}\footnotesize
[G] Galbot Team, Tables 6/14 \citep{c8galbotastra}: binary completion; 50 paired RoboDojo instances; 25 shared RoboCasa starts/group. Contextual/feedback support also changes. The aggregate RoboCasa comparison is not statistically resolved ($p=0.5235$).\par
[H] Hi-VLA v2, Table 1 \citep{hu2026orchestrating}: Gemini/Gemini Robotics On-Device; a fixed VLA for each task, five tasks/category, 200 trials/task; means $\pm$ SE. Best Hierarchy jointly changes reasoning, memory, observations and termination.\par
[R] RoboCerebra v2, Table 3 \citep{c8robocerebra}: GPT-4o planner and adapted OpenVLA; 60 tasks $\times$ 10 trials; supplied anchor-aligned transitions. SR measures completion of key state transitions, not binary full-task success. The hierarchy adds monitoring and memory together.
\par Performance shift compares the second configuration with the first within each row: \raisebox{-1.1pt}{\includegraphics[width=7.5pt,height=7.5pt]{figures/shift-up.pdf}} higher reported value, \raisebox{-1.1pt}{\includegraphics[width=7.5pt,height=7.5pt]{figures/shift-down.pdf}} lower reported value, and \raisebox{-1.1pt}{\includegraphics[width=7.5pt,height=7.5pt]{figures/shift-equal.pdf}} equal at the reported precision. Symbols indicate direction only; they do not encode the magnitude or statistical significance of the difference.
\end{minipage}
\end{table*}

\textbf{Task-dependent gains.} In [G], the policy-assisted configuration has higher reported success in the displayed seen RoboCasa conditions, whereas reported success on unseen compositions decreases from 56\% to 36\% (14 versus 9 successes among 25 starts). The aggregate RoboCasa comparison is not statistically resolved, and the configurations also change contextual support. In [H], the best hierarchy has higher reported means than the flat VLA, with the largest difference on long-horizon tasks. RoboTeam illustrates a further trade-off: across four agent configurations that completed similar numbers of simulated multi-arm tasks, the single agent used fewer tokens but took longer than the hierarchical and decentralized teams \citep{fan2026roboteam}. Together, these patterns motivate evaluating organization in relation to task demands and executor capabilities. They do not isolate individual mechanisms; in [H], termination changes together with reasoning, memory and observations, preventing attribution of the gain to the termination mechanism alone. These comparisons also do not establish a universal advantage for hierarchical control.

\textbf{Supervision has uneven outcomes.} In [R], the reported point estimates increase for memory and mixed conditions, are equal at the reported precision for disturbance, and decrease for observation mismatch and ideal conditions. These small differences warrant condition-specific assessment rather than a general claim of improved correction. The configuration adds monitoring and memory together; separate measurements of detection, intervention, and executed repair would be needed to identify their individual contributions.

\subsubsection{Synthesis and Limits of the Reported Evidence}
\label{sec:c8-synthesis}
The reported comparisons show that overall performance does not fully describe a system's strengths across task demands. Rankings vary across domains or capability conditions, while changes to execution support and organization yield gains, unchanged outcomes, or losses depending on the evaluated condition (Tables~\ref{tab:c8-system-profiles}--\ref{tab:c8-organization-results}). Resource statistics add another dimension to these comparisons: token use, estimated cost, and elapsed time describe different aspects of consumption and must be read alongside task outcomes. The selected results therefore support assessing system configurations in relation to particular tasks and operating conditions.

The notation in Sections~\ref{sec:perception}--\ref{sec:feedback} helps specify the target of a comparison: information retained in $m_t$ or supplied through $c_t$, behavior selected by $\mu$ and executed by $\pi_\omega$, assessments $y_t$, and changes retained in $\Theta$. These targets require different evidence: task success alone does not directly measure the accuracy of an internal representation or assessment, while changes to $\Theta$ must be evaluated through their effects on subsequent performance and retention under comparable evaluation conditions.

These comparisons also limit what can be attributed to agentic mechanisms. Cross-system results reflect differences in training, tools, observation access, and budgets; within-study comparisons may change several mechanisms together. Where these factors vary together, the reported differences cannot be attributed to an individual mechanism. The numerical evidence reviewed here principally concerns execution and organization in simulation. It does not by itself establish sustained physical autonomy or improvement retained across tasks. Such claims require additional evidence under the physical and cross-task evaluation conditions discussed in Sections~\ref{sec:c8-execution} and~\ref{sec:c8-feedback}. Section~\ref{sec:discussion-evaluation} considers how future studies could obtain stronger evidence for these broader claims.